\documentclass[pdflatex,sn-mathphys-num]{sn-jnl}

\usepackage{graphicx}
\usepackage{multirow}
\usepackage{amsmath,amssymb,amsfonts}
\usepackage{amsthm}
\usepackage[title]{appendix}
\usepackage{xcolor}
\usepackage{textcomp}
\usepackage{manyfoot}
\usepackage{booktabs}
\usepackage{algorithm}
\usepackage{algorithmicx}
\usepackage{algpseudocode}
\usepackage{listings}
\usepackage{setspace}
\usepackage{comment}
\usepackage{makecell}
\usepackage{float}
\usepackage{placeins}
\usepackage{subcaption}
\usepackage{array}
\usepackage{tabularx}
\usepackage{microtype}
\microtypesetup{nopatch=footnote}
\usepackage{xurl}
\def\burl#1{\url{#1}}

\theoremstyle{thmstyleone}

\theoremstyle{thmstyletwo}

\theoremstyle{thmstylethree}

\begin{document}

\title[Physics-Structured Surrogate Modeling for Wing Design]
{Physics-Structured Surrogate Modeling and Conformal Robust Multipoint Optimization for Glider Wing Design}

\author*[1]{\fnm{Arash} \sur{Fath Lipaei}}
\email{alipaei25@edu.ku.tr}
\equalcont{These authors contributed equally to this work.}

\author[2]{\fnm{AmirHossein} \sur{Ghaemi}}
\equalcont{These authors contributed equally to this work.}

\author[3]{\fnm{Melika} \sur{Sabzikari}}
\equalcont{These authors contributed equally to this work.}

\affil*[1]{
\orgdiv{Graduate School of Sciences and Engineering}, \orgname{Ko\c{c} University}, \orgaddress{\street{Rumelifeneri Yolu}, \city{Sar\i yer}, \state{Istanbul}, \postcode{34450}, \country{Turkey}}
}

\affil[2]{
\orgname{Independent Researcher}
}

\affil[3]{
\orgdiv{Department of Mechanical Engineering}, \orgname{Imperial College London}, \orgaddress{\street{Exhibition Road}, \city{London}, \postcode{SW7 2AZ}, \country{United Kingdom}}
}

\abstract{Aerodynamic design using surrogate assistance can lower the cost of concept design. Point accuracy, however, is not enough to ensure that the optimizer does not exploit any part of space which is uncertain or with low confidence. In this work we develop a locked physics-structured surrogate and robust multi-point framework for the early-stage design of glider wings. The framework utilizes a dataset of 150,000 Tornado vortex lattice simulations, which provides 16 continuous targets for the aerodynamics, root loads, and flight dynamics. A five-member dual-head ensemble distinguishes between similarity-based aerodynamic inputs and the physical structure required for dimensional dynamics, while exact decoder recovers dimensional forces and root-load proxy values. Split-conformal prediction gives simultaneous intervals for the 14 optimization outputs, while at the same time a nearest-neighbor support score limits the extrapolation. The in-distribution test resulted in a mean NRMSE of 0.0223, while the structured out-of-distribution test resulted in 0.0595. The global joint 95\% intervals covered 94.71\% of points in distribution and support conditioned calibration gave 91.08\% coverage under structured shift. Three-speed search examined 16,384 geometries and kept 2,998 feasible designs, 198 of which are nondominated designs. After freezing 20 wings, 60 Tornado simulations were conducted. The 60 simulations resulted in a mean NRMSE of 0.0228, coverage for 58 of 60 operating points, and hard feasibility success for all 60. For all 20 wings, all three objective upper bounds are conservative. The primary contribution of this paper is the combination of structured multi-output learning, simultaneous calibration, support aware robust Pareto search and locked post-selection simulations.}


\maketitle

\section{Introduction}
\label{sec:introduction}

The design of wing planform is a highly coupled problem, as chord, span, taper, sweep, twist, dihedral, and flight conditions simultaneously affect the aerodynamic performance, trim, loading, and flight dynamics derivatives. This problem is especially relevant in case of glider and other aircraft with high aspect-ratio, where drag reduction is desired, but larger wings increase the risk of root bending and sensitivity to off-design conditions. Despite the fact that high-fidelity computational fluid dynamics (CFD) and adjoint approach allow one to solve the problem of viscous and compressible flows in high detail, this method requires expensive and demanding computation process for each case in large design space \cite{martins2022aerodynamic,abergo2023aerodynamic}. Therefore, low-fidelity lifting surface methods are still valuable for preliminary filtering, given that their limitations are clearly stated. The vortex lattice method (VLM), including its Tornado version used here, provides fast calculation of integrated lifting surface derivatives for the case of attached and incompressible flow \cite{melin2000vortex,leifsson2014fast}.

Despite being designated as a low fidelity model in comparison with viscous computational fluid dynamics, it should be noted that it is not the noisy data or lack of physical meaning that distinguishes this modeling approach. VLM is a deterministic three-dimensional, lifting-surface formulation, which solves the problem of the finite-span circulation distribution, induced velocity and downwash, spanwise loading, forces and moments and drag due to lift. The classical vortex lattice research demonstrated the convergence to lifting surface solutions as well as satisfactory agreement with theoretical and experimental lift, pitching moment, span loading, and induced drag results for proper subsonic attached flow cases \cite{paulson1976applications,blackwell1976induced,
deyoung1976optimum}. Based on the current glider geometry, speed interval and altitude, the database can be considered low-subsonic flight case for finite wings, which is consistent with conceptual glider analysis.

Surrogate modeling reduces the cost of repeated aerodynamic evaluations even further by capturing the relationship between geometry, operating condition, and solver response. Classical response surfaces, radial basis functions, Kriging, Gaussian processes, and efficient global-optimization methods are well-known within engineering design \cite{Forrester2008,Jones1998,asouti2023radial}. More recent work uses neural networks, active learning, transfer learning, and multi-fidelity models for aerodynamic predictions and shape optimization \cite{li2021data,sabater2022fast,zhang2024active,tao2024multi,nikolaou2025multi,liu2025review}. Although these methods could provide both accurate and cheap response models, there are important considerations when such a model is embedded within an optimizer as opposed to being evaluated only on a random test set.

First, an optimizer will seek out regions where errors are favorable on purpose, therefore, the average regression accuracy alone is insufficient to guarantee a reliable design. Second, uncertainty estimation based on a neural ensemble does not necessarily satisfy the joint coverage requirement when multiple outputs and constraints have to be met simultaneously, and predictive uncertainty tends to become unreliable as the input distribution changes \cite{lakshminarayanan2017deep,ovadia2019trust}. Third, confidence intervals for individual coefficients are not equal to joint uncertainty for a vector of coupled aerodynamic and dynamic responses. Split conformal prediction provides a distribution-free prediction coverage in finite samples under exchangeability assumptions, and conformal uncertainty sets were proven to be directly related to robust optimization \cite{lei2018distribution,johnstone2021conformal,feldman2023calibrated}. Lastly, wing optimization requires the consideration of multiple conditions, and aerodynamic loads proxies should be distinguished from structural quantities; a low-drag design might turn out to be a bad choice when lift robustness, root bending, damping derivatives, and extrapolation risk are taken into account.

This paper addresses the mentioned issues through a locked surrogate-to-simulation pipeline. The database of 150,000 Tornado evaluations is divided into training, validation, calibration, in-distribution test, hidden optimization and structured out-of-distribution subsets. Sixteen continuous outputs are learned, as opposed to merely labelling the flight-dynamics behavior. The physics-structured two-headed ensemble splits the similarity-based aerodynamic input space from the full dimensional space used for flight-dynamics derivatives, whereas exact analytical decoding is used to restore lift, drag and root loads. The normalized maximum score conformal procedure produces an uncertainty rectangle simultaneously on the 14 outputs used for optimization. The nearest neighbor support condition prevents unsupported extrapolation, and the calibrated model is used for robust three-speed Pareto optimization. Prior to simulating 60 Tornado designs for evaluation, 20 designs are frozen.

The methodology novelty comes from the combination of these elements in a frozen information pipeline: physics-structured multi-output regression, simultaneous conformal calibration, support-conditioned multi-point Pareto search and post-selection simulation of frozen designs.

The contributions made are:
\begin{enumerate}
    \item A database for aerodynamics design which includes continuous aerodynamics parameters, normalized root loading parameters, and dimensional flight dynamics parameters, with accurate physical interpretation of dimensional force and moment.
    \item A five-member physics-structured ensemble network, where the aerodynamics head uses a compact similarity representation, while the dynamics head retains the complete physics-based feature vector, tested against ridge regression, LightGBM, unconstrained compact multilayer perceptron, and a single structured network.
    \item A simultaneously normalized split-conformal prediction band on all optimization outputs, along with support-conditioned diagnostics and an explicit training-support constraint.
    \item A multi-point formulation balancing the conservative weighted drag, the worst root bending proxy, and the direction-derivative deficiency based on convention, while satisfying lift, damping sign, support, root shear, and root twist at three different speed levels.
    \item A lock evaluation on 60 synthetic Tornado cases, enabling direct analysis of point accuracy, simultaneous coverage, hard-feasibility success, conservative objective bounds, and Pareto status post-candidate selection.
\end{enumerate}

The rest of this document is structured in the following way. Section~\ref{sec:related_work} places the current framework within the context of aerodynamic surrogate optimization, physics-based learning, offline model-based optimization, and conformal decision strategies. Section~\ref{sec:methodology} describes the solver data, objective functions, frozen partitions, surrogate structure, conformal calibration, and robust optimization method. Section~\ref{sec:results} describes the surrogate models, uncertainties, optimizations, and simulations results. Sections~\ref{sec:discussion} and~\ref{sec:limitations} analyze the results and define their practical implications, while Section~\ref{sec:conclusions} outlines the main conclusions. Mathematical details are given in the appendices.

\section{Related Work and Methodological Positioning}
\label{sec:related_work}

\subsection{Surrogate-assisted and multipoint aerodynamic optimization}

Multi-point surrogate-assisted aerodynamic design has evolved from response surfaces and efficient global optimizations to adaptive, neural, and multi-fidelity models that enable expansive searches at a lower number of high-cost evaluations. Early versions of multi-point studies have used cokriging to share information through operating conditions and decrease the number of aerodynamic evaluations \cite{toal2011multipoint}. High-fidelity gradient-based studies then demonstrated how implementing multiple design points helps mitigate the off-design performance drop of single-point optimizers, including investigations of the Common Research Model wing \cite{kenway2016multipoint}. Recent advances in aerodynamic optimization have followed this path via off-design constraint models \cite{li2023efficient}, distributionally robust formulations \cite{chen2024data}, hierarchical-Kriging, expected-improvement multi-fidelity approaches for robust design \cite{zhang2024robustkriging}, transfer learning and convolutional multi-fidelity surrogates \cite{tao2024multi,wu2024efficient}, and multi-fidelity wing optimization at early designs \cite{nikolaou2025multi}. In operation-aware multi-point optimization, clustered operational data have been used to define representative wing-design conditions, avoiding solely relying on manually-prescribed points \cite{yang2025operationaware}.

Though these papers show the potential of multi-condition and multi-fidelity aerodynamic design, they answer a reliability question distinct from this work. The proposed framework seeks to address whether a multi-output low-fidelity surrogate model can allow expansive continuous exploration within the design space while maintaining calibrated simultaneous uncertainty estimation, allowing explicit control of the support region, and guaranteeing a locked solver testing after selection. It is important to note that a design may be multi-point and still utilize the surrogate in a region with low support, and a very precise surrogate may still provide inappropriate uncertainty estimates.

\subsection{Physics-guided and dimensionally structured learning}

Physical insight and scientific principles can be implemented in machine learning by governing-equation residuals, constrained losses, invariant or equivariant representations, hybrid model components, and analytic decoders \cite{karniadakis2021piml,willard2023scientific}. Dimensional analysis offers an especially relevant inductive bias for engineering regression. Buckingham-$\Pi$ constrained learning can discover or impose constraints that generate dimensionless groups \cite{bakarji2022buckingham}, whereas units-equivariant learning generates dimensionless inputs and analytically restores output dimensions \cite{villar2023dimensionless}. Such strategies suggest employing aspect ratio, taper, sweep, twist, dihedral angle, and angle of attack as a compact, similarity-oriented representation, along with dimensional reconstruction of forces and root-load proxies.

This paper, thus, uses the term \emph{physics-structured} in an intentionally constrained manner. The proposed network is neither a physics-informed neural network where the loss function minimizes a governing partial differential equation (PDE) residual, nor does it replace the vortex lattice solver. The structure is achieved via the following elements: (i) separate feature sets for normalized aerodynamic/load response and dimensional flight dynamics response; (ii) use of dimensionless or similarity-oriented variables where relevant; (iii) normalized coefficient targets; and (iv) exact decoding of known dynamic pressure, area, span, and chord scalings. In this sense, the architecture is closer to a knowledge-guided surrogate rather than a learned flow solver.

\subsection{Optimization against learned objectives and support control}

Since the training process of a design surrogate is based on a static dataset, the optimizer changes the effective query distribution. It prefers certain points that are either absent or rare in the training data as the prediction error there is favorable. It should be noted that the optimizer shift is different from the usual random-test generalization and is one of the key problems of offline model-based optimization. Conservative Objective Models for example, actively penalize the optimistic predictions in out-of-distribution designs \cite{trabucco2021com}. Generally speaking, deep predictive uncertainties are known to suffer under dataset shift \cite{ovadia2019trust}.

In the current case, the support control itself is much simpler compared to learning a conservative objective model. A standard 20-nearest-neighbor distance is implemented as an observable measure for proximity to the training design could, and the optimizer is restricted from choosing geometries which exceed the frozen calibration-derived threshold. The distance is the geometric constraint that complements calibrated residual-based uncertainty instead of substituting it. Afterward, 60-run external test evaluates whether both constraints were conservative enough for the selected continuous geometries.

\subsection{Conformal uncertainty for downstream optimization}

Split conformal prediction offers finite-sample marginal coverage for exchangeable sequences independent of a specific probabilistic model \cite{lei2018distribution}. For covariate shift, on the other hand, the extensions generally require further information or weighting assumptions, such as an estimable test-to-training density ratio \cite{tibshirani2019covshift}. In case of multi-output decisions, individual calibration of each response separately is insufficient to simultaneously guarantee the entire constraint vector. Multi-output and max-score constructions calibrate a joint nonconformity score to address this issue \cite{feldman2023calibrated}. Conformal uncertainty sets have also been applied in robust optimization tasks \cite{johnstone2021conformal}. When it comes to multi-output problems, calibrating each output independently generally does not lead to a simultaneous guarantee for the entire constraint vector. The multivariate conformal constructions solve this problem through calibration of the joint nonconformity static, while conformal uncertainty sets have also been introduced in robust optimization.

There exist approaches where the construction of uncertainty sets is guided by the downstream decision problem. Conformal Contextual Robust Optimization implements non-convex conformal regions in predict-then-optimize problems \cite{patel2024ccro}. End-to-end Conditional Robust Optimization, on the other hand, simultaneously trains the conditional uncertainty sets and the decisions \cite{chenreddy2024ecro}. End-to-end conformal calibration is another approach that uses the downstream loss function in training while preserving conformal calibration \cite{yeh2025endtoend}.

The proposed method instead involves a modular post-hoc construction. First, the ensemble is trained and then, the normalized maximum residual is calibrated on a disjoint set, and then a 14-output rectangular uncertainty region is passed to the optimizer. This method is less efficient in terms of task-specificity than decision-aware uncertainty sets. However, it allows comparison of the raw ensemble, marginal, global-joint, and support-conditioned intervals in the same experiment.

\begin{figure}[H]
\centering
\includegraphics[width=\textwidth]{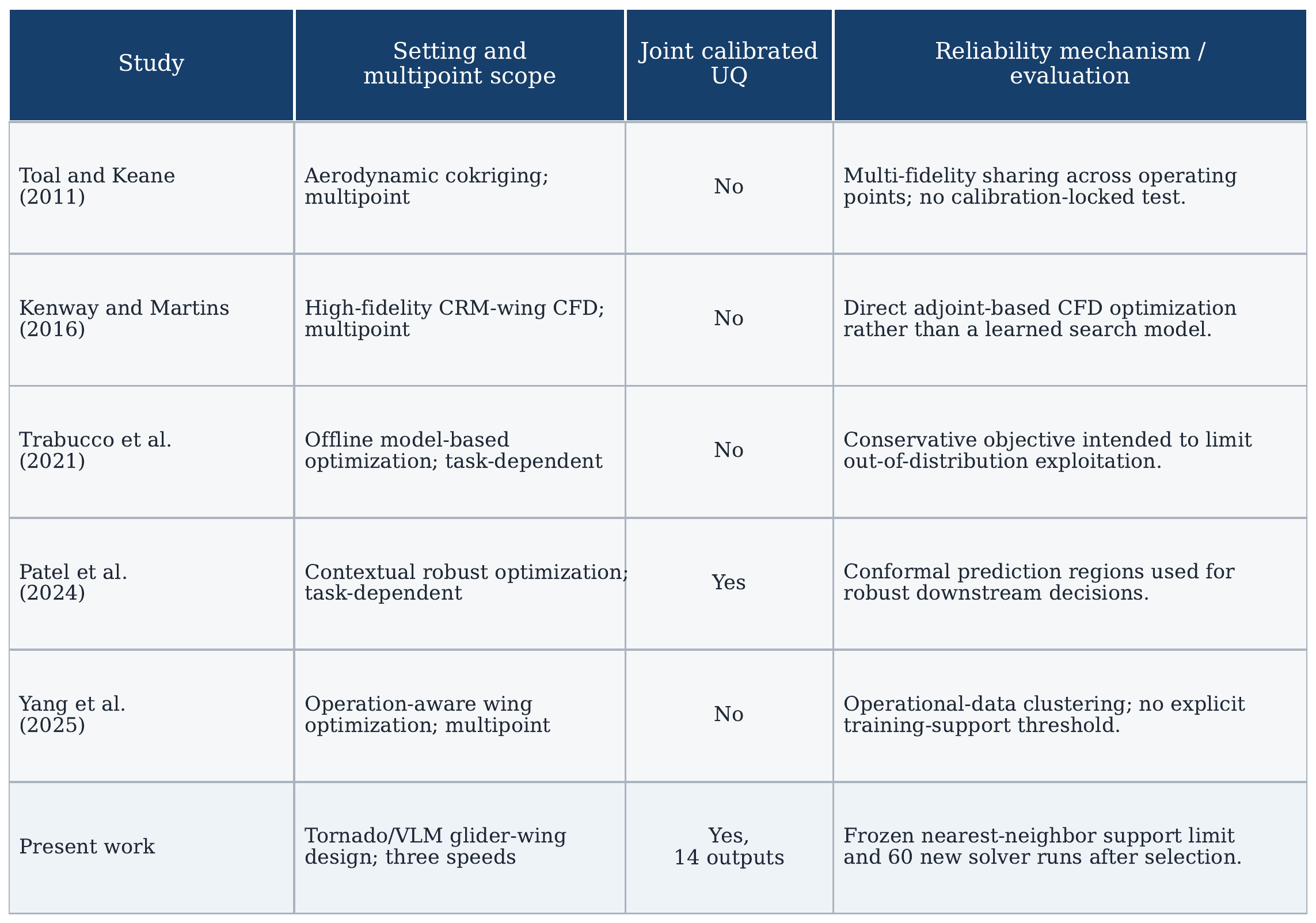}
\caption{Suggested methodology positioning with respect to multi-point aerodynamic optimization, offline model-based optimization, and conformal robust decision methods \cite{toal2011multipoint,kenway2016multipoint,trabucco2021com,patel2024ccro,yang2025operationaware}.}
\label{fig:related_work_positioning}
\end{figure}

Figure~\ref{fig:related_work_positioning} represents the key methodological difference between representative past works and the current workflow.

\section{Methodology}
\label{sec:methodology}

\subsection{Framework overview}
\label{sec:framework_overview}

This framework integrates aerodynamic simulations, physics-structured surrogate modeling, distribution-free uncertainty calibration, and robust multi-point optimization in a locked workflow. As the start, we use a database of 150,000 wings and operating-condition samples generated via Tornado vortex lattice method (VLM). VLM is an adequate approach to fast exploration of the design space for lifting-surface planforms, whereas surrogate modeling makes use of repeated solutions from the solver in the process of large-scale design space exploration \cite{melin2000vortex,leifsson2014fast}. In this study we focus on early-stage aerodynamic design; the solver-generated trends will be used for screening and comparison of candidate wings prior to more higher-fidelity viscous, structural, or aeroelastic studies.

Eight raw geometrical and operating parameters along with four analytically extracted physical properties are used to describe each sample. Sixteen continuous response values are learned: six aerodynamic or moment coefficient outputs, seven dimensional flight dynamics derivatives, and three normalized root loads coefficient. Dimensional lift, drag, root shear, root bending, and root twisting quantities are reconstructed via exact physical decoder without learning them as separate outputs. Such separation allows the neural networks to learn compact response surfaces with preserved known scaling with dynamic pressure, reference area, span, and mean aerodynamic chord.

The surrogate is a five-member physics-structured ensemble. The first similarity-based head takes aspect ratio, taper ratio, sweep, twist, dihedral, and angle of attack as inputs and outputs the aerodynamic and normalized load quantities. The second head takes the full physical feature vector and outputs the dimensional flight-dynamics derivatives. Bootstrap-style diversity in the ensemble predictors in utilized to reveal model disagreement, following the general idea behind ensemble predictors and deep ensembles \cite{breiman1996bagging,lakshminarayanan2017deep}. The spread of the ensemble members is not interpreted as a calibrated confidence interval on its own.

A separate calibration partition is used to convert the network mean and variance into finite-sample conformal intervals. For each target, a normalized split-conformal score is calculated, and the highest score from among the 14 quantities entering optimization is used to provide a simultaneous global certificate. This construction follows the distribution-free regression framework of split conformal inference and the more general use of multivariate conformal regions and uncertainty sets for conformal inference in downstream optimization tasks \cite{lei2018distribution,feldman2023calibrated,johnstone2021conformal}. Support-conditioned calibration is maintained for out-of-distribution analysis, while the global 95\% simultaneous interval is the official certificate for optimization.

\begin{figure}[t]
\centering
\includegraphics[width=\textwidth]{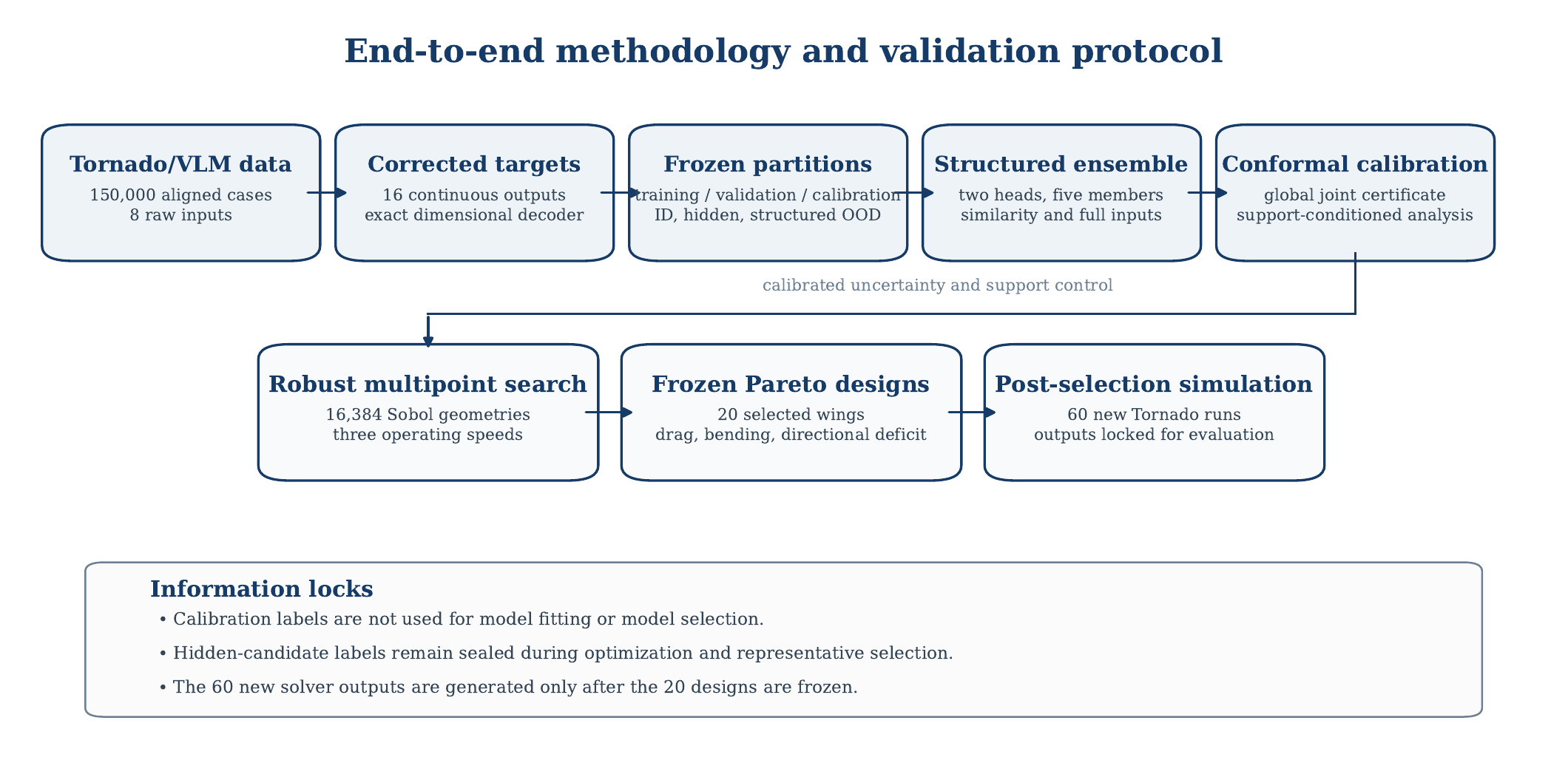}
\caption{Locked end-to-end workflow and information access protocol. Separated information used for training, calibration, candidate selection, and post-selection simulation is separated; the 60 Tornado outputs are produced only after freezing the 20 representative designs.}
\label{fig:framework_overview}
\end{figure}

The calibrated surrogate is part of a three-condition design problem. Using a scrambled Sobol sequence, 16,384 continuous geometries are created, and a separate angle of attack value is specified at each of three prescribed speeds, in order to trim the ensemble-mean lift to the aircraft weight. Designs are kept only if the entire conformal lift interval, derivative-sign conditions, root-load limit criteria, and nearest-neighbor support criterion hold for all three operating conditions. Next, nondominated sorting is performed with regard to the conservative weighted drag, worst root bending load estimate, and directional stability--damping deficit objectives. Then twenty Pareto designs are frozen before examining any new solver outputs. Finally, all 20 frozen geometries are analyzed at all three operating conditions, creating 60 Tornado simulations which are used solely for the purpose of external validation. The process is summarized in Figure~\ref{fig:framework_overview}.

\subsection{Wing design problem and modeling scope}
\label{sec:problem_formulation}

An evaluation of a candidate design can be represented using a geometry vector with six dimensions and an operating-state vector with two dimensions,
\begin{equation}
\mathbf{g}
=
\begin{bmatrix}
 c_r & \Lambda & b & \theta & \lambda & \Gamma
\end{bmatrix}^{\mathsf T},
\qquad
\mathbf{z}
=
\begin{bmatrix}
 \alpha & V
\end{bmatrix}^{\mathsf T},
\label{eq:geometry_state_vectors}
\end{equation}
where $c_r$ is the wing root chord, $\Lambda$ is the quarter-chord sweep angle fed to the Tornado geometry definition, $b$ is the semi-span, $\theta$ is geometric twist, $\lambda=c_t/c_r$ is the taper ratio, $\Gamma$ is dihedral, $\alpha$ is the angle of attack, and $V$ is the freestream speed. The network input state comprises $\mathbf{x}=[\mathbf{g}^{\mathsf T},\mathbf{z}^{\mathsf T}]^{\mathsf T}$. The bounds used to construct the solver database are presented in Table~\ref{tab:design_bounds}.

\begin{table}[t]
\centering
\caption{Raw variables used for geometry and operating conditions in the Tornado database. Here, $b$ refers to semi-span throughout the dataset, codes and manuscript.}
\label{tab:design_bounds}
\begin{tabular}{llll}
\toprule
Symbol & Meaning & Lower bound & Upper bound \\
\midrule
$c_r$ & Root chord [m] & 0.6 & 1.2 \\
$\alpha$ & Angle of attack [deg] & $-2.5$ & 7.5 \\
$\Lambda$ & Quarter-chord sweep angle [deg] & 0.0 & 5.0 \\
$b$ & Semi-span [m] & 5.0 & 10.0 \\
$\theta$ & Geometric twist [deg] & 0.0 & 5.0 \\
$\lambda$ & Taper ratio, $c_t/c_r$ & 0.3 & 0.6 \\
$\Gamma$ & Dihedral [deg] & 0.0 & 6.0 \\
$V$ & Freestream speed [m s$^{-1}$] & 36.0 & 54.0 \\
\bottomrule
\end{tabular}
\end{table}

The span convention is critical since the original simulator input uses the distance from the aircraft centerline to one wingtip. Therefore,
\begin{equation}
B=2b,
\label{eq:full_span}
\end{equation}
where $B$ refers to full span. In the case of symmetric linearly tapered planform, the reference area, aspect ratio, and mean aerodynamic chord are
\begin{equation}
S=b c_r(1+\lambda),
\qquad
AR=\frac{(2b)^2}{S},
\qquad
\bar{c}=\frac{2}{3}c_r\frac{1+\lambda+\lambda^2}{1+\lambda}.
\label{eq:derived_geometry_main}
\end{equation}
These formulas are used both to calculate the surrogate features and to decode dimensional forces and root-load proxies. Their derivation can be found in Appendix~\ref{app:geometry_decoder}. All through this paper, the variable $b$ used in the simulator refers to semi-span; therefore, $2b$ is the tip-to-tip full span.

The multi-point mission fixes the aircraft weight to
\begin{equation}
W=5886~\mathrm{N}
\label{eq:aircraft_weight}
\end{equation}
and evaluates every geometry at
\begin{equation}
V_k\in\{38,45,52\}~\mathrm{m\,s^{-1}},
\qquad
w_k\in\{0.25,0.50,0.25\},
\qquad k=1,2,3.
\label{eq:mission_conditions}
\end{equation}
While the geometry $\mathbf{g}$ is shared by all operating conditions, a separate $\alpha_k\in[-2.5^{\circ},7.5^{\circ}]$ is obtained via bisection on the ensemble-mean lift for each speed. This results in a trimmed multi-point state
\begin{equation}
\mathcal{X}(\mathbf{g})
=
\left\{
(\mathbf{g},\alpha_k,V_k)
\right\}_{k=1}^{3}.
\label{eq:multipoint_state}
\end{equation}
The following robust-feasibility test demands that the complete calibrated lift interval stays within $\pm5\%$ of $W$ at all conditions, as opposed to only requiring the point estimate to match the target.

The learned output vector consists of 16 continuous variables,
\begin{equation}
\mathbf{y}
=
\begin{bmatrix}
C_L, C_D, C_m, C_{m_\alpha}, C_{\ell_p}, C_{m_q},\\
F_{YV}, F_{ZW}, M_\alpha, L_\beta, N_\beta, L_p, N_r,\\
C_{\mathrm{shear}}, C_{\mathrm{bend}}, C_{\mathrm{twist}}
\end{bmatrix}^{\mathsf T}.
\label{eq:learned_outputs}
\end{equation}
The first six terms are force or moments coefficients and longitudinal or rotational derivatives. The next seven are dimensional flight-dynamics derivatives that are produced via accepted MATLAB post-processing. The final three are normalized aerodynamic root-load coefficients. With $q=\rho V^2/2$, the dimensional quantities can be recovered precisely as
\begin{equation}
L=qSC_L,
\qquad
D=qSC_D,
\qquad
F_{\mathrm{root}}=qS C_{\mathrm{shear}},
\label{eq:force_decoder_main}
\end{equation}
\begin{equation}
M_{\mathrm{bend}}=qS(2b)C_{\mathrm{bend}},
\qquad
M_{\mathrm{twist}}=qS\bar{c}C_{\mathrm{twist}}.
\label{eq:moment_decoder_main}
\end{equation}
Normalization and decoding make sure the network does not have to relearn the known dimensional scalings. Complete definitions and the conformal propagation needed to construct conservative bounds are detailed in Appendix~\ref{app:geometry_decoder}.

The derivatives $C_{m_\alpha}$, $C_{\ell_p}$, and $C_{m_q}$ are chosen as conservative sign conditions, whereas $N_\beta$ and $N_r$ are included in the directional stability--damping trade-off objective. The adoption of coefficient-to-derivatives, dimensionanlization, and sign treatment are described in Appendix~\ref{app:retained_dynamics_derivation}. However, the derivative $L_\beta$ is still predicted and reported only as a diagnosis as no time-domain validation for the lateral-directional dynamics has been done, whereas $N_\beta$ and $N_r$ are included in a convention-qualified directional-deficit objective.

\subsection{Vortex-lattice data generation}
\label{sec:vlm_data_generation}

The aerodynamic database was generated using the Tornado MATLAB vortex-lattice solver which was developed for linear lifting surface problems \cite{melin2000vortex}. In this case, for the specified wing geometry and operating conditions, Tornado represents the lifting planform by a lattice of horseshoe vortices and solves for the circulation strengths that satisfy the linearized flow-tangency. The resulting circulation distribution field provides spanwise load distributions, integrated force and moment coefficients, induced-like drag, root loads, and coefficients slopes needed for downstream fixed-wing post-processing \cite{leifsson2014fast}.

Tornado is a fully three-dimensional method in the lifting-surface sense, however, it is not a three-dimensional volumetric CFD solver. The geometry of the finite wing, which is characterized by span, taper, sweep, twist, and dihedral, exists in a three-dimensional space, and each vortex element induces three-component velocities at any point on the lifting planform. Aerodynamic variables, however, are defined on the wing lifting surface rather than the fluid volume surrounding it. Therefore, Tornado resolves three-dimensional finite-wing circulation, downwash, spanwise lift distribution, tip-vortex effect, forces and moments, and aerodynamic derivatives without performing volumetric meshing or solving the Navier-Stokes equations \cite{melin2000vortex,katzPlotkin2001,drelaYoungrenAVL}.

Given the geometry of the present glider, flight-speed interval, and reference altitude, the corresponding Mach and mean-aerodynamic-chord Reynolds numbers indicate low-subsonic flight at the full-scale glider dimensions. The conditions are hence, favorable for the incompressible component of the VLM approximation due to negligible compressibility effects. Moreover, the moderate sweep and specified angle-of-attack intervals are consistent with the intended linear and attached-flow application of Tornado \cite{melin2000vortex,drelaYoungrenAVL}. It should be noted that attachment is an assumption of the aerodynamic model and not a feature provided by the VLM solution itself. The cases, which would have significant laminar separation, trailing-edge separation, or stall may possess larger model-form errors than those indicated by the surrogate residuals.

The appropriateness of the solver must therefore be assessed relative to the purpose of the study. The present work investigates a surrogate, conformal-calibration, support-control, and robust multi-point optimization methodology over a broad conceptual planform domain. For this purpose, a deterministic and computationally efficient lifting-surface model is preferable to a small and sparsely distributed set of viscous simulations. The selected geometries should nonetheless be interpreted as candidates for subsequent higher-fidelity analysis rather than as aerodynamically certified glider wings.

\begin{figure}[t]
\centering
\includegraphics[width=0.78\textwidth]{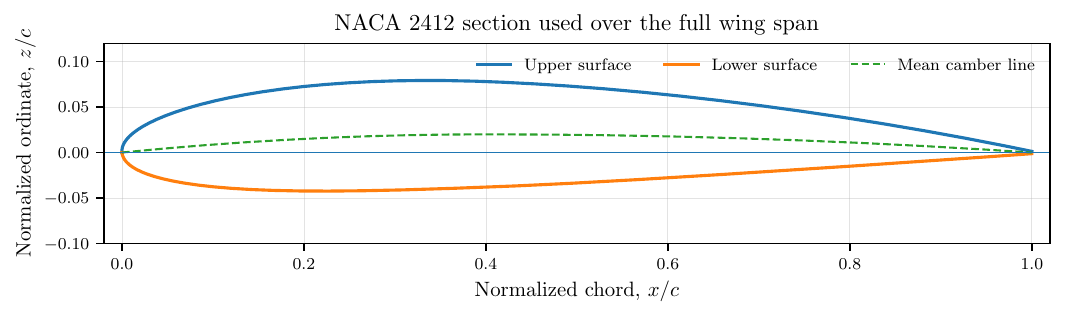}
\caption{NACA~2412 section used over the complete wing span. Coordinates are normalized by the local chord.}
\label{fig:naca2412}
\end{figure}

In total, $150{,}000$ configurations were examined from the continuous box defined in Table~\ref{tab:design_bounds}. The NACA~2412 profile was used over the entire span as shown in Fig.~\ref{fig:naca2412}, and the lattice contained 20 chordwise by 40 spanwise panels. Simulations were performed at a fixed altitude of $1200~\mathrm{m}$ under International Standard Atmosphere (ISA) setting.

Table~\ref{tab:vlm_configuration} summarizes the solver configuration.

\begin{table}[t]
\centering
\caption{Configuration used for the $150{,}000$ Tornado evaluations.}
\label{tab:vlm_configuration}
\begin{tabular}{ll}
\toprule
Item & Setting \\
\midrule
Aerodynamic solver & Tornado vortex-lattice method \\
Flow model & Steady, incompressible, inviscid potential flow \\
Wing section & NACA~2412, fixed over the span \\
Lattice resolution & $20\times40$ chordwise--spanwise panels \\
Atmospheric setting & $1200~\mathrm{m}$ altitude \\
Air density & $1.089960964052~\mathrm{kg\,m^{-3}}$ \\
Raw design/operating variables & 8 \\
Evaluated configurations & $150{,}000$ \\
\bottomrule
\end{tabular}
\end{table}

The number of simulations was chosen so that there would be a large coverage of the eight-dimensional geometry-operating-condition space, and sufficient statistical separation for the developed workflow. With 16 retained continuous responses per each of the $150{,}000$ configurations, the database includes $2.4 \times 10^{6}$ retained target values. Given the $20 \times 40$ lattice convention, each of these instances contains 800 unknowns for panel circulation, which means that there are nominally $1.20 \times 10^{8}$ panel-level unknown circulations that have been computed throughout the entire simulation. The values above are not equivalent to a single $1.20 \times 10^{8}$ unknown system or volumetric CFD mesh cells.

The database is also statistically useful apart from its aerodynamic purposes. The database allows for separate training, validation, conformal calibration, in-distribution testing, hidden candidate, and structured out-of-distribution datasets while still providing sufficient coverage in each dataset. The database also allows for a large design space exploration that does not require repeated querying of sparsely sampled spaces. More VLM cases lead to less sampling and more characterization of the VLM response surfaces; however, this does not eliminate the assumptions in the VLM model. Database volume and physical fidelity are thus treated as separate effects of the study.

Converged 3D viscous CFD is impractical from a computational point of view even if the costs of setting up the geometry and generating the volume mesh are disregarded. The exact cost ratio will vary depending on the flow solver, turbulence/transition modeling, mesh density, convergence tolerance, hardware, and initial conditions of the simulation.

Therefore, the computational argument leads to a preference for a hierarchy of fidelity options, with VLM being an important part of it, allowing the creation of a large database to train the surrogate.

The Tornado output table included aerodynamic coefficients, coefficient derivatives, dimensional forces and moments, summary of the spanwise loads, and root-load quantities. Next, a separate post-processing stage in MATLAB implemented the coefficient derived from VLM, combined with a fixed configuration of the fuselage and the tail surfaces to form dimensional flight-dynamics derivatives. The wing geometry was varied in accordance with Table~\ref{tab:design_bounds}, while the geometry of the fuselage and tail surfaces was fixed in order to study the effect of the wing planform only. The fixed configuration and the coefficient-to-derivative mapping are provided in Appendix~\ref{app:solver_configuration}. This post-processing step is a deterministic map from the VLM coefficient slopes and the reference geometry of the fixed-wing aircraft \cite{MathWorksAeroFixedWing,MathWorksForcesAndMoments,
MathWorksStaticStability,roskam1995flightdynamics}.

\subsection{target construction and physical decoding}
\label{sec:target_construction}

\begin{figure}[t]
\centering
\includegraphics[width=\textwidth]{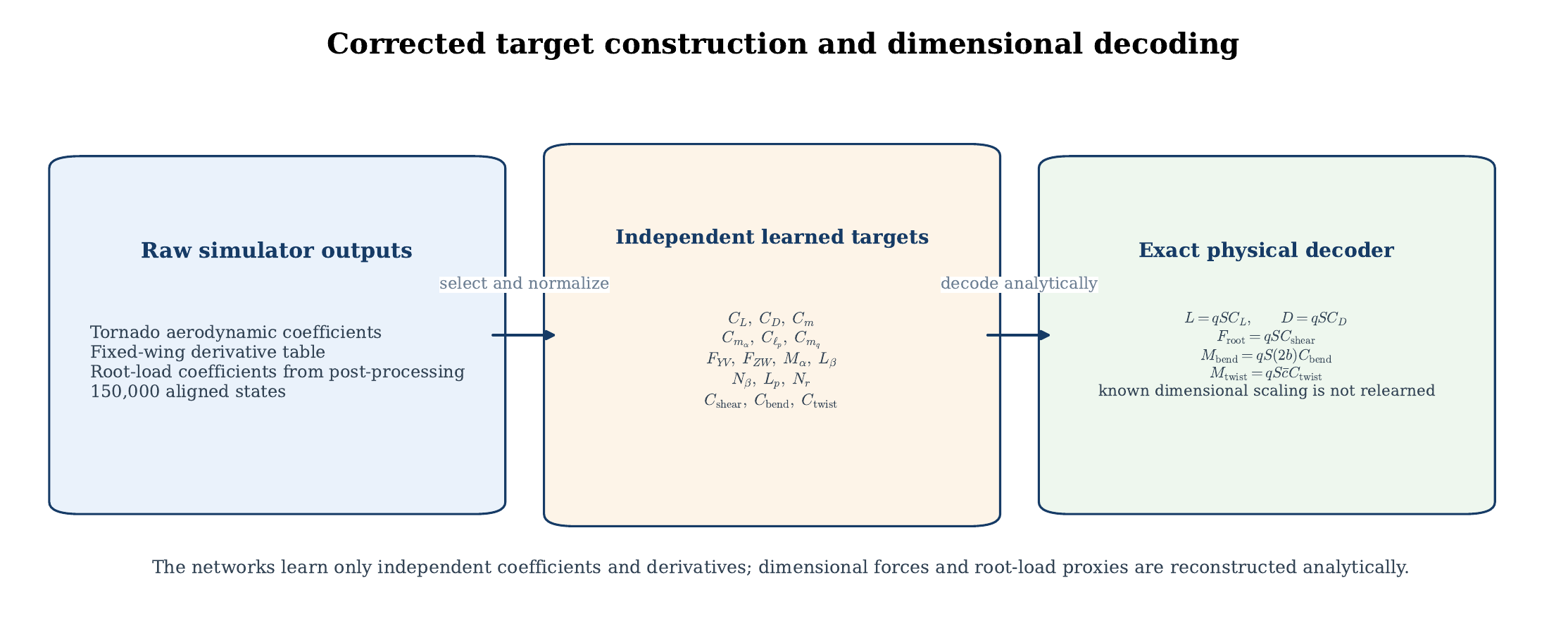}
\caption{Construction of the target vector and dimensional decoding. The ensemble learns 16 independent coefficients and derivatives; known dynamic pressure, area, span, and chord scaling allow reconstruction of dimensional forces and root loads.}
\label{fig:target_decoder}
\end{figure}

The raw simulation outputs were first checked for identification, dimensional consistency, redundancy, and dependency inspection prior to model training. The input table, aerodynamic-load table, and flight-dynamics table shared the same $150{,}000$ identifiers with no missing or non-finite entries. Deterministic dimensional values and redundant or affine-equivalent columns were not considered independent targets for learning. Rather, the accepted model learns the concise set of 16 continuous entities shown in Eq.~\eqref{eq:learned_outputs}, and physical dependencies of force and root load components are obtained from Eqs.~\eqref{eq:force_decoder_main} and~\eqref{eq:moment_decoder_main}. Refer to Fig.~\ref{fig:target_decoder}.

The learned outputs are organized in Table~\ref{tab:target_groups}. The aerodynamic head keeps the primitive coefficients necessary for trim, drag estimation, and moments, as well as the three sign constraints that are checked. Seven continuous outputs from the fixed-wing post-processing keep the magnitudes of the lateral-force, vertical-force, pitch, roll, and yaw derivatives. Lastly, root shear, bending, and torsion are nondimensionalized by the known scaling factors. In addition, the continuous regression keeps more information and allows uncertainty-aware margins instead of discontinuous class decisions.

\begin{table}[t]
\centering
\caption{Groups of continuous outputs learned by the surrogate.}
\label{tab:target_groups}
\begin{tabular}{
>{\raggedright\arraybackslash}p{0.20\textwidth}
>{\raggedright\arraybackslash}p{0.35\textwidth}
>{\raggedright\arraybackslash}p{0.34\textwidth}}
\toprule
Group & Learned quantities & Main role \\
\midrule
Aerodynamic primitives & $C_L$, $C_D$, $C_m$ & Lift trim, induced-like drag, pitch-moment diagnostic \\
Nondimensional stability and damping & $C_{m_\alpha}$, $C_{\ell_p}$, $C_{m_q}$ & Conservative sign constraints \\
Dimensional flight dynamics & $F_{YV}$, $F_{ZW}$, $M_\alpha$, $L_\beta$, $N_\beta$, $L_p$, $N_r$ & Continuous response margins and directional objective \\
Normalized root loads & $C_{\mathrm{shear}}$, $C_{\mathrm{bend}}$, $C_{\mathrm{twist}}$ & Load limits and root-bending objective \\
\bottomrule
\end{tabular}
\end{table}

Complete chain-rule derivation for the seven-dimensional derivatives is provided in Appendix~\ref{app:retained_dynamics_derivation}, which also includes native units as well as speed dependency. The difference between the wing-level coefficient primitives and the total fixed airframe derivatives that arise after including fixed fuselage and tail is also made clear in the derivation.

Lineage review of the dimensional dynamics shows two velocity scaling families in the accepted table. The velocity-scaled variables $F_{YV}$, $F_{ZW}$, $L_p$, and $N_r$ involve a factor proportional to $V$, while the factors involved in the variables $M_\alpha$, $L_\beta$, and $N_\beta$ are proportional to $V^2$. Dimensional transformation, units, and velocity scaling of these variables are provided in Appendix~\ref{app:retained_dynamics_derivation}. The multiplicative factors in each case are always positive over the speed interval of interest, and hence the dimensionalization will only alter the magnitude and not the sign.

Each row of the decoder was matched to corresponding rows of the simulator outputs. In this case, $L=qSC_L$, $D=qSC_D$, and three identities of root loads match the saved dimensions within the limits of the floating-point precision. This verification holds significance as it prevents incoherency between the quantities used for the optimization of the surrogate and the same quantities reconstructed by Tornado. It reduces the number of independent outputs that the network must learn as well and applies the exact dependence on $V^2$, $S$, $2b$, and $\bar c$.

\subsection{Leakage-free data partitioning}
\label{sec:data_partitioning}

\begin{figure}[t]
\centering
\includegraphics[width=\textwidth]{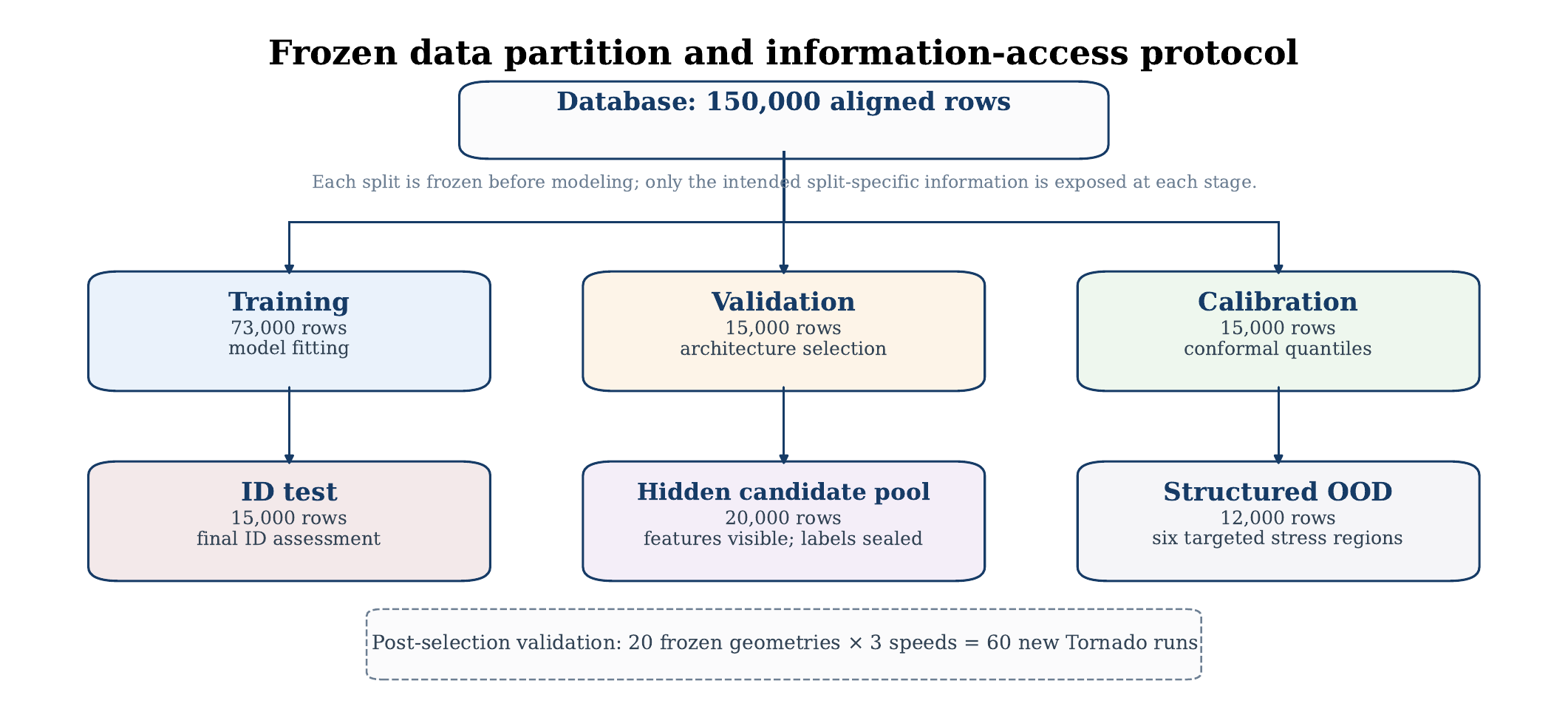}
\caption{Frozen partition protocol. Database rows used for training, validation, calibration, ID testing, hidden candidate selection, and structured-OOD stress testing are separate, and sixty post-selection simulations are independently generated.}
\label{fig:partition_protocol}
\end{figure}

The database was split into six separate partitions once, each with its own unique role. The distribution of the database is detailed in Table~\ref{tab:frozen_splits}, while Figure~\ref{fig:partition_protocol} illustrates it graphically. The training set is used for parameter estimation only; the validation set is used for architecture/model selection; the calibration set is kept only for conformal quantiles; the ID test set remains unused until the model selection is completed; the hidden candidate pool is used for feature exposure without label disclosure during model selection, and the structured OOD test set is kept in reserve for stress testing. Sixty Tornado simulations for final external validation are not part of any of the six partitions.

\begin{table}[t]
\centering
\caption{Frozen data partitions. The percentages refer to the original $150{,}000$ rows of the database.}
\label{tab:frozen_splits}
\begin{tabular}{lrr>{\raggedright\arraybackslash}p{0.37\textwidth}}
\toprule
Partition & Rows & Fraction & Role \\
\midrule
Training & 73,000 & 48.67\% & Fit ensemble members and feature/target scalers \\
Validation & 15,000 & 10.00\% & Select architecture and training configuration \\
Calibration & 15,000 & 10.00\% & Estimate conformal quantiles only \\
ID test & 15,000 & 10.00\% & Final in-distribution performance assessment \\
Hidden candidate pool & 20,000 & 13.33\% & Feature-only optimization challenge; labels sealed \\
Structured OOD & 12,000 & 8.00\% & Six targeted boundary and extreme-region tests \\
\bottomrule
\end{tabular}
\end{table}

After the structured OOD rows had been cut off, the remaining ID partitions were graded using quartile bins of aspect ratio, angle of attack, and velocity. Therefore, their feature means and ranges are closely matched. The hidden candidate set was first chosen only based on these input-space bins before performing the ID test, calibration, validation, and training partitions. Such an order avoids any output-based selection of the hidden candidate set.

\begin{figure}[t]
\centering
\includegraphics[width=0.86\textwidth]{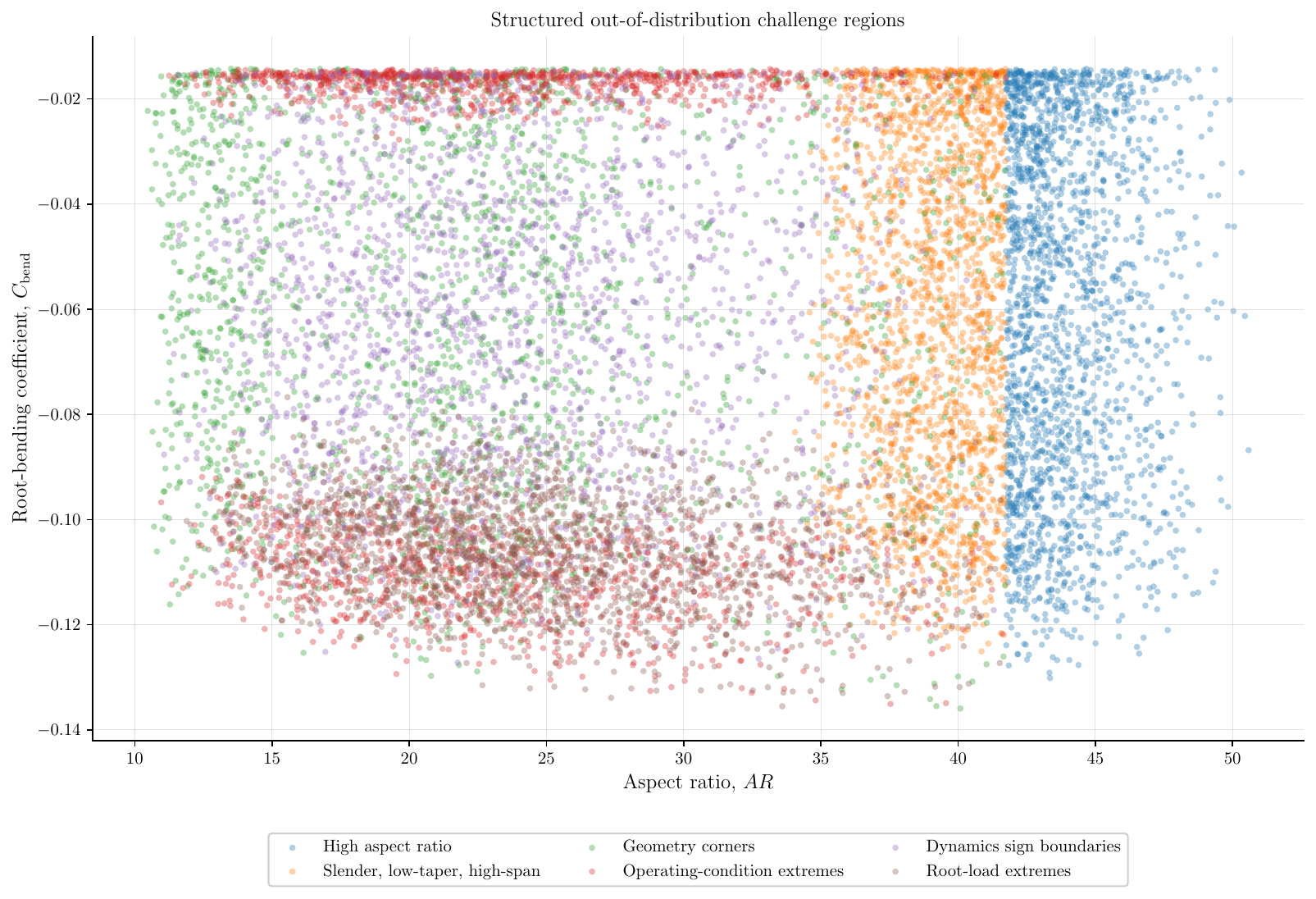}
\caption{The Structured OOD challenge regions were selected in the aspect ratio/root-bending space. All categories include 2,000 entries each and are independent from the ID partitions.}
\label{fig:ood_map}
\end{figure}

The structured OOD set consists of six disjoint categories of $2{,}000$ rows each as follows: highest aspect-ratio configurations; slender, low taper and high semi-span configurations; geometric corners of the six fixed geometry features; extreme combinations of angle of attack and velocity; configurations near the $F_{YV}$, $N_\beta$, or $N_r$ sign boundaries; and extreme normalized root load states. These categories have been constructed through robust standardized scores and create six mutually disjoint challenge sets. Figure~\ref{fig:ood_map} shows the distribution of these categories within aspect ratio and root-bending response.

The hidden label lock and the external validation lock are part of the statistical protocol. All candidate geometries and optimization hyperparameters were frozen prior to checking any further simulator outputs. Therefore, the external validation scores represent true generalization to unseen continuous geometries.

\subsection{Physics-structured ensemble surrogate}
\label{sec:physics_surrogate}

\subsubsection{Feature representations and dual-head architecture}
\label{sec:dual_head_architecture}

All feature and target transformations were learned on the training partition only. The 12-feature full vector is
\begin{equation}
\mathbf{x}_{\mathrm{full}}
=
\left[
 c_r,\alpha,\Lambda,b,\theta,\lambda,\Gamma,V,
 S,AR,\bar c,q
\right]^{\mathsf T},
\label{eq:full_feature_vector}
\end{equation}
whereas the aerodynamic similarity vector is limited to
\begin{equation}
\mathbf{x}_{\mathrm{sim}}
=
\left[
 AR,\lambda,\Lambda,\theta,\Gamma,\alpha
\right]^{\mathsf T}.
\label{eq:similarity_feature_vector}
\end{equation}
The restriction in Eq.~\eqref{eq:similarity_feature_vector} represents an inductive bias rather than being a claim that all the VLM predictions are invariant to scale. This is to make the aerodynamic and normalized load predictions basely dependent on planform similarity and attitude, with the dimensional flight-dynamics model having access to the whole geometry, velocity, dynamic pressure, area, and chord for representing the magnitude of the derivatives.

\begin{figure}[t]
\centering
\includegraphics[width=\textwidth]{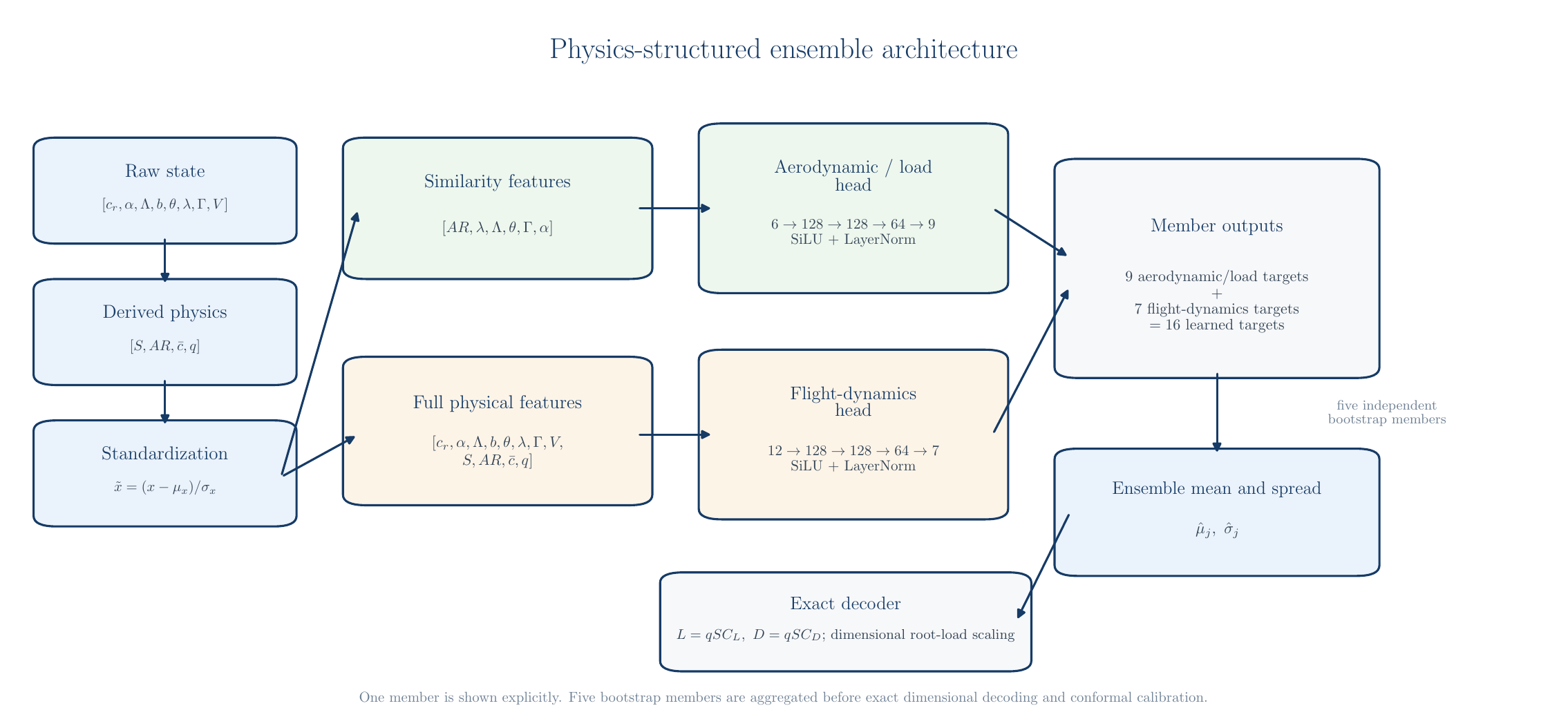}
\caption{Physics-structured dual-head ensemble. Each individual member maps similarity feature vectors to nine aerodynamic/load targets and the full physical feature vector to seven flight-dynamics targets, with five bootstrap models aggregated before conformal calibration and exact dimensional decoding.}
\label{fig:surrogate_architecture}
\end{figure}

The ensemble member contains two separate feed-forward heads. The aerodynamic/load head takes the six similarity features and maps them to nine predictions using the hidden layers with widths 128, 128, and 64: $C_L$, $C_D$, $C_m$, $C_{m_\alpha}$, $C_{\ell_p}$, $C_{m_q}$, $C_{\mathrm{shear}}$, $C_{\mathrm{bend}}$, and $C_{\mathrm{twist}}$. The flight-dynamics head has 12 full features, the same hidden widths, and seven outputs: $F_{YV}$, $F_{ZW}$, $M_\alpha$, $L_\beta$, $N_\beta$, $L_p$, and $N_r$. Both heads implement sigmoid-weighted linear unit activations and layer normalization after the first two affine layers \cite{elfwing2018silu,ba2016layernorm}. One dual-head member has 54,160 trainable parameters, while the five-member ensemble has 270,800 parameters. Figure~\ref{fig:surrogate_architecture} shows the complete architecture and the exact decoder applied after ensemble prediction.

The dual-head separation is intentionally small-scale. It does not involve differential equations, conservation residual, or the VLM operator within the neural network. Instead, it implements known differences between scale-normalized aerodynamic outputs and dimensional dynamic derivatives while maintaining enough flexibility to support nonlinear regression. The exact force and load relations stay outside of the mapping. Refer to Sec.~\ref{sec:target_construction} for details.

\subsubsection{Training protocol and ensemble diversity}
\label{sec:surrogate_training}

Training involved 5 independent members, 12 epochs, and a mini-batch size of 8192 (CUDA) or 4096 (CPU), as determined by the hardware. Both the inputs and all 16 outputs were standardized with training-partition statistics. AdamW optimization \cite{loshchilov2019adamw} was used with a learning rate of $2\times10^{-3}$ and decoupled weight decay of $10^{-5}$. The smooth $L_1$ loss with transition parameter $\delta=0.5$ was applied in standardized target space, limiting the effect of rare high residual errors while maintaining the quadratic loss function in a neighborhood near zero \cite{huber1964robust}.

For diversification purposes, online Poisson bootstrap weights were used. In other words, for each mini-batch and model member, each observation is assigned a weight sampled independently from $\operatorname{Poisson}(1)$. This is an efficient bootstrap approximation that complements independent initialization \cite{oza2001online}. The loss function first calculates the average of nine aerodynamic/load residuals and the average of seven dynamics residuals, followed by the averaging of five bootstrap-weighted member losses. A compact unrestricted multilayer perceptron was trained using the same routine but with separate parameters for black-box comparison. Detailed loss function, ensembles, and checkpoint criteria are provided in Appendix~\ref{app:surrogate_math}.

\begin{figure}[t]
\centering
\includegraphics[width=0.82\textwidth]{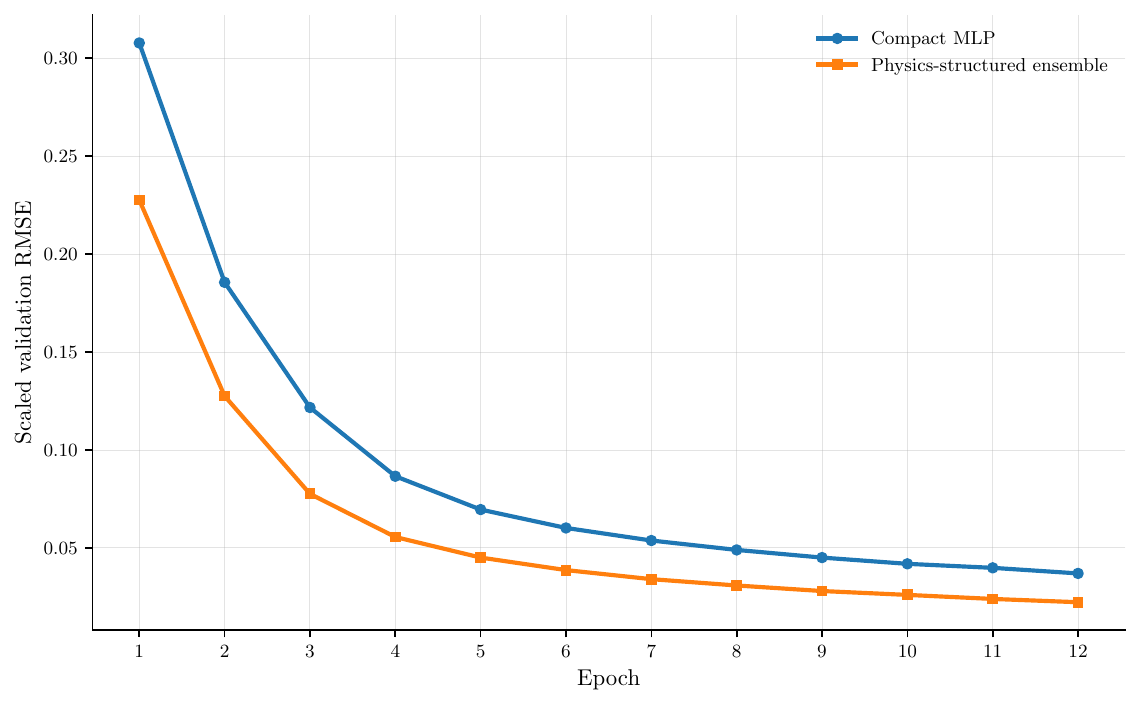}
\caption{Validation history in standardized target space. Epochs are indexed starting from one on the plot; the last checkpoint reached the scaled validation RMSE of 0.02230 for the physics-structured ensemble and 0.03700 for the compact MLP.}
\label{fig:training_history}
\end{figure}

The validation error decreased during the entire 12-epoch run for both neural networks, and the ensemble remained under the compact MLP (Fig.~\ref{fig:training_history}). The best checkpoints were chosen based on the scaled validation RMSE, instead of training loss. No calibration, ID-test, OOD-test, hidden label, or external simulation outputs were used for checkpoint selection.

\subsection{Baseline models and frozen selection rule}
\label{sec:model_selection}

Five predictors were inspected using the same frozen splits and targets. Ridge regression provides a Five predictors were tested using the same frozen splits and targets. Linear ridge regression gives a baseline reference using an $ \ ell_2$ penalty \cite{hoerl1970ridge}. A nonlinear tree-ensemble baseline was provided by LightGBM. LightGBM was trained separately for each target \cite{ke2017lightgbm}. The compact MLP predictor uses all 12 features and maps layers with widths of 128, 128, and 64 to all 16 outputs. The physics single-predictor model is a single fixed Poisson-bootstrap member of the dual-headed ensemble and represents the advantage of the averaging technique without needing to change the architecture. The final physics ensemble predicts all five members. The baseline hyperparameters are all listed in Appendix~\ref{app:surrogate_math}.

As for model selection, the validation partition alone was used, requiring the meeting of three criteria: low normalized regression error through all 16 targets, reliable sign recovery at low-magnitude derivative boundaries, and rank preservation according to aerodynamic optimization. For target $j$, the RMSE was normalized to the standard deviation of the training target. The criterion for boundary sign accuracy was evaluated over the lowest decile of absolute magnitude for the nine sign-sensitive derivatives. The rank criterion was measured using the mean Spearman correlation between $C_D$ and $|C_{\mathrm{bend}}|$. The composite score was fixed before the ID, and OOD results were evaluated,
\begin{equation}
\mathcal S
=
\overline{\mathrm{NRMSE}}
+0.15\left(1-A_{\mathrm{boundary}}\right)
+0.15\left(1-R_{\mathrm{rank}}\right),
\label{eq:model_selection_score_main}
\end{equation}
with lower values preferred. The detailed metric definitions appear in Appendix~\ref{app:surrogate_math}. The protocol also specified that the structured ensemble would be retained when its score was within 5\% of the lowest score, because subsequent uncertainty calibration requires member-level predictions. In the experiment, the ensemble itself achieved the lowest composite score, so the tolerance rule did not alter the numerical winner.

\subsection{Conformal calibration and simultaneous uncertainty sets}
\label{sec:conformal_method}

The ensemble containing five members produces a mean prediction and an observational member dispersion for each output; however, the dispersion is not explicitly a calibrated predictive standard deviation. A calibration set of 15,000 observations is thus created to transform the model disagreement to finite-sample prediction intervals. The use of split conformal prediction appears to be appealing here as it allows one to calibrate a previously trained regression without having to specify a likelihood \cite{lei2018distribution}. When there are multiple outputs, this study defines a single joint score based on the largest standardized residual such that the whole response vector is accounted for all at once and not target by target.

For target $j$, the calibration residual is scaled by the sum of the ensemble standard deviation and the target-specific floor,
\begin{equation}
 r_{ij}
 =
 \frac{|y_{ij}-\widehat\mu_j(\mathbf{x}_i)|}
 {\widehat\sigma_j(\mathbf{x}_i)+\epsilon_j}.
 \label{eq:normalized_conformal_score_main}
\end{equation}
The floor is the larger of the 10th percentile of calibration-set dispersion and 1\% of the training-target standard deviation. This ensures that nearly perfect member agreement does not result in unreliable scores. Four interval constructions were kept for comparison: the uncalibrated Gaussian ensemble interval; target-wise marginal normalized conformal intervals; a global simultaneous interval over the 14 variables entering optimization; and finally a support-conditioned Mondrian variant. The optimization certificate is the global 95\% interval,
\begin{equation}
 \left[
 \widehat\mu_j(\mathbf{x})
 -\widehat q_{0.95}\{\widehat\sigma_j(\mathbf{x})+\epsilon_j\},
 \widehat\mu_j(\mathbf{x})
 +\widehat q_{0.95}\{\widehat\sigma_j(\mathbf{x})+\epsilon_j\}
 \right],
 \qquad j\in\mathcal J_{\mathrm{opt}},
 \label{eq:global_conformal_interval_main}
\end{equation}
where $\widehat q_{0.95}=3.33918$ is the finite-sample order statistic of the maximum normalized calibration residual. The 14 certified outputs are $C_L$, $C_D$, $C_{m_\alpha}$, $C_{\ell_p}$, $C_{m_q}$, $F_{YV}$, $F_{ZW}$, $M_\alpha$, $N_\beta$, $L_p$, $N_r$, and the three root-load coefficients. Diagnostic outputs $C_m$ and $L_\beta$ are predicted and calibrated but are not included in the hard optimization certificate. The construction of the maximum score, the quantile of finite-samples, and the coverage assertion are described in more detail in Appendix~\ref{app:conformal_math}.

The conformal region in Eq.~\eqref{eq:global_conformal_interval_main} is a rectangle. This is conventional when compared to a fully learned multivariate conformal set, but it is transparent and allows exact component-wise propagation via the force and load decoder. This formulation is closely related to robust optimization, where decisions are evaluated at the worst-case point in the calibrated uncertainty set \cite{johnstone2021conformal}. The coverage guarantee is marginal under exchangeability, meaning it makes no claims about exact conditional coverage at each geometric configuration, nor is it by itself a guarantee of coverage for an adaptive design-selection procedure.

\subsection{Support-aware calibration and extrapolation control}
\label{sec:support_method}

The geometric support score is used to separate interpolation-like candidates from those that are distant from the training cloud. All 12 physical features are normalized using the training statistics, and the score $d_{20}(\mathbf{x})$ represents the average Euclidean distance to the 20 closest training points. Thresholds of the calibration score at the 50th, 80th, and 95th percentiles are 1.19541, 1.28303, and 1.38104, respectively. The four support bins are defined by the said thresholds. A separate max-score conformal quantile is predicted for each support bin according to the general Mondrian principle of calibrating in pre-defined difficulty categories \cite{bostrom2020mondrian}.

At the 95\% nominal coverage level, the support bin quantiles grow monotonically from 2.8523 in the best-supported bin to 5.9625 in the most distant bin. This is both a physical and statistical desired property: the confidence interval grows wider when a candidate's distance to the observed training distribution increases. The support-Mondrian intervals are implemented for OOD diagnosis and an exploratory alternative, whereas the global joint interval serves as a formal certificate for the optimization process. Furthermore, any optimized operating point must satisfy
\begin{equation}
 d_{20}(\mathbf{x})\leq 1.38104,
 \label{eq:support_constraint_main}
\end{equation}
which is the 95th-percentile threshold of the calibration set. The support constraint ensures the limitation of extrapolation until the robust objective and derivative constraints apply.

\subsection{Support-aware robust multipoint optimization}
\label{sec:robust_optimization}

The calibrated ensemble was applied in order to perform a continuous search in a six-variable geometry domain for three flight regimes. The frozen mission weight was $W=5886$~N, and three freestream velocities were $V_k \in \{38, 45, 52\}$~m\,s$^{-1}$, with mission weights $w_k \in \{0.25, 0.50, 0.25\}$. For each geometry, an angle of attack was determined at each speed within $-2.5^\circ \leq \alpha \leq 7.5^\circ$. Fifteen vectorized bisection iterations were performed on the mean lift response of the ensemble. The mean lift was used only to identify the trim-state candidate, while all subsequent acceptance decisions and objectives were evaluated with the 95\% global joint conformal bounds.

The scrambling of the Sobol sequence resulted in the creation of $2^{14}=16{,}384$ geometries within the frozen bounds of the training domain \cite{sobol1967distribution}. A geometry could only be regarded as robustly feasible if three operating points met five sets of criteria: (i) the full conformal lift bound remained within $\pm5\%$ of $W$; (ii) the upper bounds for $C_{m_\alpha}$, $C_{\ell_p}$, and $C_{m_q}$ were negative; (iii) the nearest-neighbor support score remained below 1.38104; and (iv--v) the worst case of the nondimensional bounds for the root-shear and root-twist coefficients was below the 95th percentile threshold from the training set, 0.465921 and 0.00220706, respectively. Root bending was retained as an objective rather than a hard limit.

The remaining designs were evaluated against the following three conservative objectives: mission-weighted drag, worst root-bending proxy, and a normalized directional stability-damping deficit. In the first two cases, decoded upper bounds are implemented. For the latter, the lower bound of $N_\beta$ that goes below zero or the upper bound of $N_r$ that goes above zero is penalized, after being normalized by the median absolute magnitudes, $s_{N_\beta}=331.5103$ and $s_{N_r}=44.1025$, observed in the training set. These numerical signs are consistent with the adopted post-processing convention, which is described in Appendix~\ref{app:retained_dynamics_derivation}. The directional quantities were not forced as simultaneous hard signs, as only 0.6356\% of the training rows satisfy both sign conditions, and there was no hidden candidate that satisfied both signs at 95\%.

Nondominated sorting was then performed for the three objective values. Twenty candidates were frozen prior to the external solver evaluation. This selection includes the minimum-drag design, the minimum-bending design, the design with the minimum directional deficit, the design closest to the normalized three-objective ideal (the balanced knee), and some additional designs chosen by farthest-point sampling in the normalized objective space. Mathematical details of this optimization step are provided in Appendix~\ref{app:optimization_math}.

\FloatBarrier
\section{Results}
\label{sec:results}

\subsection{Surrogate benchmark and model selection}
\label{sec:surrogate_results}

The frozen validation, ID test, and structured-OOD benchmark are presented in Table~\ref{tab:model_benchmark}. The physics-structured ensemble achieved an average NRMSE of 0.02181 in the validation set, 0.02234 in the ID test set, and 0.05951 in the structured-OOD test set. In comparison to the compact MLP, the average NRMSE was improved by 39.5\% in the ID test and 22.4\% in structured OOD test cases. Compared to LightGBM, the reductions are 32.8\% and 31.2\%, respectively. The mean NRMSE reductions for the ID test and OOD test were 17.0\% and 9.2\%, respectively, compared to a single dual-head member.

\begin{table}[t]
\centering
\caption{Surrogate benchmark. NRMSE is averaged through all 16 outputs normalized by the training target standard deviation. ``Boundary" is the near-boundary sign accuracy, ``Rank" is the aerodynamics ranking statistic, and ``Score'' is the validation compound in Eq.~\eqref{eq:model_selection_score_main}. Lower NRMSE and score are better.}
\label{tab:model_benchmark}
\scriptsize
\setlength{\tabcolsep}{2.0pt}
\begin{tabular}{lrrrrrr}
\toprule
Model & Val. & Boundary & Rank & Score & ID & OOD \\
\midrule
\textbf{Physics ensemble} & \textbf{0.02181} & \textbf{0.93250} & 0.99976 & \textbf{0.03197} & \textbf{0.02234} & \textbf{0.05951} \\
Physics single & 0.02651 & 0.92377 & 0.99962 & 0.03801 & 0.02692 & 0.06555 \\
LightGBM & 0.03323 & 0.91098 & \textbf{0.99978} & 0.04661 & 0.03326 & 0.08651 \\
Compact MLP & 0.03644 & 0.89812 & 0.99903 & 0.05187 & 0.03690 & 0.07670 \\
Ridge & 0.24585 & 0.80464 & 0.99656 & 0.27567 & 0.24662 & 0.35830 \\
\bottomrule
\end{tabular}
\end{table}

\begin{figure}[t]
\centering
\includegraphics[width=0.88\textwidth]{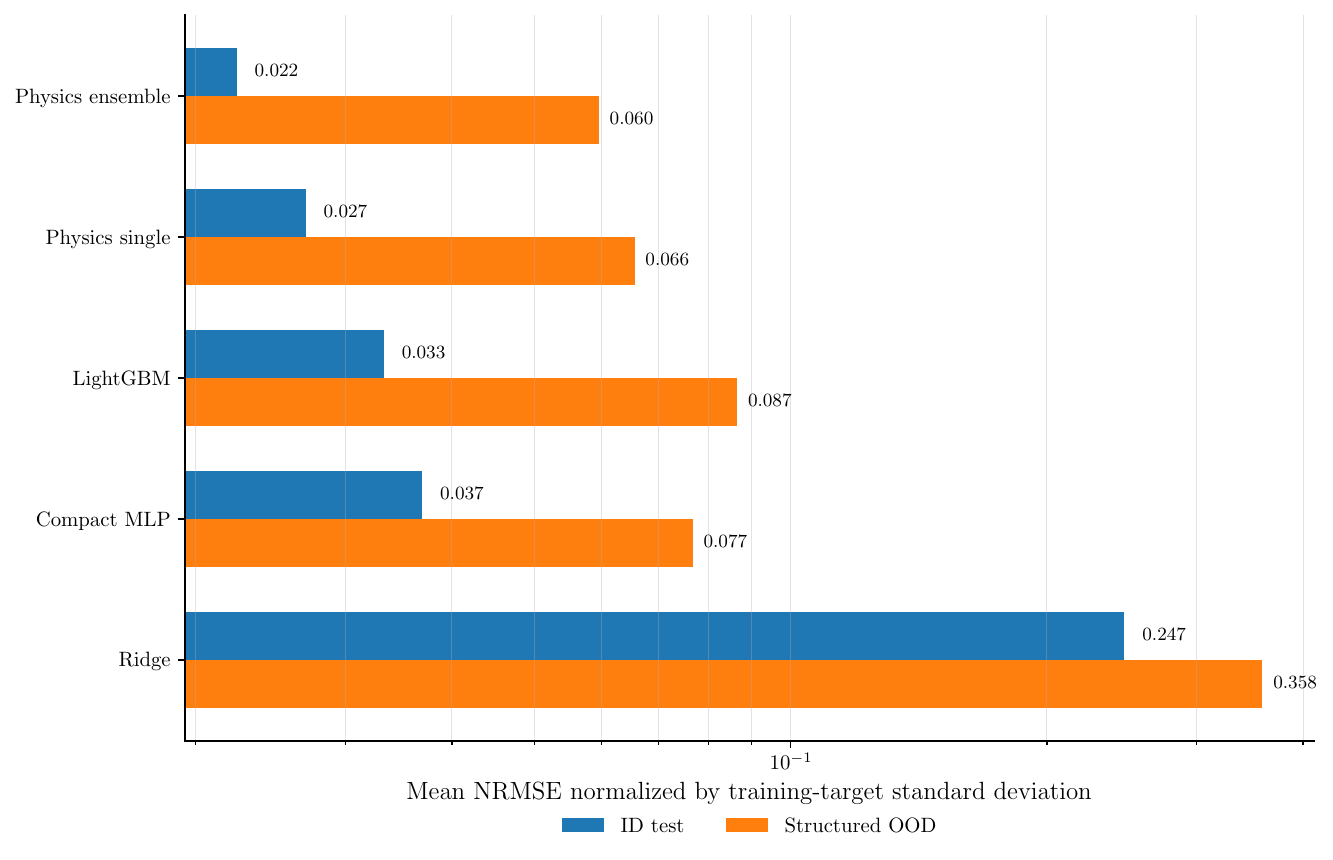}
\caption{Mean NRMSE on the ID and structured OOD tests for the five frozen model families. The x-axis is logarithmic due to significantly lower accuracy of the ridge baseline compared to nonlinear models.}
\label{fig:model_benchmark}
\end{figure}

The comparison in Fig.~\ref{fig:model_benchmark} emphasizes two points. On one hand, the structured ensemble outperforms the unrestricted neural baseline despite equal hidden layer widths and identical training objectives in both cases. On the other hand, each model is degraded under the structured OOD scenario; however, the degradation is least severe in terms of absolute values in the case of the selected ensemble. The ensemble achieved $R^2>0.9990$ in all ID test cases and had ID NRMSE below 0.031 for every target. The largest ID normalized errors were observed for $N_r$, root shear, root bending, and $N_\beta$; although these values are relatively small compared to training variances, they still necessitate uncertainty calibration instead of point prediction application.

\begin{figure}[t]
\centering
\includegraphics[width=\textwidth]{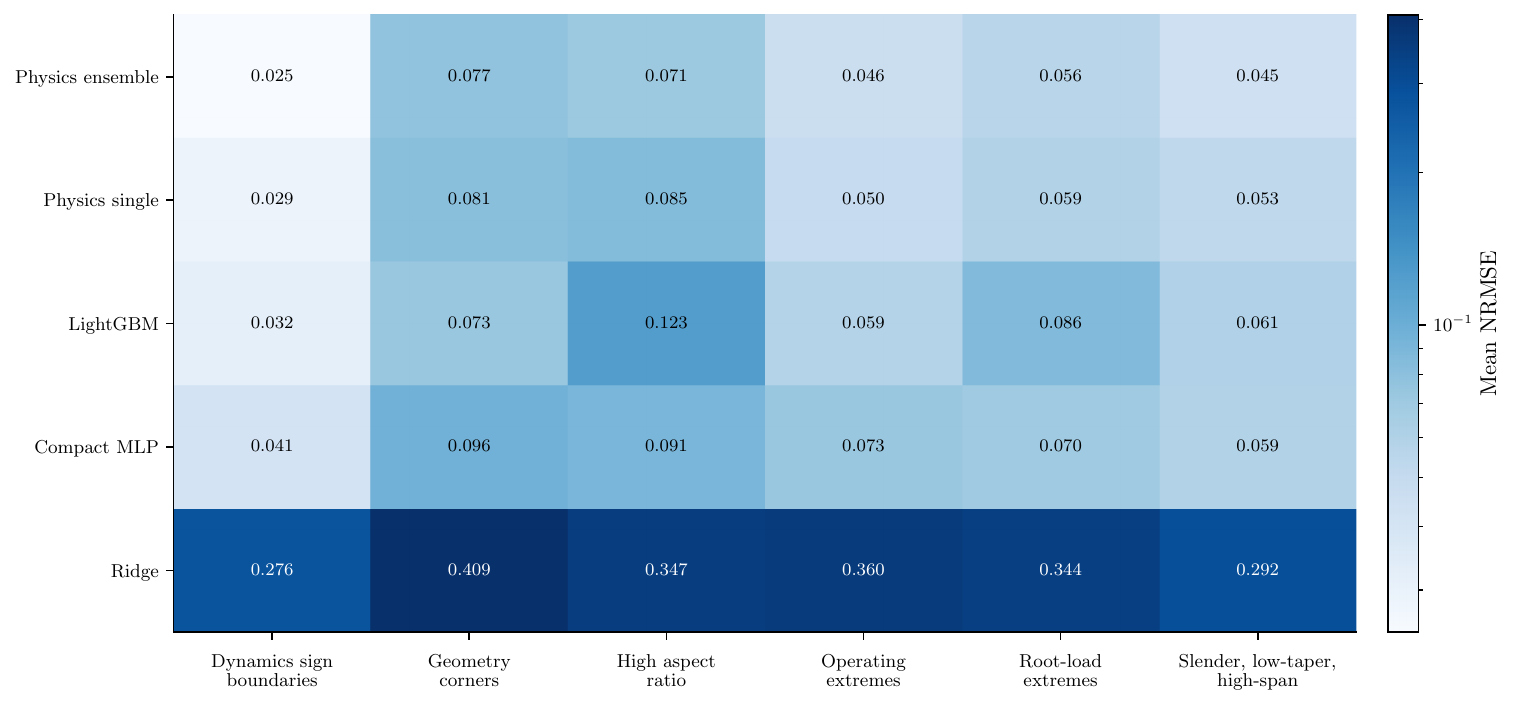}
\caption{Mean NRMSE for each structured-OOD category. Values are annotated and color intensity is based on a logarithmic scale to maintain contrast between the nonlinear models and the ridge reference.}
\label{fig:ood_heatmap}
\end{figure}

The ensemble performed best with the lowest average NRMSE values in five out of the six OOD categories (Fig.~\ref{fig:ood_heatmap}). LightGBM performed slightly better with the isolated geometry-corner OOD category (0.0731 vs 0.0773), but the ensemble was significantly better with respect to high aspect ratio geometries (0.0712 vs 0.1230), root-load extremes (0.0561 vs 0.0855), operating extremes (0.0458 vs 0.0588), slender and high-span designs (0.0447 vs 0.0607), and finally dynamics sign boundaries(0.0248 vs 0.0318). These findings do not provide evidence of superiority of the proposed representation across all domains. The findings only indicate the inductive splitting strategy works well on these six challenge regions.

Ranking performance of the model also provides an explanation for why regression accuracy on its own is not enough for downstream optimization. The best rank-quality statistic for validation was provided by LightGBM with only a slight margin, yet its mean OOD error was 45.4\% larger than that of the physics ensemble. In turn, the physics ensemble demonstrated the smallest regression error along with the highest ranking fidelity and the best average near-boundary sign accuracy. Hence, it was frozen and became the only model passing the conformal calibration. There was also only a weak correlation between the absolute ID error (0.067--0.304), confirming that the ensemble spread indeed required calibration before it could be interpreted as an uncertainty interval.

\subsection{Conformal calibration and support-aware coverage}
\label{sec:conformal_results}

Table~\ref{tab:conformal_coverage} summarizes how the four uncertainty constructions compare to each other at 95\% nominal coverage. For the uncalibrated ensemble interval, the joint coverage on ID was 17.51\%, while for the structured OOD data it was even smaller at 11.68\%. Even after conformal calibration on a target-by-target basis, though the marginal coverage improved, the issue of simultaneous coverage remained, with coverage of 58.29\% on ID and 27.81\% on OOD. In comparison, the max-score interval had 94.71\% joint coverage on the ID data, being quite close to the nominal level. Its joint coverage on OOD was 73.81\%, which was expected as the structured OOD data were selected from boundary and extreme regions rather than being sampled from the calibration distribution.

\begin{table}[t]
\centering
\caption{Coverage and mean normalized interval width at 95\% nominal coverage. Convergence is evaluated throughout all 14 outputs. Support-Mondrian result serves as a diagnosis; the global joint interval is the formal optimization certificate.}
\label{tab:conformal_coverage}
\scriptsize
\setlength{\tabcolsep}{3.2pt}
\begin{tabular}{lrrrr}
\toprule
Method & ID coverage & OOD coverage & ID width & OOD width \\
\midrule
Raw ensemble & 0.1751 & 0.1168 & 0.0625 & 0.0967 \\
Marginal normalized & 0.5829 & 0.2781 & 0.0817 & 0.1093 \\
\textbf{Global joint} & \textbf{0.9471} & 0.7381 & 0.1733 & 0.2316 \\
Support-Mondrian & 0.9497 & \textbf{0.9108} & 0.1763 & 0.3599 \\
\bottomrule
\end{tabular}
\end{table}

\begin{figure}[t]
\centering
\includegraphics[width=0.90\textwidth]{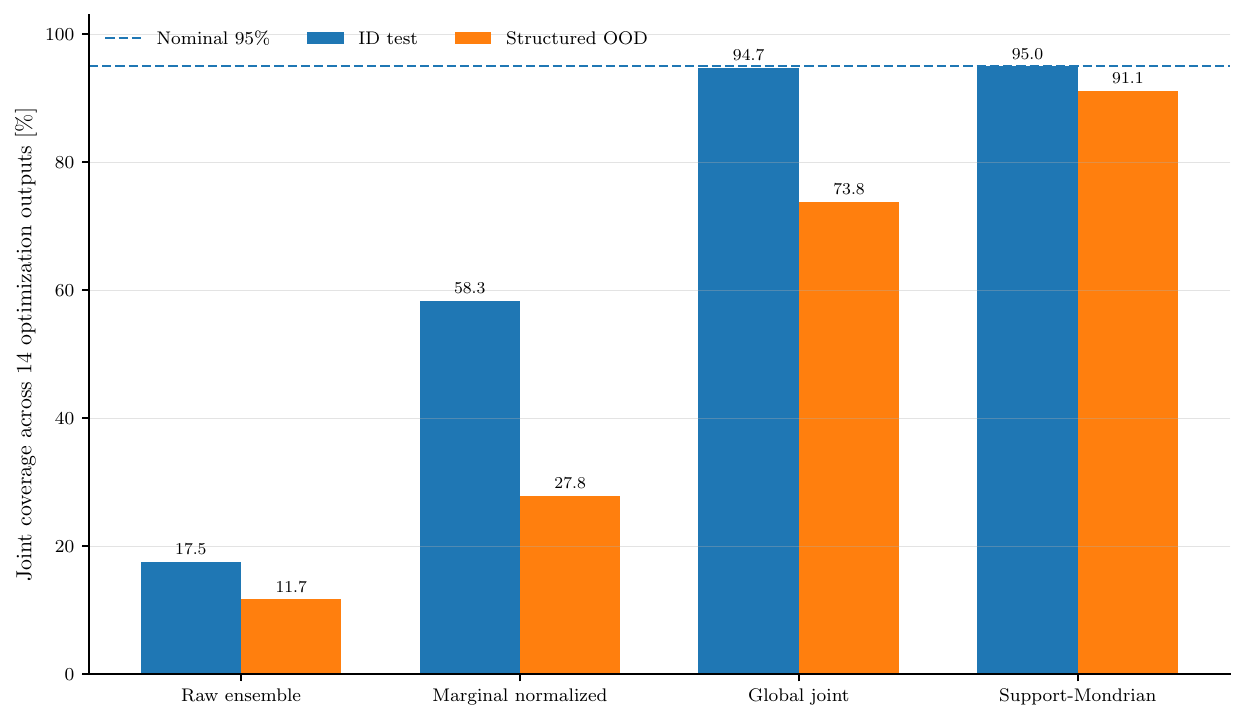}
\caption{Joint coverage of the 14 optimization outputs at 95\% nominal coverage level. Marginal target-wise calibration fails to provide a simultaneous guarantee, while the max-score global interval satisfies the nominal level on the ID test.}
\label{fig:conformal_joint_coverage}
\end{figure}

\begin{figure}[t]
\centering
\includegraphics[width=0.82\textwidth]{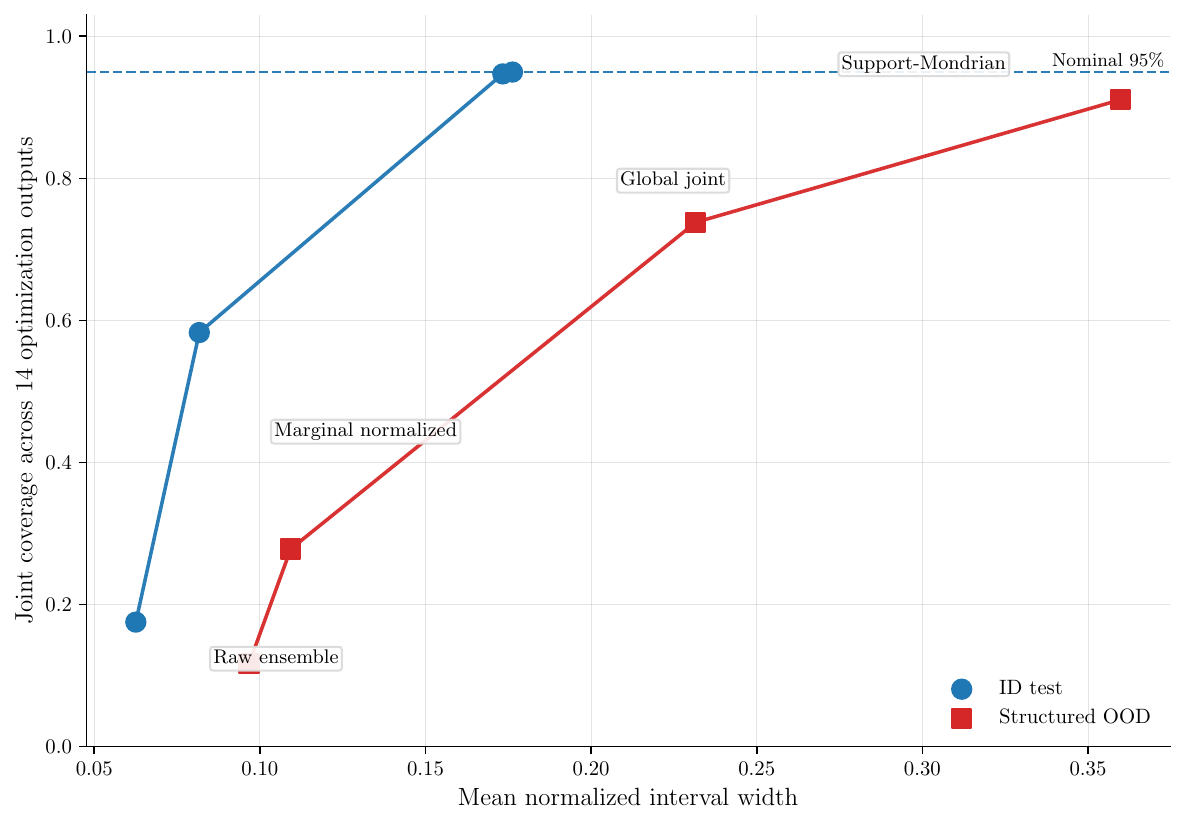}
\caption{Joint coverage versus mean normalized interval width at 95\% nominal coverage. Support conditioning improves structured-OOD coverage at the cost of wider intervals in low-support regions.}
\label{fig:coverage_width_tradeoff}
\end{figure}

Figure~\ref{fig:conformal_joint_coverage} presents the corresponding coverage comparison, while Figure~\ref{fig:coverage_width_tradeoff} puts into perspective the cost of OOD adaptation. The support condition improved OOD joint coverage compared to the global interval by 17.28 percentage points, increasing joint coverage from 73.81\% to 91.08\%, yet increased the mean normalized OOD width by 55.4\%. The same conditioning managed to improve the ID test joint coverage by only 0.26 percentage points, while confidence widths expanded by 1.7\%. These results confirm the division of roles between the global interval, which acts as a reliable formal certificate, and support-Mondrian calibration, which is required in sparse regions.

\begin{figure}[t]
\centering
\includegraphics[width=0.84\textwidth]{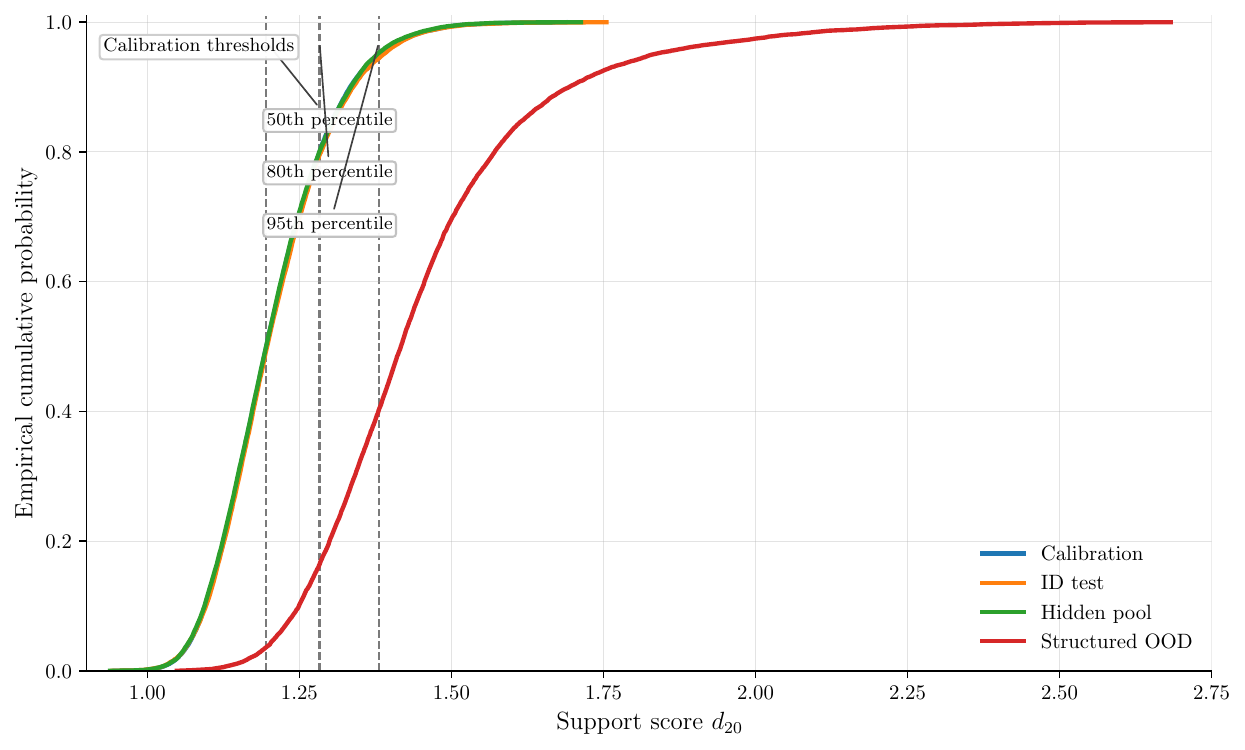}
\caption{Empirical cumulative distributions of 20 nearest-neighbors support scores. Dashed lines are the calibration 50th, 80th, and 95th percentiles used to define the four support bins.}
\label{fig:support_ecdf}
\end{figure}

Explanation of OOD gap from the support-score distributions is depicted in Fig.~\ref{fig:support_ecdf}. Calibration and ID curves are almost overlapping, and the hidden candidate pool is concentrated in the same range. However, out of the total number of OOD samples equal to 12,000, 7,163 are in the distant support bin, while OOD support score reaches up to 2.683, compared to a calibration maximum of 1.687.

\begin{figure}[t]
\centering
\includegraphics[width=0.94\textwidth]{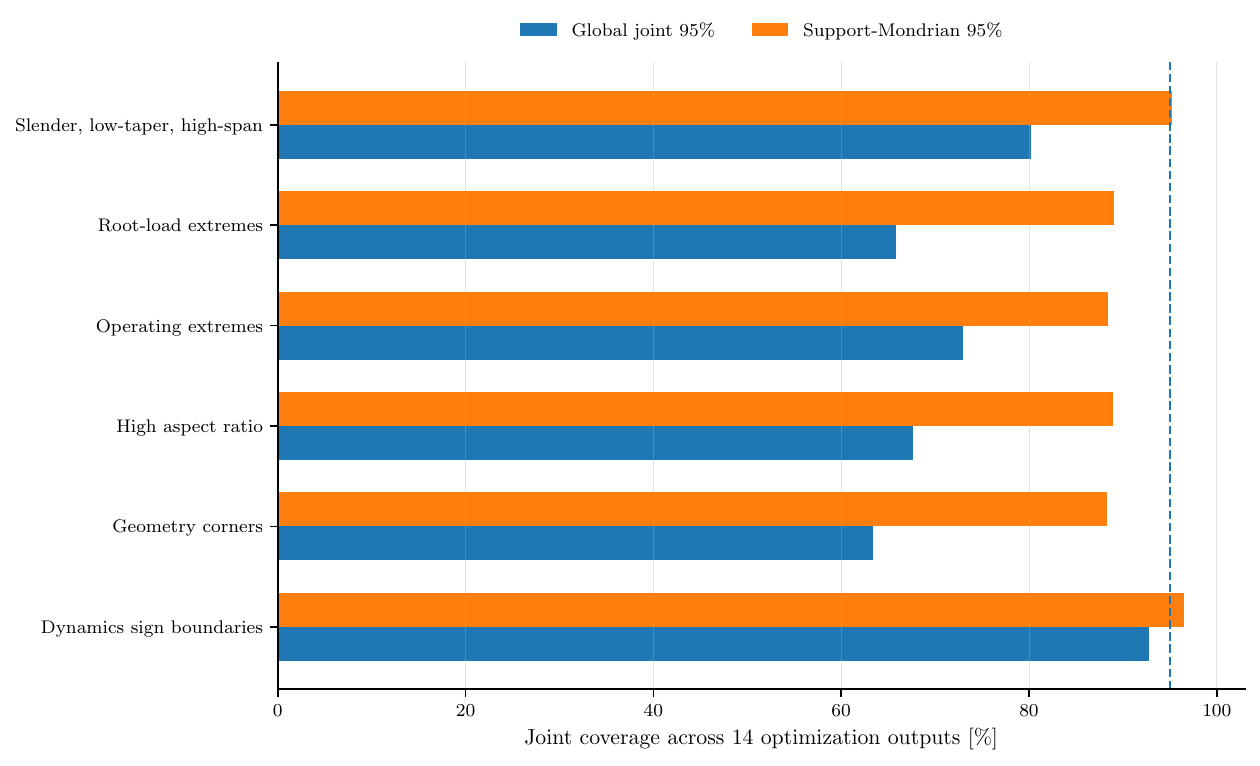}
\caption{Structured-OOD joint coverage by challenge category. Support-conditioned calibration improves all categories, and the largest expansion of the intervals occurs for the sparsest high-aspect ratio and geometry corners regions.}
\label{fig:ood_category_conformal}
\end{figure}

Category-wise results in Fig.~\ref{fig:ood_category_conformal} demonstrate that support conditioning does improve each structured-OOD category. Categories with the largest global-interval deficits include geometry corners, high-aspect ratio wings, operating extremes, and root-load extremes. The support-Mondrian interval brings all six categories above 88\% and achieves 96.55\% for dynamics sign boundaries and 95.20\% for slender, low-taper ratio, high-span wings. This is an empirical diagnostic result; the adaptive interval does not substitute the formal global certificate in optimization.

\subsection{Robust multipoint screening and Pareto structure}
\label{sec:optimization_results}

The continuous Sobol search generated 16,384 geometries. The independent number of passes for each group of constraints and the number of passes for simultaneous intersections are recorded in Table~\ref{tab:constraint_screening}. The robust lift condition was the most constraining individual screen, having allowed 4,747 geometries to pass through. The derivative-sign constraints were nonbinding in this search, while the support, root-shear, and root-twist screens allowed 13,418, 11,519, and 13,365 geometries to pass through, respectively. Their intersection resulted in 2,998 robustly feasible geometries, of which 198 were nondominated in the three-objective space. Twenty geometries were frozen for solver evaluation.

\begin{table}[t]
\centering
\caption{Independent continuous-search screening counts. Individual constraint counts are not sequential; ``all hard constraints'' refers to their simultaneous intersection.}
\label{tab:constraint_screening}
\begin{tabular}{lrr}
\toprule
Screen & Geometries & Fraction of 16,384 \\
\midrule
Robust lift at all three speeds & 4,747 & 28.97\% \\
Derivative signs at all three speeds & 16,384 & 100.00\% \\
Support at all three speeds & 13,418 & 81.90\% \\
Root-shear limit at all three speeds & 11,519 & 70.31\% \\
Root-twist limit at all three speeds & 13,365 & 81.57\% \\
All hard constraints & 2,998 & 18.30\% \\
Nondominated geometries & 198 & 1.21\% \\
Frozen representatives & 20 & 0.12\% \\
\bottomrule
\end{tabular}
\end{table}

\begin{figure}[t]
\centering
\includegraphics[width=0.82\textwidth]{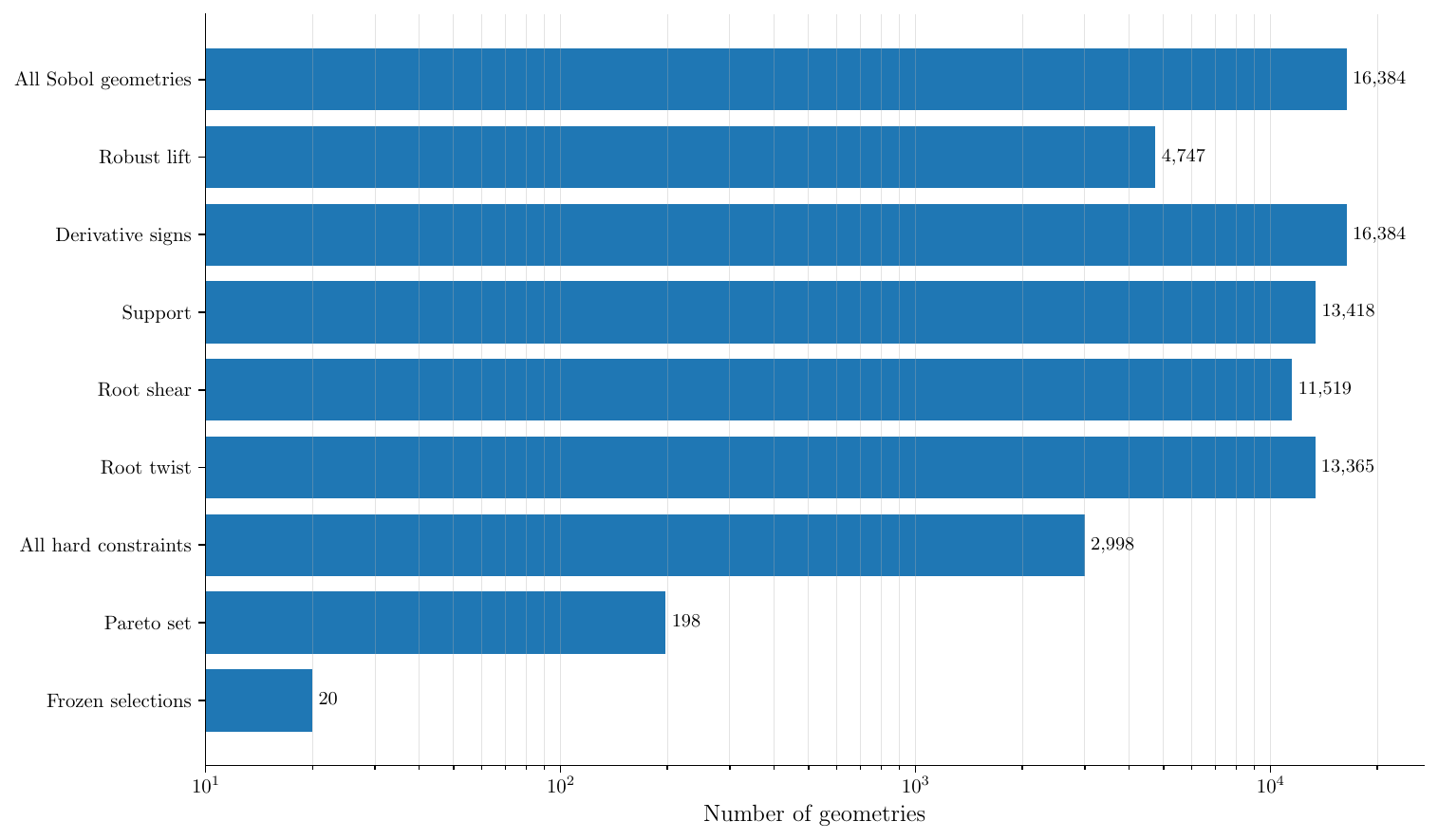}
\caption{Independent screening counts for the continuous three-speed search.}
\label{fig:constraint_screening}
\end{figure}

\begin{figure}[t]
\centering
\includegraphics[width=0.84\textwidth]{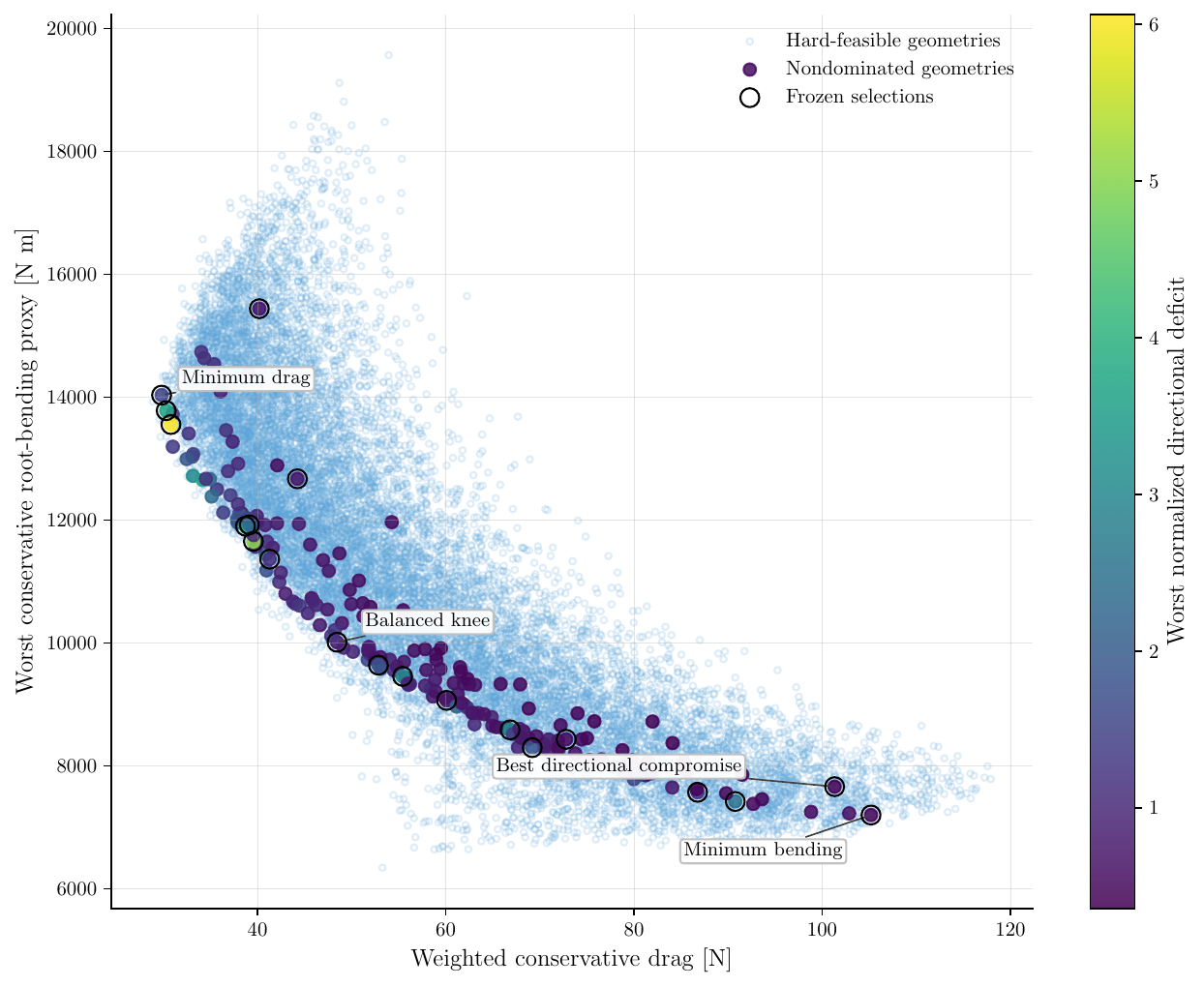}
\caption{Conservative trade-off between drag and root-bending. Light blue circles represent all 2,998 hard-feasible geometries, colored markers represent the 198 nondominated geometries, and the black circles represent the 20 frozen geometries; the color represents the directional-deficit objective.}
\label{fig:drag_bending_pareto}
\end{figure}

\begin{figure}[t]
\centering
\includegraphics[width=0.84\textwidth]{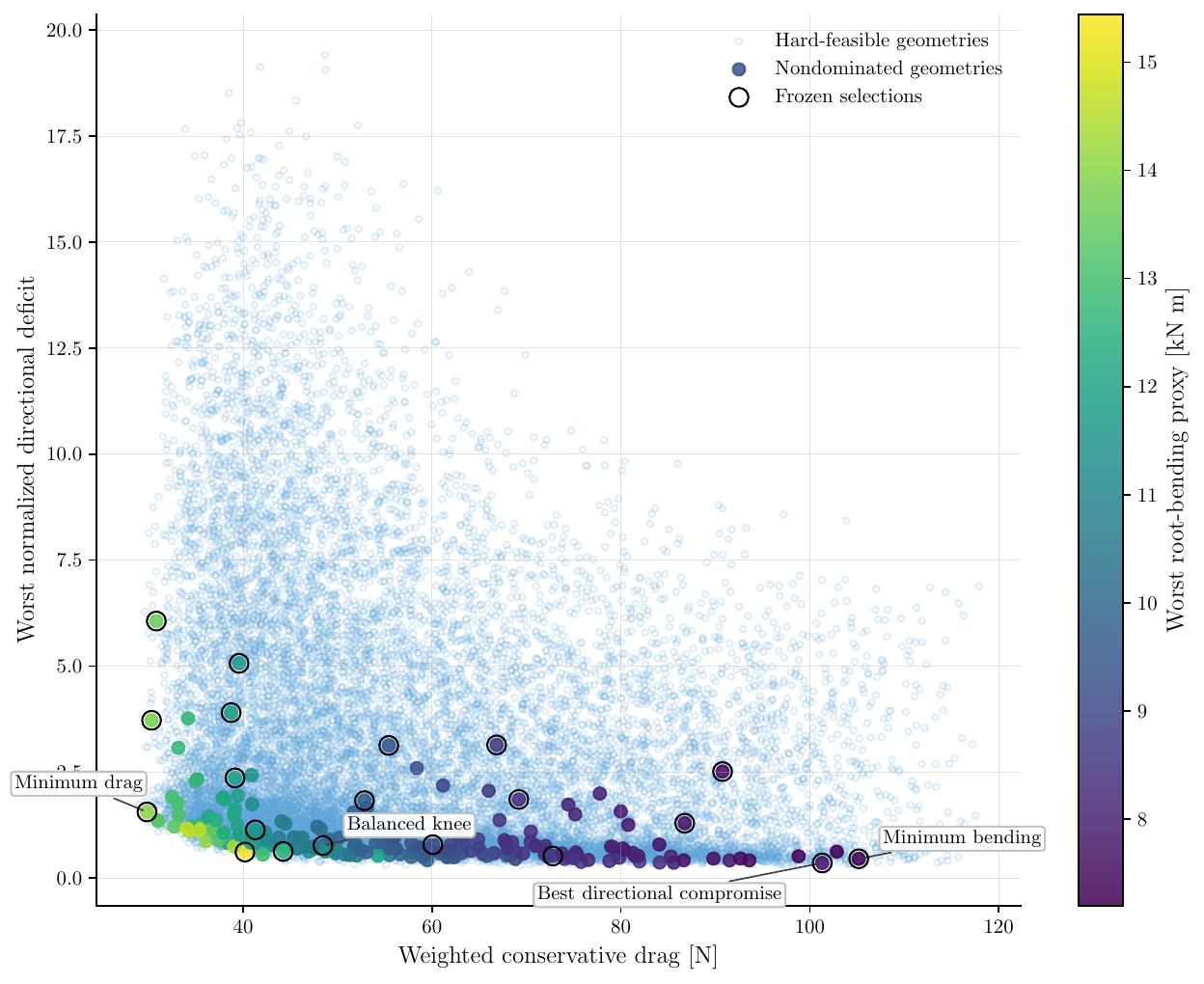}
\caption{Conservative drag vs. normalized directional deficit. The color indicates the worst root-bending proxy, and black circles indicate the 20 frozen candidates; no design can simultaneously optimize all three objectives.}
\label{fig:drag_directional_pareto}
\end{figure}

The Pareto set covers an extensive trade-off between drag and root-bending, where the conservative weighted drag varies between 29.81 and 105.19~N, and the worst conservative root-bending proxy varies between 7.20 and 15.44~kN\,m. The long and slender wings exhibit small drag, but generate a large root-bending load, while the shorter wings decrease the bending proxy with the increase in induced drag. Fig.~\ref{fig:drag_bending_pareto} depicts this relationship and uses color to expose the third objective. Fig.~\ref{fig:drag_directional_pareto} also demonstrates that the smallest directional deficit exists close to the high-drag and low-bending region instead of the minimum-drag.

\begin{figure}[t]
\centering
\includegraphics[width=0.72\textwidth]{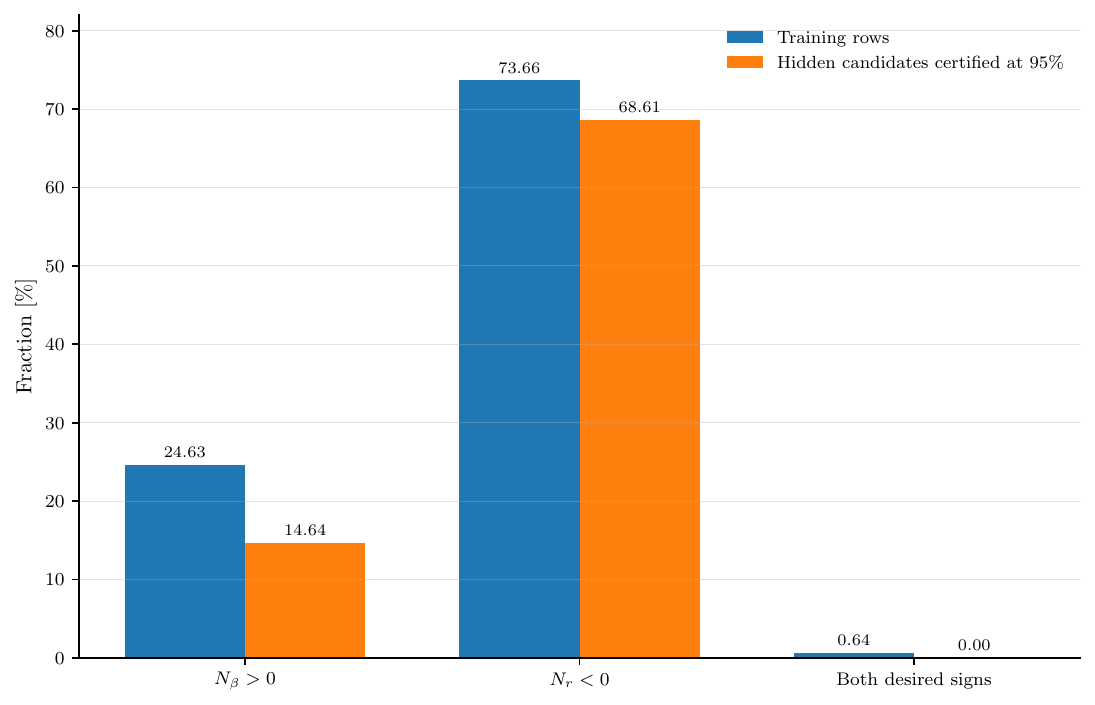}
\caption{Coexistence of the adopted $N_\beta$ and $N_r$. Counts in hidden pools use 95\% conformal sign certification; the interpretation depends on the documented derivative convention.}
\label{fig:directional_sign_coexistence}
\end{figure}

It was necessary to include the directional objective as appropriate signs do not often coexist in the database. While 24.63\% of training cases have $N_\beta>0$ and 73.66\% have $N_r<0$, only 0.6356\% satisfy both requirements. Among the candidates in the hidden database, 14.65\% and 68.61\% confirm the individual requirements at 95\%, but none confirm both. Figure~\ref{fig:directional_sign_coexistence} thus demonstrates the reason why they have to be treated as a graded compromise instead of rejecting the full continuous search through two conflicting hard sign constraints. The issue is stated conservatively, as its physical interpretation relies on the exact solver axis definition.

\subsection{Representative candidate wings}
\label{sec:selected_wing_results}

The table (Table~\ref{tab:representative_wings}) demonstrates the four main representatives. Representative 3055 is the design with the lowest drag. It has a semi-span of 9.93~m, which makes a full span of 19.86~m, and its conservative weighted drag is 29.81~N; the worst bending proxy associated with it is 14.03~kN\,m. Representative 12349 is the representative with the minimum-bending design, having a full span of 10.26~m and the worst bending proxy of 7.20~kN\,m; however, its weighted drag increases up to 105.19~N. Representative 16074 has the lowest directional deficit, of 0.3573, at similar drag and bending levels to the minimum-bending wing. Representative 6996 is the normalized knee and offers an intermediate 15.21~m span, 48.45~N drag, 10.01~kN\,m bending proxy, and a  0.7679 directional deficit.

\begin{table}[t]
\centering
\caption{Four primary frozen Pareto representatives. Span values shown in the table represent full span $2b$, while the simulator input value of $b$ represents semi-span. Objective values for all cases are conservative 95\%.}
\label{tab:representative_wings}
\scriptsize
\setlength{\tabcolsep}{3.2pt}
\begin{tabularx}{\textwidth}{>{\raggedright\arraybackslash}X c c c c c c c c}
\toprule
Role & ID & $c_r$ & $2b$ & $\lambda$ & $\Lambda$ & $J_D$ & $J_B$ & $J_{\mathrm{dir}}$ \\
 & & [m] & [m] & & [deg] & [N] & [kN m] & \\
\midrule
Minimum drag & 3055 & 0.702 & 19.863 & 0.400 & 2.194 & 29.81 & 14.03 & 1.561 \\
Minimum bending & 12349 & 1.092 & 10.255 & 0.528 & 0.639 & 105.19 & 7.20 & 0.456 \\
Best directional compromise & 16074 & 1.151 & 10.415 & 0.385 & 2.842 & 101.32 & 7.66 & 0.357 \\
Balanced knee & 6996 & 0.900 & 15.210 & 0.318 & 2.898 & 48.45 & 10.01 & 0.768 \\
\bottomrule
\end{tabularx}
\end{table}

\begin{figure*}[t]
\centering
\begin{subfigure}[t]{0.48\textwidth}
\centering
\includegraphics[width=\linewidth]{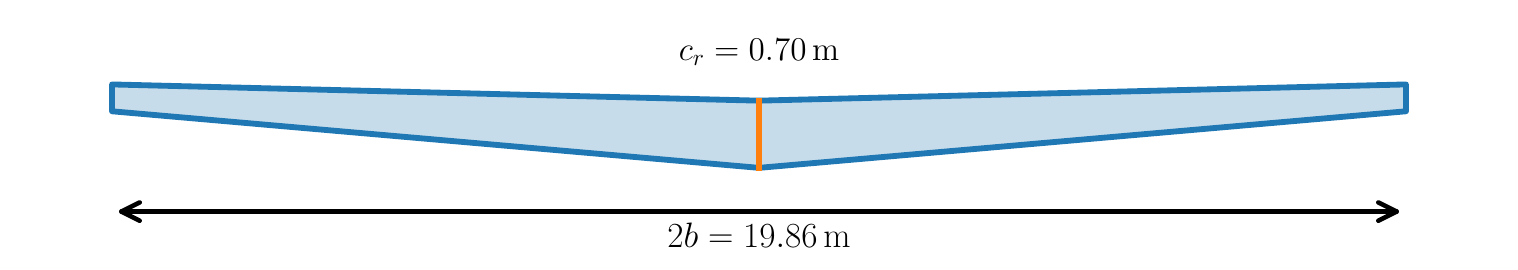}
\caption{Minimum drag (3055).}
\end{subfigure}\hfill
\begin{subfigure}[t]{0.48\textwidth}
\centering
\includegraphics[width=\linewidth]{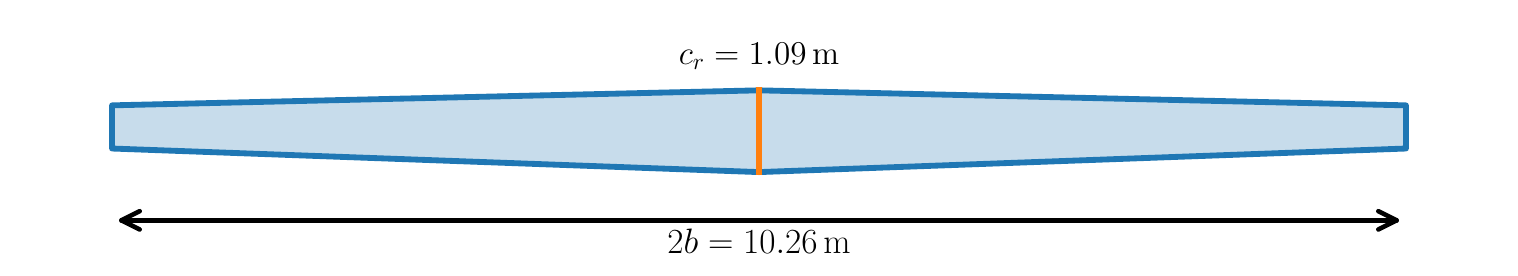}
\caption{Minimum bending (12349).}
\end{subfigure}

\begin{subfigure}[t]{0.48\textwidth}
\centering
\includegraphics[width=\linewidth]{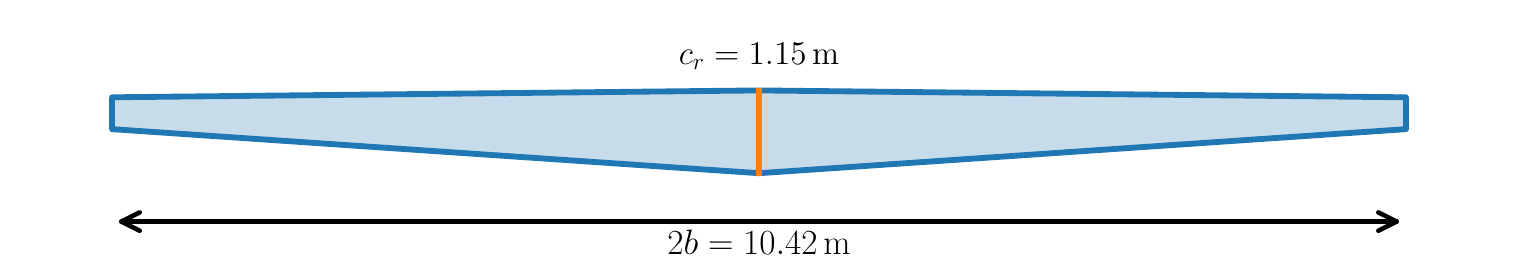}
\caption{Best directional compromise (16074).}
\end{subfigure}\hfill
\begin{subfigure}[t]{0.48\textwidth}
\centering
\includegraphics[width=\linewidth]{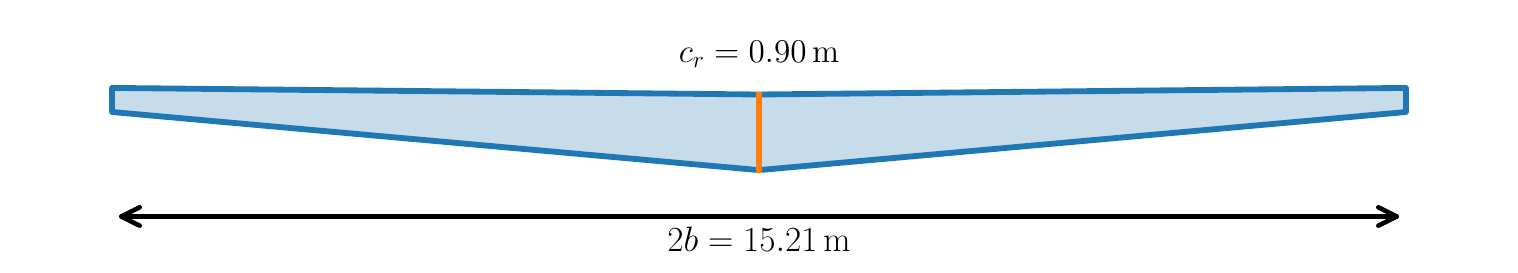}
\caption{Balanced knee (6996).}
\end{subfigure}

\begin{subfigure}[t]{0.48\textwidth}
\centering
\includegraphics[width=\linewidth]{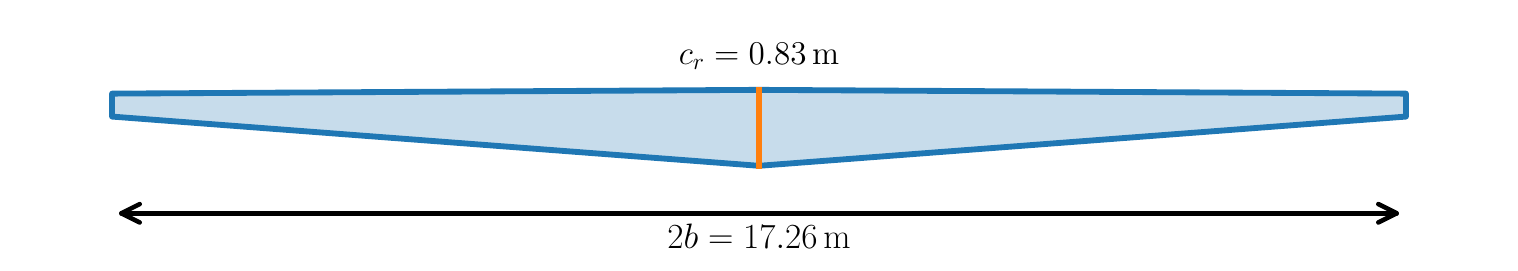}
\caption{Diverse Pareto (13595).}
\end{subfigure}\hfill
\begin{subfigure}[t]{0.48\textwidth}
\centering
\includegraphics[width=\linewidth]{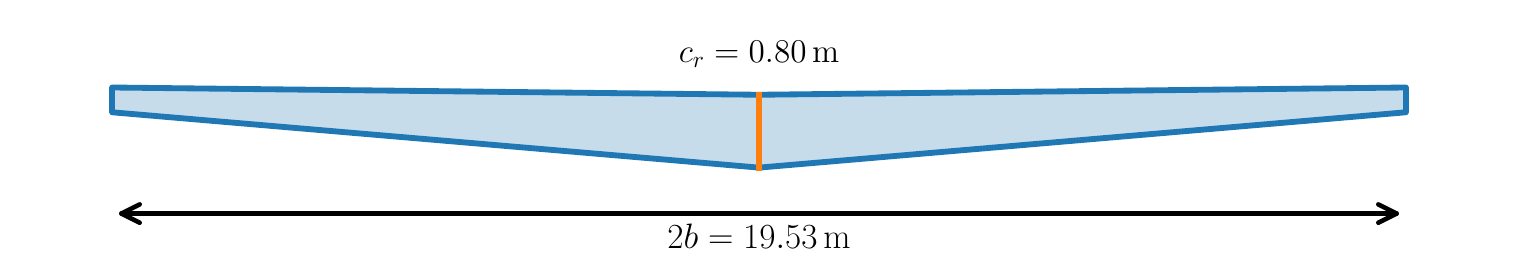}
\caption{Diverse Pareto (15411).}
\end{subfigure}

\begin{subfigure}[t]{0.48\textwidth}
\centering
\includegraphics[width=\linewidth]{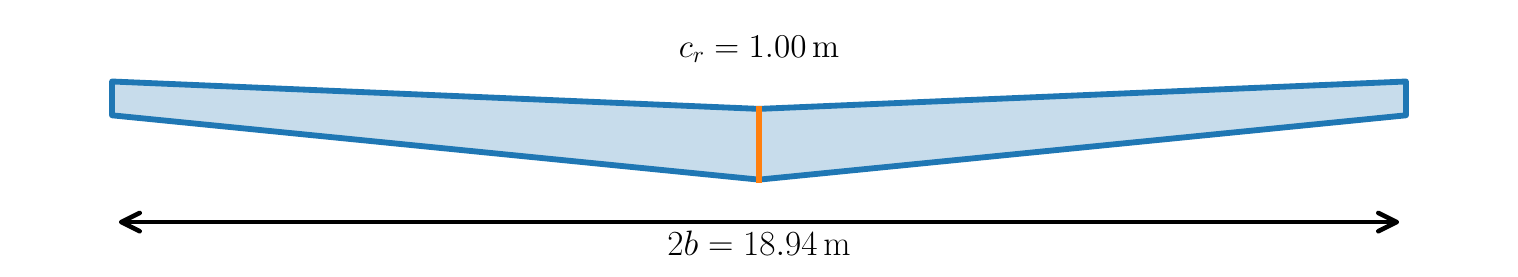}
\caption{Diverse Pareto (8926).}
\end{subfigure}\hfill
\begin{subfigure}[t]{0.48\textwidth}
\centering
\includegraphics[width=\linewidth]{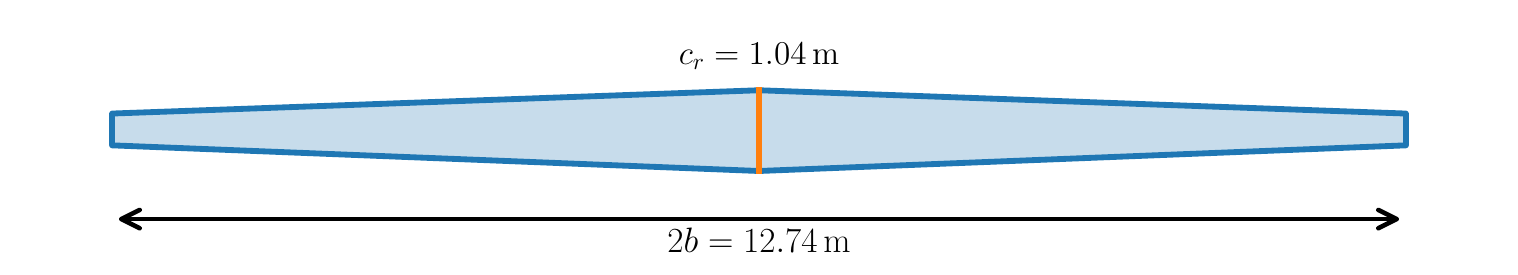}
\caption{Diverse Pareto (4273).}
\end{subfigure}
\caption{Eight frozen final wing designs. Numbers within the parentheses refer to candidate IDs. The first four plots refer to the primary Pareto representatives as previously mentioned, while the remaining four refer to additional diverse representatives obtained from the predicted nondominated set. $c_r$ represents the root chord, the arrow represents the full span $2b$, and finally, the stored simulator variable $b$ represents the semi-span.}
\label{fig:representative_planforms}
\end{figure*}

\begin{figure}[t]
\centering
\includegraphics[width=0.72\textwidth]{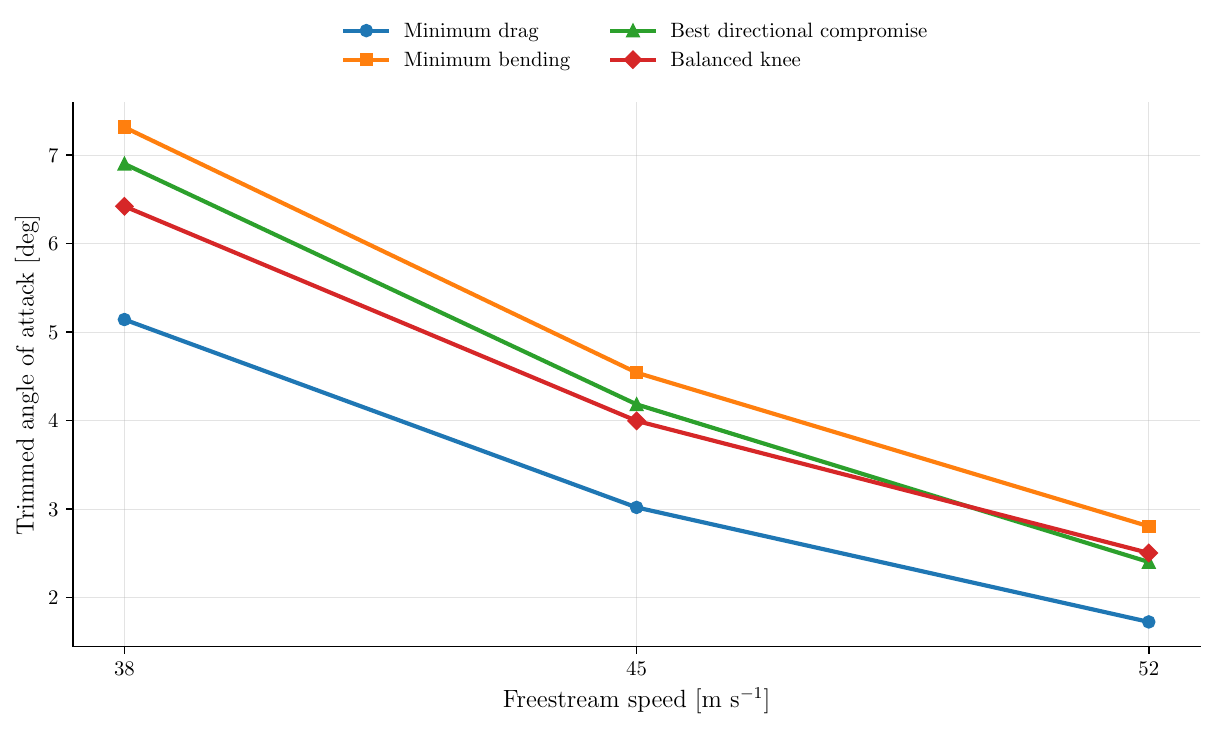}
\caption{Trimmed angle of attack for the four primary Pareto representatives. Trim is found via mean prediction; interval feasibility is evaluated afterwards with the 95\% global conformal bounds.}
\label{fig:representative_trim_angles}
\end{figure}

The trim angle requirement monotonically decreases with speed for all four representative wings (see Fig.~\ref{fig:representative_trim_angles}). For 38~m\,s$^{-1}$, the trimmed angles vary from 5.14$^\circ$ for the minimum-drag design to 7.31$^\circ$ for the minimum-bending design. For 52~m\,s$^{-1}$, the range reduces to 1.73--2.81$^\circ$. The complete lift intervals at 95\% are fully contained within the imposed $\pm5\%$ band; the maximum worst-case relative error for any of the four wings is 4.70\%.

Four additional diverse representatives selected from the previously considered four primary Pareto sets are included in Fig.~\ref{fig:representative_planforms} to demonstrate the geometric width of the frozen final set.

\FloatBarrier

\subsection{External validation on 60 new Tornado simulations}
\label{sec:external_validation_results}

Once the candidate geometries, operating points, conformal quantiles, support threshold, and selection roles had been frozen, the 20 wings were re-assessed using Tornado at 38, 45, and 52 m s$^{-1}$. The 60 operating points generated matched the frozen simulator input file to within $5.0\times10^{-10}$ in absolute input value. These points were not used in order to retrain the network, architecture selection, conformal recalibration, optimizer retuning, or candidate reselection.

\subsubsection{External point-prediction accuracy}
\label{sec:external_point_accuracy}

\begin{figure}[H]
\centering
\includegraphics[width=0.76\textwidth]{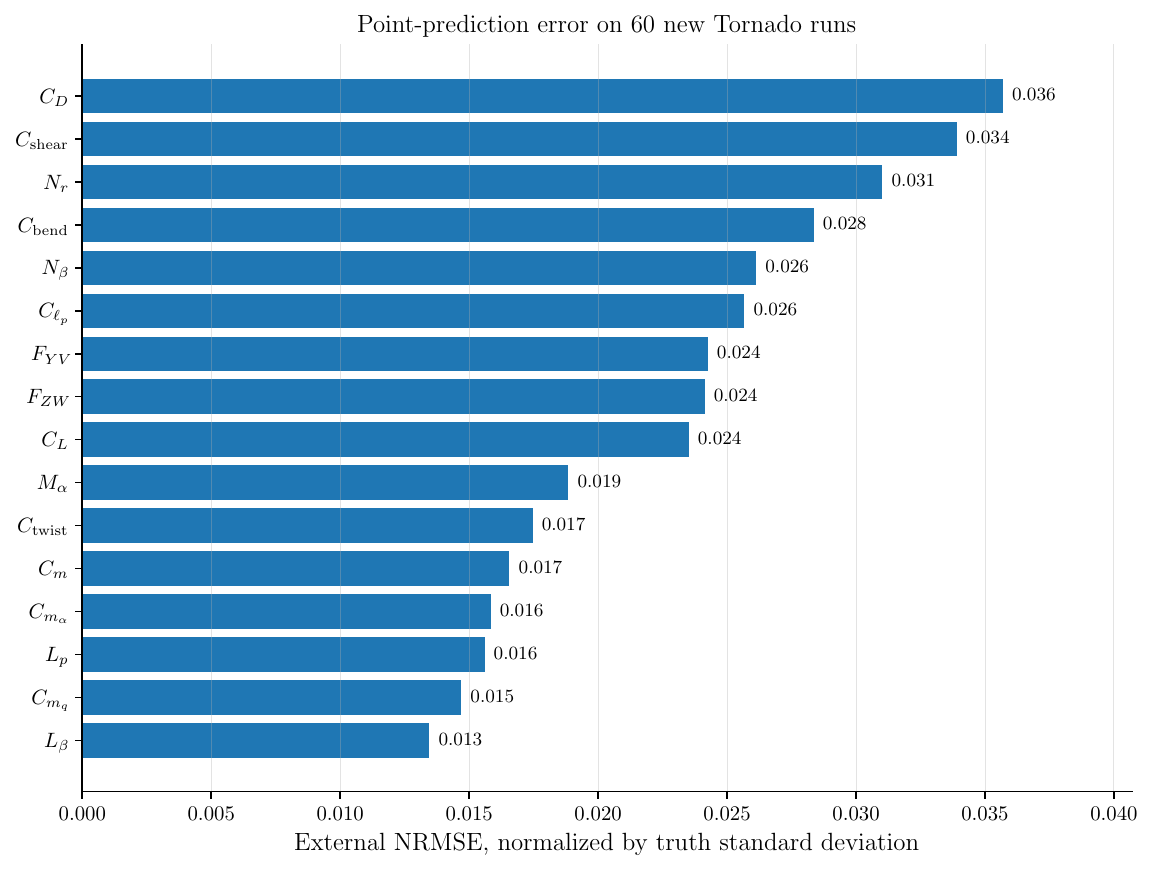}
\caption{Target-wise point-prediction error on the 60 new Tornado runs. NRMSE is normalized by the standard deviation of the external truth for each target.}
\label{fig:external_target_nrmse}
\end{figure}

For the 16 learned targets, the mean NRMSE normalized with respect to the external truth standard deviation was 0.0228, the median was 0.0238, and each target maintained $R^2\geq 0.9987$, with a mean $R^2$ of 0.9994. The target-specific NRMSE varied from 0.0135 for $L_\beta$ to 0.0357 for $C_D$ (Fig.~\ref{fig:external_target_nrmse}). These results indicate that the accuracy achieved on the frozen ID and structured-OOD partitions transferred to optimized, non-tabulated geometries.

Table~\ref{tab:external_dimensional_accuracy} lists decoded dimensional errors. The mean absolute errors were 35.34~N for lift, 0.726~N for drag, 28.70~N for root shear, 81.75~N\,m for the root-bending proxy, and 0.0581~N\,m for the root-twist proxy. In relation to the mean absolute value of each simulated quantity, these values correspond to 0.60\%, 1.33\%, 0.98\%, 0.82\%, and 0.67\%, respectively. As trim compresses the variation of lift and root shear over the selected cases, variance-based $R^2$ does not carry any information for these two quantities; absolute and relative errors are more suitable measures.

\begin{table}[!htbp]
\centering
\caption{External dimensional prediction accuracy. Relative MAE is normalized with respect to the mean absolute Tornado value. Coverag implements the global joint 95\% interval decoded into dimensional units.}
\label{tab:external_dimensional_accuracy}
\begin{tabular}{lrrrr}
\toprule
Quantity & MAE & RMSE & Relative MAE [\%] & Coverage [\%] \\
\midrule
Lift [N] & 35.336 & 45.784 & 0.60 & 100.00 \\
Drag [N] & 0.726 & 1.215 & 1.33 & 96.67 \\
Root shear [N] & 28.705 & 36.000 & 0.98 & 100.00 \\
Root bending [N m] & 81.75 & 113.09 & 0.82 & 100.00 \\
Root twisting [N m] & 0.0581 & 0.0796 & 0.67 & 100.00 \\
\bottomrule
\end{tabular}
\end{table}

\begin{figure}[H]
\centering
\begin{subfigure}[t]{0.46\textwidth}
\centering
\includegraphics[width=\linewidth]{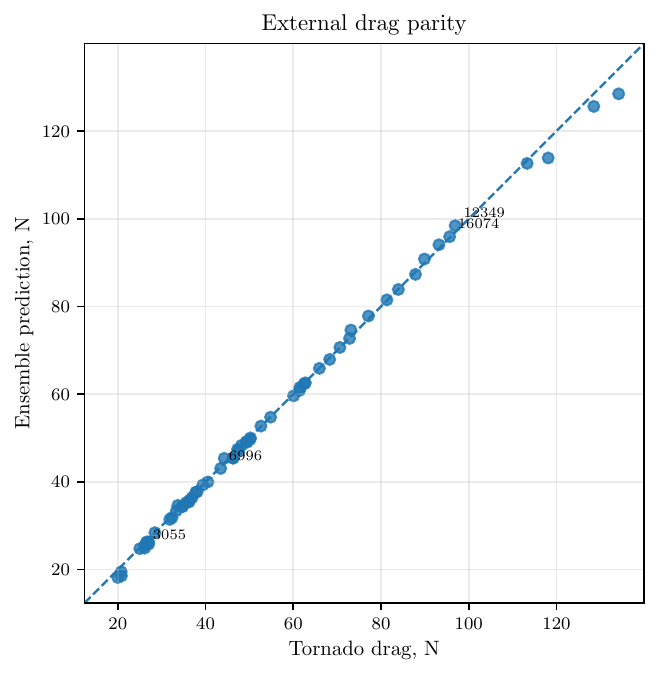}
\caption{Mission-point drag.}
\end{subfigure}\hfill
\begin{subfigure}[t]{0.46\textwidth}
\centering
\includegraphics[width=\linewidth]{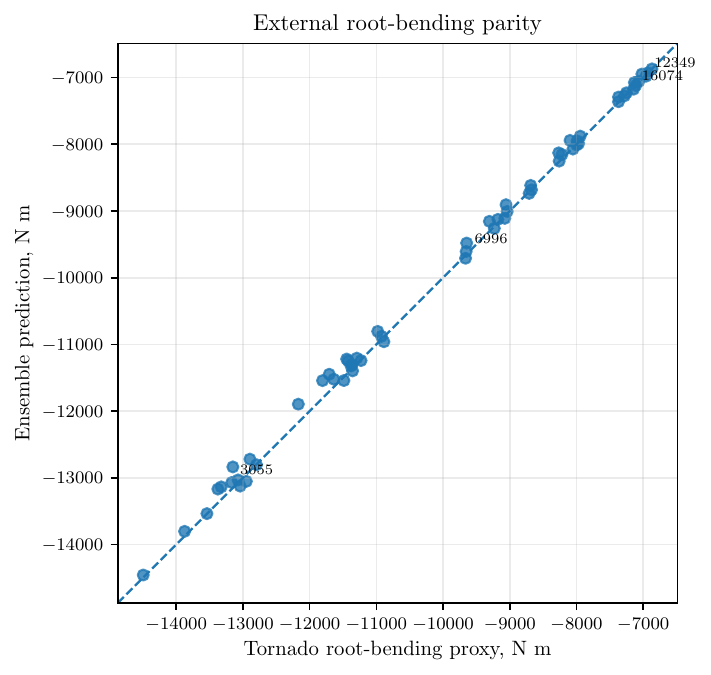}
\caption{Root-bending proxy.}
\end{subfigure}
\caption{External parity plots for two primary optimization quantities. Candidate labels refer to the four named representatives at the middle speed.}
\label{fig:external_dimensional_parity}
\end{figure}

\subsubsection{Simultaneous uncertainty coverage}
\label{sec:external_joint_coverage}

The raw ensemble interval covered the full set of 14 optimization outputs simultaneously in only 10 runs out of 60 runs (16.67\%). On the other hand, the global 90\% conformal interval covered 57 out of 60 runs (95.00\%), and the formal global 95\% interval covered 58 out of 60 runs (96.67\%). The support-Mondrian 95\% interval yielded the same 58 out of 60 result since all optimized operating points satisfied the frozen support requirement. Through all 16 learned outputs and 60 runs, the global 95\% intervals achieved 99.79\% marginal coverage, and no individual target achieved less than 96.67\% coverage.

\begin{table}[!htbp]
\centering
\caption{Empirical simultaneous coverage across the 60 new Tornado simulations. The chosen points are not claimed to be exchangeable with the calibration set; this is an external empirical verification of the frozen intervals.}
\label{tab:external_conformal_coverage}
\begin{tabular}{lrr}
\toprule
Method & Covered runs & Joint coverage \\
\midrule
Raw ensemble 95\% & 10/60 & 0.1667 \\
Global joint 90\% & 57/60 & 0.9500 \\
\textbf{Global joint 95\%} & \textbf{58/60} & \textbf{0.9667} \\
Support-Mondrian 95\% & 58/60 & 0.9667 \\
\bottomrule
\end{tabular}
\end{table}

\begin{figure}[H]
\centering
\includegraphics[width=0.62\textwidth]{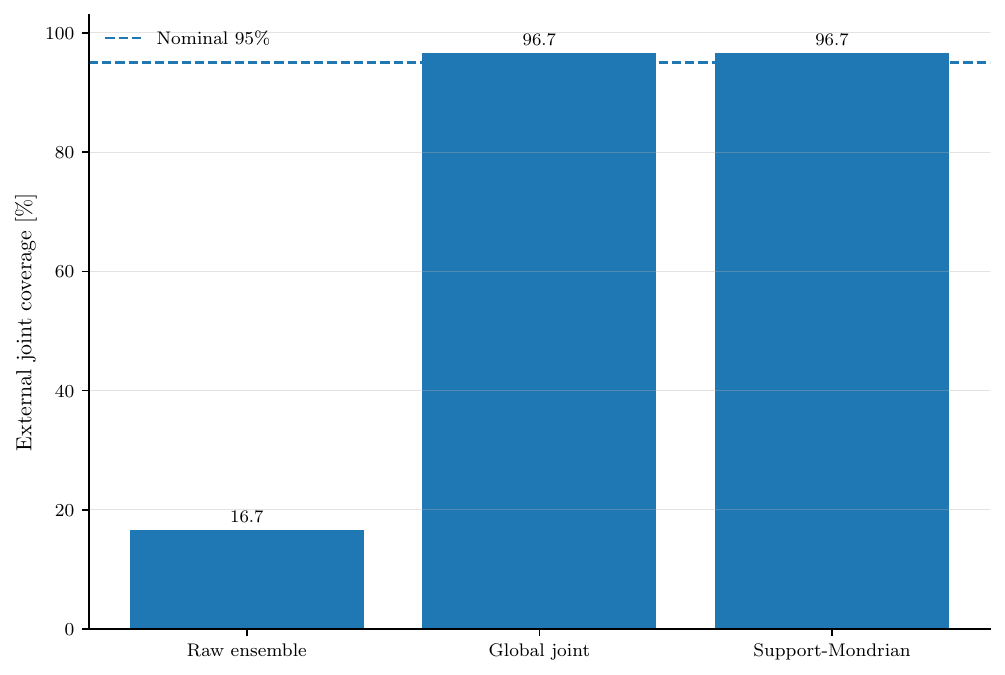}
\caption{Empirical joint coverage over the 60 simulated points. While the global joint interval and support-Mondrian intervals were held near-nominal simultaneous coverage after design selection, the raw ensemble dispersion remained undercalibrated.}
\label{fig:external_conformal_coverage}
\end{figure}

The two joint misses occurred only for $C_D$ at 38~m,s$^{-1}$. For representative 12349, the simulated coefficient was 0.0199305, and the upper global bound was 0.0196182. For representative 16074, the simulated coefficient was 0.0196731, and the upper bound was 0.0196406. All other certified outputs were covered in both rows. Neither miss caused a failure of the dimensional drag upper bound or a violation of any hard design constraint.

\subsubsection{Actual hard-constraint and objective-bound verification}
\label{sec:external_constraint_verification}

\begin{figure}[H]
\centering
\includegraphics[width=0.78\textwidth]{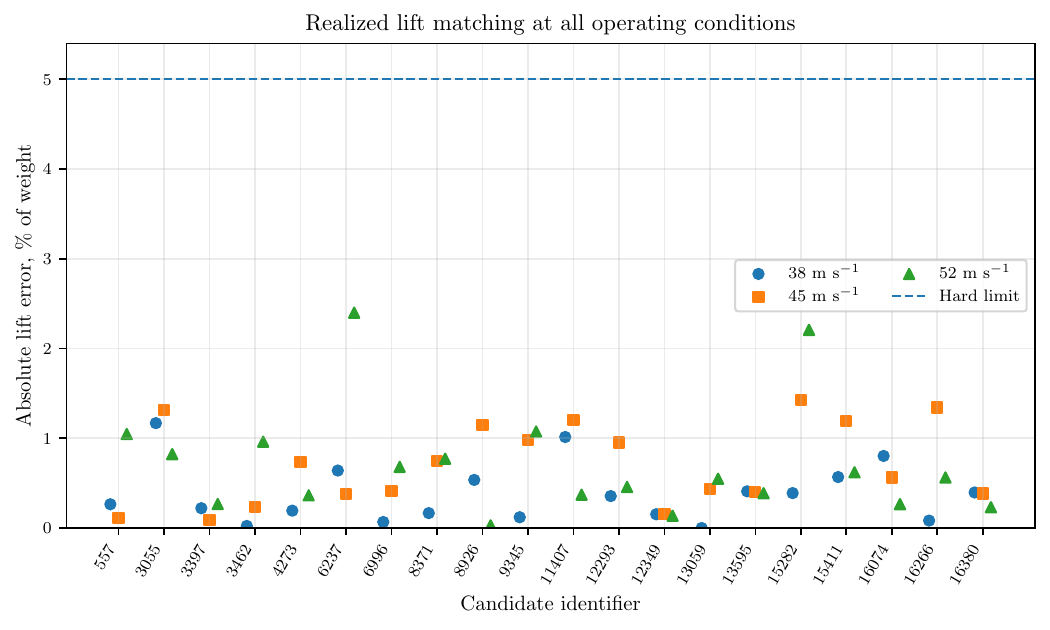}
\caption{Lift-matching error for all frozen candidates and speeds. The largest perceived error was 2.40\%, compared with the 5\% hard limit.}
\label{fig:external_lift_error}
\end{figure}

All the operating points met the entire hard feasibility definition. In particular, the 60 newly generated outputs were within $\pm5\%$ of the required lift, had negative $C_{m_\alpha}$, $C_{\ell_p}$, and $C_{m_q}$, and complied with the constraints on root-shear and root-twist coefficients; moreover, all corresponding inputs remained below the frozen support threshold prior to simulation. The largest lift mismatch was 2.40\% of the required weight, still well below the 5\% bound (Fig.~\ref{fig:external_lift_error}). Therefore, all 20 wings could be considered feasible at all three speeds.

\begin{table}[!htbp]
\centering
\caption{Verification of the hard feasibility components for 60 post-selection operating points. Aerodynamic quantities are evaluated using the new Tornado outputs; support is the frozen input-state criterion, checked independently prior to simulation.}
\label{tab:external_constraint_success}
\begin{tabular}{lrr}
\toprule
Constraint & Passing runs & Success [\%] \\
\midrule
Lift within $\pm5\%$ of $W$ & 60/60 & 100 \\
$C_{m_\alpha}<0$, $C_{\ell_p}<0$, and $C_{m_q}<0$ & 60/60 & 100 \\
Input-state support below the frozen limit & 60/60 & 100 \\
Root-shear coefficient below its frozen limit & 60/60 & 100 \\
Root-twist coefficient below its frozen limit & 60/60 & 100 \\
All hard constraints simultaneously & 60/60 & 100 \\
\bottomrule
\end{tabular}
\end{table}

\begin{figure}[H]
\centering
\begin{subfigure}[t]{0.32\textwidth}
\centering
\includegraphics[width=\linewidth]{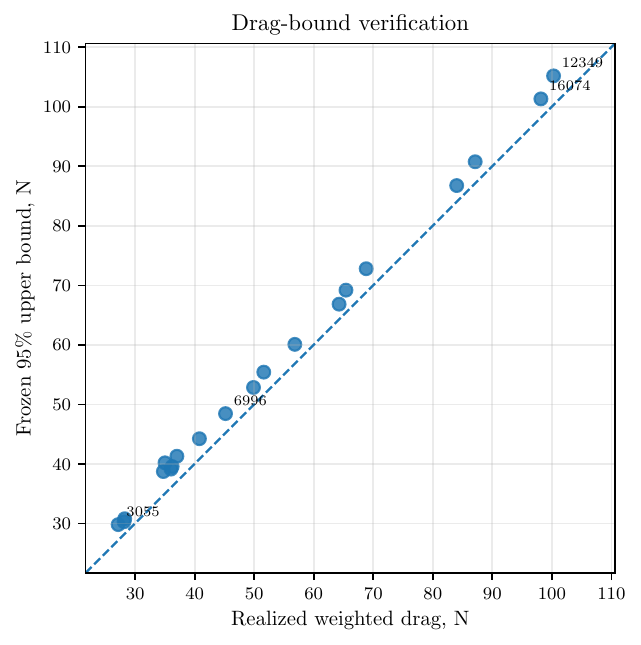}
\caption{Weighted drag.}
\end{subfigure}\hfill
\begin{subfigure}[t]{0.32\textwidth}
\centering
\includegraphics[width=\linewidth]{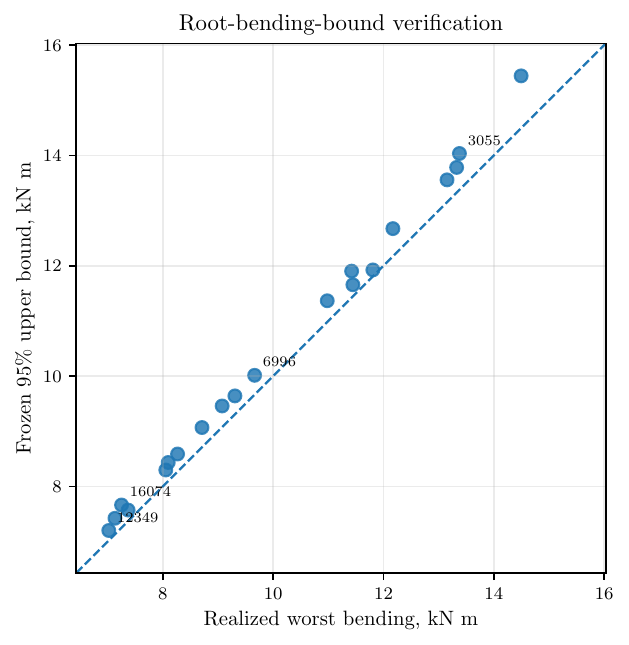}
\caption{Worst bending.}
\end{subfigure}\hfill
\begin{subfigure}[t]{0.32\textwidth}
\centering
\includegraphics[width=\linewidth]{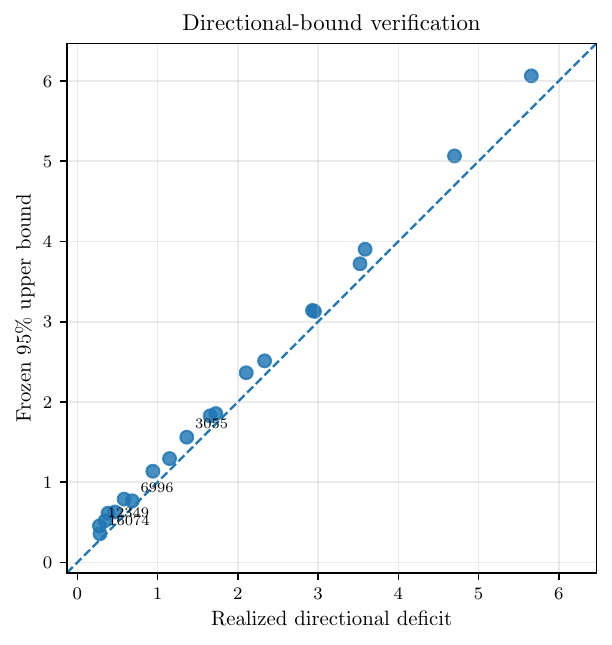}
\caption{Directional deficit.}
\end{subfigure}
\caption{Verification of external objectives for the four named candidates. Each cell contains realized value / frozen 95\% upper bound.}
\label{fig:external_bound_verification}
\end{figure}

From the design perspective, each frozen conservative objective bound was successful: realized weighted drag was less than its upper bound for 20/20 wings, worst root bending was less than its upper bound for 20/20, and realized directional deficit was less than its upper bound for 20/20. 
Figure~\ref{fig:external_bound_verification} depicts the realized values against their frozen bounds; all points remain below the identity line.
For the four designed wings, Table~\ref{tab:principal_external_objectives} provides realized and bounded values.

\begin{table}[!htbp]
\centering
\caption{External objective verification for the four named candidates. Each cell reports realized value / frozen 95\% upper bound.}
\label{tab:principal_external_objectives}
\scriptsize
\begin{tabularx}{\textwidth}{l c >{\centering\arraybackslash}X >{\centering\arraybackslash}X >{\centering\arraybackslash}X}
\toprule
Role & ID & Drag [N] & Bending [kN m] & Directional deficit \\
\midrule
Minimum drag & 3055 & 27.15 / 29.81 & 13.38 / 14.03 & 1.363 / 1.561 \\
Minimum bending & 12349 & 100.29 / 105.19 & 7.02 / 7.20 & 0.274 / 0.456 \\
Best directional compromise & 16074 & 98.16 / 101.32 & 7.25 / 7.66 & 0.281 / 0.357 \\
Balanced knee & 6996 & 45.18 / 48.45 & 9.66 / 10.01 & 0.684 / 0.768 \\
\bottomrule
\end{tabularx}
\end{table}

The identities of the minimum drag and minimum bending candidates were preserved by the simulations, and candidate 6996 remained the point closest to the realized normalized three-objective ideal. The smallest realized directional deficit was 0.274 for candidate 12349, just slightly below 0.281 for candidate 16074, chosen by the conservative predicted objective. This minor change in ranks does not contradict the bound, as both of the realized deficits remained below their frozen upper bounds.

\subsubsection{Externally realized Pareto structure}
\label{sec:external_pareto_structure}

\begin{figure}[H]
\centering
\includegraphics[width=0.76\textwidth]{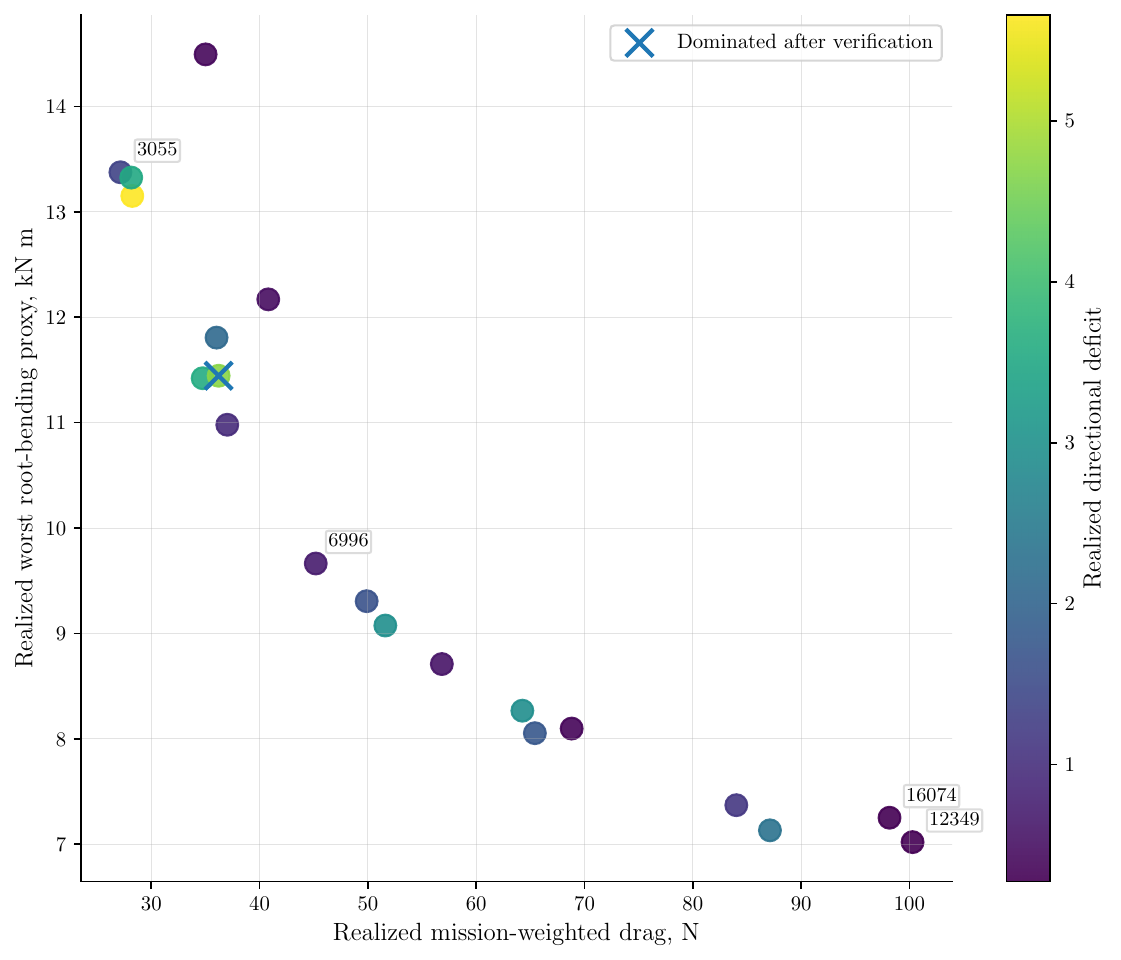}
\caption{Objective trade-offs realized for 20 frozen candidates. The color indicates directional deficit. Nineteen designs are nondominated through the evaluated set, and the cross identifies candidate 13595 13595, which is dominated by candidate 8371.}
\label{fig:external_realized_pareto}
\end{figure}

Nineteen out of twenty frozen designs were nondominated in the externally simulated subset (Fig.~\ref{fig:external_realized_pareto}). Candidate 13595 became dominated by candidate 8371 after evaluation by the solver. This behavior is expected as the purpose of the final farthest-point procedure is the wide coverage of the predicted Pareto surface with surrogate uncertainty instead of exact preservation of nondomination of every representative after further evaluations. Note, however, that candidate 13595 still satisfied all the hard constraints and conservative objective bounds.

Taken together, the external study demonstrates four separate properties of the framework: high point prediction accuracy on simulated optimized geometries, near-nominal simultaneous conformal coverage,  hard-feasibility success, and conservative design-level objective bounds. This is not cross-fidelity aerodynamic validation, as both the training data and verification runs implement the same Tornado solver; however, it validates the end-to-end surrogate optimization pipeline at geometries absent in the training table.

\section{Discussion}
\label{sec:discussion}

\subsection{Effect of physics-structured surrogate modeling}
\label{sec:discussion_surrogate}

The benchmark outcomes support implementing structured inductive decomposition for this particular set of responses. In comparison to the compact unrestricted multilayer perceptron and LightGBM, the five-member ensemble reduced the mean in-distribution NRMSE by 39.5\% and 32.8\%, respectively, while also producing the lowest aggregated error in five of the six structured out-of-distribution categories. It is not an indication of the fact that a dual-head architecture is going to dominate every single aerodynamic surrogate problem. Instead, it indicates that the separation of a similarity-based aerodynamic mapping from dynamics mapping proved useful when the learned outputs mixed nondimensional aerodynamic coefficients, normalized root-load coefficients, and dimensional derivatives with distinct scaling behavior.

The exact decoder plays an essential role in the results. Lift, drag, and root loads were not assumed to be unrelated regression targets. Instead, lift, drag, and root loading were constructed from predicted coefficients by using dynamic pressure, area, full span, and mean aerodynamic chord. With this construction, redundant learning is minimized, and dimensional consistency becomes transparent. This enables conservative coefficient intervals to be mapped into dimensional objectives and constraints without needing to learn a second network. This approach is also less restrictive than a fully physics-informed neural network, as it does not force flow equations, yet it is more structured than black-box regression models.

However, the remaining OOD gap is important. The ensemble's OOD NRMSE of 0.0595 was about 2.7 times that of its in-distribution value, and the geometry-corner cases were the only OOD category where the LightGBM model slightly outperformed the ensemble model. This behavior is in accordance with the general challenge of preserving both accuracy and reliable uncertainty in the presence of dataset shift \cite{ovadia2019trust}. The support constraint therefore plays a different role from the ensemble model, as the network improves the response approximation while the support metric prevents the optimizer from treating that approximation as equally trustworthy everywhere.

\subsection{Role of conformal calibration in design decisions}
\label{sec:discussion_conformal}

The uncertainty results explain why the raw ensemble spread did not make for a design certificate directly. Despite being highly accurate, its nominal raw 95\% intervals simultaneously cover all 14 optimization results in just 16.67\% of the 60 new simulations. After being calibrated by a global max-score conformal algorithm, the empirical simultaneous coverage improved to 96.67\%. The contrast is critical for the suggested framework: while point accuracy and ensemble diversity proved useful for predictions, an additional calibration partition was necessary in order to convert dispersion into interpretable intervals.

Joint calibration was also more suitable than target-wise marginal calibration for the optimization problem since a candidate is acceptable only if all the required lift, damping signs, support, and root-load conditions are satisfied simultaneously at every operating point. Even strong marginal coverage may produce poor vector coverage when numerous outputs are evaluated simultaneously. The maximum normalized nonconformity score generates a rectangular uncertainty set, which is conservative yet simple to decode and impose. The connection between conformal regions and robust optimization has already been established in a general sense \cite{johnstone2021conformal}, although the current setup is specific to multiple aerodynamic and dynamic outputs and evaluated after adaptive candidate selection.

The difference between formal and empirical statements still holds. The calibration guarantee applies with respect to the calibration distribution under exchangeability. Since the 20 optimized geometries were adaptively chosen, they cannot be considered exchangeable random test samples; thus, the 58/60 result must be viewed as empirical post-selection verification rather than a novel finite-sample guarantee. Support-Mondrian calibration improved structured OOD coverage from 73.81\% to 91.08\%, although its adaptively constructed bins were treated as an exploratory diagnostic. The global joint interval was held as the formal certificate in the frozen optimization pipeline.

\subsection{Aerodynamic and root-load trade-offs}
\label{sec:discussion_tradeoffs}

A robust search resulted in an evident conflict between VLM-derived drag and the aerodynamic root-bending proxy. The minimum-drag wing has a full span of 19.86~m and a weighted drag of 29.81~N, while its worst bending proxy was 14.03~kN\,m. For the minimum bending proxy wing, the full span decreased to 10.26~m, the bending proxy decreased to 7.20~kN\,m, while the weighted drag increased to 105.19~N. Finally, the balanced-knee wing design remained in the middle of this trade-off, with a full span of 15.21~m, a weighted drag of 48.45~N, and a worst bending proxy of 10.01~kN\,m. Such tendencies seem realistic in regard to the induced-drag-dominated lifting surfaces, as an increase in span and aspect ratio improves induced deficiency, yet increases the moment arm in which distributed aerodynamic loading acts at the root.

Therefore, the optimization process must be referred to as a root load-aware aerodynamic design instead of structural optimization. The bending, shear, and twisting loads are aerodynamic resultants obtained from the solver. These loads do not account for material stiffness, stress concentration, skin or spar sizing, buckling, fatigue, manufacturing, or aeroelasticity. Their importance lies in ensuring that the low-fidelity aerodynamic optimization process does not pick wing geometry based only on induced drag without accounting for the large variations in load transmission. An aerostructural analysis can utilize the current Pareto set as a reduced candidate pool.

The directional objective highlighted yet another, more conservative trade-off. The desired signs of $N_\beta$ and $N_r$ rarely occur together in the accepted database and could not be verified for any hidden design at 95\%. If both were considered as hard conditions, the design space would have been ruled out. However, the normalized deficit allowed maintaining a graded comparison. Still, it is not easy to provide a physical interpretation of this trade-off as it depends on the exact body-axis, perturbation, and post-processing signs. That is the reason why the paper states the trade-off as a convention-dependent directional-derivative objective and not as a complete lateral-directional stability guarantee.

\subsection{Meaning of the simulation validation}
\label{sec:discussion_external}

The 60 final Tornado evaluations assess the entire process starting from continuous geometry generation to trim, surrogate inference, conformal decoding, constraint screening, Pareto selection, and ultimately final simulation. Each of the 60 operational points met all hard constraints, each of the 20 designs met all three mission conditions, and all of the frozen objective upper bounds were conservative. The low-dimensional errors in drag, root-bending, and root twisting demonstrate that the decoder and coefficient surrogates have stayed accurate at the selected geometries. Nineteen out of twenty wing designs remained nondominated in the tested subset; the one ranking loss came from a wing deliberately kept for diversity, and not as a named optimum.

This verification is stronger than merely reporting performance only on random database rows, as the geometries evaluated were outputs of a continuous optimization process and are not present in the original table.

Post-selection evaluation serves as validation of the surrogate and robust-selection workflow with respect to the Tornado solver. Since the 60 new cases were obtained by the same aerodynamic model that generated the training database, they demonstrate that the surrogate successfully reproduces Tornado at optimized, previously unevaluated continuous geometries and that the confidence intervals still remain conservative for the solver outputs. They do not validate the VLM aerodynamic model against Navier-Stokes CFD and wind tunnel tests or flight data.

\section{Limitations and future work}
\label{sec:limitations}

The main drawbacks are directly related to the aerodynamic model, the optimization scope, and the statistical assumptions.

First and foremost, Tornado is a potential flow VLM. Although it is well suited to attached-flow trends of the lifting surfaces, it is unable to resolve boundary layers, profile and skin friction drag, transitions, large separated zones, shock waves, and detailed interference between the wings and fuselage. Therefore, the drag reported is VLM-derived and induced-like, instead of the total aircraft drag. The fixed NACA~2412 airfoil, the fixed fuselage and tail geometry, and the reference atmosphere also further restrict the generalization. Reynolds-averaged Navier-Stokes computations with higher fidelity, wind-tunnel testing, or flight tests are still needed for quantitative validation of the selected configurations.

The adopted flight regime makes the use of VLM for the intended attached-flow analysis more compelling. Therefore, the logically defensible interpretation is that the workflow deliberately selects robust solutions with regard to uncertainties of VLM response reproduction in the sampled design space. Prior to final aerodynamic designs, representative Pareto designs need to be analyzed using transition-informed 3-D RANS computations, possibly supplemented with higher-fidelity unsteady computations, wind-tunnel measurements, or flight tests. These data can also be used in the discrepancy estimation between VLM and viscous aerodynamics and in order to construct a multi-fidelity correction model.

Second, the loads are aerodynamic proxies and not structural responses. No finite element model, material selection, spar or skin sizing, local stress analysis, buckling, fatigue, flutter, aeroelastic effects, or mass feedback are considered here. Aircraft weight is held constant at 5886~N, with only three airspeeds examined. Future research should integrate the current uncertainty-aware aerodynamic modeling approach with structural and aeroelastic simulations, permit weight variation with geometry, and increase the mission set to climbing, maneuvering, gust, and off-design conditions. Multi-fidelity transfer learning provides a natural approach for integrating the VLM dataset with fewer viscous and aeroelastic simulations \cite{tao2024multi,wu2024efficient}.

Third, the control-surface deflections and control-effectiveness derivatives are missing. The predicted static and damping derivatives characterize the aircraft response tendencies about the design configuration; they do not represent the elevator, aileron, and rudder authority. Fixed tail and fuselage parameters are used during the optimization process. A joint wing-tail design, control-surface sizing, trim analysis with actuator variables, and time-domain six-degree-of-freedom simulation are necessary to establish controllability claims.

Fourth, the conformal validity relies on distribution. Global split-conformal interval has finite-sample coverage under exchangeability, not any form of covariate shift or adaptive optimization. Support score and support-Mondrian bins can provide better diagnostics but cannot prove validity out of the sampled domain. A possible future work could consider the usage of weighted conformal predictions, covariate-shift correction, active learning, or sequential acquisition to improve weakly supported regions. New validation sets are to be preserved after every model, calibration, or optimization change; the current 60 simulations should not be reused for tuning.

Finally, the Sobol search is a finite global exploration approach rather than a proof of global Pareto optimality. The 16,384-point design ensures broad deterministic coverage while avoiding differential calculation through trim, nearest neighbor support, and interval constraints. However, small feasible regions may still go unnoticed due to their small size. Appendix~\ref{app:sensitivity} describes the sensitivity of the frozen surrogate to lift tolerance, load threshold, confidence level, and the support limit. The number of feasible geometries varies significantly across these settings, meaning the baseline choices of ±5\% lift tolerance, 95th percentile load limits, global 95\% interval, and the support limit should be interpreted as the declared study parameter configuration.

\section{Conclusions}
\label{sec:conclusions}

A physics-structured surrogate model and conformal robust multipoint optimization framework was created for the preliminary design of glider wings. The framework utilized a 150,000-case Tornado database, a five-member dual-headed neural ensemble, exact physical decoding, normalized simultaneous conformal intervals, an explicit training-support constraint, and a three-objective search over three velocities.

The structured ensemble managed to achieve a mean NRMSEs value of 0.0223 on the in-distribution test set and 0.0595 for the structured OOD set, and outperformed the ridge regression, LightGBM, unrestricted compact MLP, and a single structured network in the aggregate benchmark. The dispersion of the raw ensemble was significantly undercalibrated for simultaneous decisions. Global 95\% max-score conformal calibration yielded 94.71\% in-distribution joint coverage, and support-condition calibration raised structured-OOD joint coverage up to 91.08\%.

The robust search filtered 2,998 of 16,384 continuous geometries after applying all hard conditions and identified 198 nondominated designs. The Pareto set captured the anticipated trade-off between induced-like drag and aerodynamic root bending and highlighted the existence of a  convention-dependent direction derivative compromise. Twenty candidate wings were frozen for further solver evaluation.

The 60 Tornado simulations served as a direct post-selection validation for the entire procedure. The mean target NRMSE was 0.0228, the global 95\% interval covered all 14 optimization quantities simultaneously in 58/60 cases, and all 60 operating points satisfied every single hard constraint. All 20 mission-weighted drag, worst-bending, and directional-deficit upper bounds were conservative, and 19 out of 20 designs remained nondominated among those tested.

The presented study demonstrates the effectiveness of physics-structured surrogates and conformal robust optimization for an efficient, uncertainty-aware screening of continuous-wing planforms.

\FloatBarrier

\makeatletter
\renewcommand*{\theHequation}{\thesection.\arabic{equation}}
\makeatother

\begin{appendices}

\section{Simulation configuration and dimensional flight-dynamics derivations}
\label{app:solver_configuration}

\subsection{Solver and fixed-aircraft configuration}

The Tornado solver receives input of wing geometry and the operating condition defined in Eq.~\eqref{eq:geometry_state_vectors}. Angles are stored in degrees in the data files and then converted via the simulation interface whenever required. The full span is $B=2b$; the tip chord is $c_t=\lambda c_r$; the reference area and MAC are given by Eq.~\eqref{eq:derived_geometry_main}. The VLM results contain integrated coefficients and coefficient slopes with respect to angle, sideslip, and nondimensional body rates.

\begin{table}[h]
\centering
\caption{Frozen aerodynamic-solver settings.}
\label{tab:solver_settings_appendix}
\begin{tabular}{ll}
\toprule
Setting & Frozen value \\
\midrule
Lifting-surface solver & Tornado vortex-lattice method \\
Airfoil section & NACA~2412 over the complete span \\
Lattice resolution & $20\times40$ chordwise--spanwise panels \\
Reference altitude & 1200~m, International Standard Atmosphere \\
Recovered density & $1.089960964052~\mathrm{kg\,m^{-3}}$ \\
Sweep input & Quarter-chord sweep angle $\Lambda$ \\
Span convention & $b$ is semi-span; $B=2b$ is full span \\
Angle storage & Degrees in input files \\
\bottomrule
\end{tabular}
\end{table}

The fixed-wing derivative post-processing used the fuselage and tail configuration is listed in Table~\ref{tab:fixed_aircraft_appendix}. These values remained fixed for all of the $150{,}000$ cases. They represent a common reference aircraft design, around which the wing design varies; they are not part of the optimization problem and are not claimed to be optimal for each individual wing design.

\begin{table}[h]
\centering
\caption{Fixed geometry and mass parameters used in the flight-dynamics post-processing.}
\label{tab:fixed_aircraft_appendix}
\begin{tabular}{lclc}
\toprule
Parameter & Value & Parameter & Value \\
\midrule
Fuselage length [m] & 6.00 & Fuselage diameter [m] & 0.50 \\
Horizontal-tail root chord [m] & 1.95 & Horizontal-tail tip chord [m] & 0.60 \\
Horizontal-tail span [m] & 1.00 & Vertical-tail root chord [m] & 1.95 \\
Vertical-tail tip chord [m] & 1.00 & Vertical-tail span [m] & 1.30 \\
Wing position $x$ [m] & 1.75 & Wing position $z$ [m] & 1.90 \\
Horizontal-tail position $x$ [m] & 4.05 & Horizontal-tail position $z$ [m] & 1.90 \\
Vertical-tail position $x$ [m] & 4.05 & Vertical-tail position $z$ [m] & 1.90 \\
Fuselage mass per unit length [kg/m] & 1.00 & Tail mass per unit area [kg/m$^2$] & 10.00 \\
\bottomrule
\end{tabular}
\end{table}

\subsection{Body-axis loads and perturbation variables}

Let $V_\infty$ denote the unperturbed flight speed and let
\begin{equation}
Q=\frac{1}{2}\rho V_\infty^2
\label{eq:dynamic_pressure_derivation}
\end{equation}
denote dynamic pressure in this appendix. The symbol $Q$ is used here to distinguish the quantity from the pitch rate; they denote the same quantity as $q$ in the main text. Let $\mathbf C_F=[C_X,C_Y,C_Z]^{\mathsf T}$ and $\mathbf C_M=[C_\ell,C_m,C_n]^{\mathsf T}$ denote the total fixed-aircraft body-axis coefficient vectors after the VLM wing slopes have been assigned to the MATLAB fixed-wing model. Their dimensionalization is
\begin{equation}
\begin{bmatrix}X\\Y\\Z\end{bmatrix}=QS\mathbf C_F,
\qquad
\begin{bmatrix}L\\M\\N\end{bmatrix}
=QS
\begin{bmatrix}
B C_\ell\\ \bar c C_m\\ B C_n
\end{bmatrix},
\label{eq:body_load_dimensionalization}
\end{equation}
where $B=2b$ is the full span. The superscript ``FW'' used below denotes derivatives of these total fixed-aircraft coefficients. They were not assumed to be equal to the wing-only Tornado column values, as the fixed fuselage and tail contributions are included in the downstream post-processing.

Let $(U,v_b,w_b)$ denote the body-axis translational velocity components, with the reference state $(U,v_b,w_b)=(V_\infty,0,0)$. Locally,
\begin{equation}
\alpha=\tan^{-1}\!\left(\frac{w_b}{U}\right),
\qquad
\beta=\sin^{-1}\!\left(\frac{v_b}{\sqrt{U^2+v_b^2+w_b^2}}\right).
\label{eq:body_velocity_angles}
\end{equation}
Therefore, at the reference state and with $U$ held fixed,
\begin{equation}
\left.\frac{\partial\alpha}{\partial w_b}\right|_0=\frac{1}{V_\infty},
\qquad
\left.\frac{\partial\beta}{\partial v_b}\right|_0=\frac{1}{V_\infty},
\qquad
\left.\frac{\partial Q}{\partial v_b}\right|_0=
\left.\frac{\partial Q}{\partial w_b}\right|_0=0.
\label{eq:velocity_angle_chain_factors}
\end{equation}

For the nondimensional body-rate variables,
\begin{equation}
\widehat p=\frac{pB}{2V_\infty},
\qquad
\widehat q=\frac{q_{\rm pitch}\bar c}{2V_\infty},
\qquad
\widehat r=\frac{rB}{2V_\infty},
\label{eq:nondimensional_rates_appendix}
\end{equation}
where $q_{\rm pitch}$ separates pitch rate from dynamic pressure.

\subsection{Derivation of the seven retained dimensional derivatives}
\label{app:retained_dynamics_derivation}

The dimensionalizations based on the chain rule obtained here are consistent with conventional linearized derivatives for fixed-wing aircraft, as described in \cite{roskam1995flightdynamics}. The slopes of aerodynamic coefficients for the stationary aircraft are given with angular quantities expressed in radians by
\begin{equation}
C_{Y_\beta}^{\rm FW}=\frac{\partial C_Y^{\rm FW}}{\partial\beta},\quad
C_{Z_\alpha}^{\rm FW}=\frac{\partial C_Z^{\rm FW}}{\partial\alpha},\quad
C_{m_\alpha}^{\rm FW}=\frac{\partial C_m^{\rm FW}}{\partial\alpha},
\label{eq:fixed_wing_coefficient_slopes_1}
\end{equation}
\begin{equation}
C_{\ell_\beta}^{\rm FW}=\frac{\partial C_\ell^{\rm FW}}{\partial\beta},\quad
C_{n_\beta}^{\rm FW}=\frac{\partial C_n^{\rm FW}}{\partial\beta},\quad
C_{\ell_{\widehat p}}^{\rm FW}=\frac{\partial C_\ell^{\rm FW}}{\partial\widehat p},\quad
C_{n_{\widehat r}}^{\rm FW}=\frac{\partial C_n^{\rm FW}}{\partial\widehat r}.
\label{eq:fixed_wing_coefficient_slopes_2}
\end{equation}
Degree-valued input angles should be converted to radians before these derivatives take effect. Note that if a coefficient slope is quoted per degree, it should be multiplied by $180/\pi$ to express the value per radian as defined above.

Using Eqs.~\eqref{eq:body_load_dimensionalization} and \eqref{eq:velocity_angle_chain_factors}, the two translational-velocity derivatives may be derived via the chain rule:
\begin{align}
F_{YV}
&\equiv \left.\frac{\partial Y}{\partial v_b}\right|_0
 =QS\,C_{Y_\beta}^{\rm FW}
 \left.\frac{\partial\beta}{\partial v_b}\right|_0
 =\frac{QS}{V_\infty}C_{Y_\beta}^{\rm FW},
\label{eq:FYV_derivation}\\
F_{ZW}
&\equiv \left.\frac{\partial Z}{\partial w_b}\right|_0
 =QS\,C_{Z_\alpha}^{\rm FW}
 \left.\frac{\partial\alpha}{\partial w_b}\right|_0
 =\frac{QS}{V_\infty}C_{Z_\alpha}^{\rm FW}.
\label{eq:FZW_derivation}
\end{align}
These derivatives have units of $\mathrm{N\,m\,s\,rad^{-1}}$. The new surrogate model does not learn a separate dimensional $M_q$ target; it instead uses the coefficient-level slope  $C_{m_q}\equiv C_{m_{\widehat q}}$.

Regarding the angular static derivatives, $Q$, $S$, $B$, and $\bar c$, they are constant with respect to the angular perturbation; thus,
\begin{align}
M_\alpha
&\equiv\frac{\partial M}{\partial\alpha}
 =QS\bar c\,C_{m_\alpha}^{\rm FW},
\label{eq:Malpha_derivation}\\
L_\beta
&\equiv\frac{\partial L}{\partial\beta}
 =QSB\,C_{\ell_\beta}^{\rm FW},
\label{eq:Lbeta_derivation}\\
N_\beta
&\equiv\frac{\partial N}{\partial\beta}
 =QSB\,C_{n_\beta}^{\rm FW}.
\label{eq:Nbeta_derivation}
\end{align}
Their units are $\mathrm{N\,m\,rad^{-1}}$. The radian is a dimensionless quantity in the SI system; however, "per radian" is used for the sake of differentiating variables.

Finally, using the chain rule on Eq.~\eqref{eq:nondimensional_rates_appendix} leads to
\begin{align}
L_p
&\equiv\frac{\partial L}{\partial p}
 =QSB\,C_{\ell_{\widehat p}}^{\rm FW}
 \frac{\partial\widehat p}{\partial p}
 =QSB\left(\frac{B}{2V_\infty}\right)C_{\ell_{\widehat p}}^{\rm FW},
\label{eq:Lp_derivation}\\
N_r
&\equiv\frac{\partial N}{\partial r}
 =QSB\,C_{n_{\widehat r}}^{\rm FW}
 \frac{\partial\widehat r}{\partial r}
 =QSB\left(\frac{B}{2V_\infty}\right)C_{n_{\widehat r}}^{\rm FW}.
\label{eq:Nr_derivation}
\end{align}
with units $\mathrm{N\,m\,s\,rad^{-1}}$ . The updated surrogate does not learn a separate dimensional $M_q$ target, only the coefficient-level slope $C_{m_q}\equiv C_{m_{\widehat q}}$ .

\begin{table}[h]
\centering
\caption{Construction of dimensions, native unit, and speed scaling of the seven fixed-wing derivatives. Speed scaling assumes fixed geometry, density, and coefficient slopes.}
\label{tab:retained_dynamics_derivation}
\small
\begin{tabular}{llll}
\toprule
Target & Dimensional construction & Native unit & Explicit scaling \\
\midrule
$F_{YV}$ & $(QS/V_\infty)C_{Y_\beta}^{\rm FW}$ & $\mathrm{N\,s\,m^{-1}}$ & $V_\infty$ \\
$F_{ZW}$ & $(QS/V_\infty)C_{Z_\alpha}^{\rm FW}$ & $\mathrm{N\,s\,m^{-1}}$ & $V_\infty$ \\
$M_\alpha$ & $QS\bar c C_{m_\alpha}^{\rm FW}$ & $\mathrm{N\,m\,rad^{-1}}$ & $V_\infty^2$ \\
$L_\beta$ & $QSB C_{\ell_\beta}^{\rm FW}$ & $\mathrm{N\,m\,rad^{-1}}$ & $V_\infty^2$ \\
$N_\beta$ & $QSB C_{n_\beta}^{\rm FW}$ & $\mathrm{N\,m\,rad^{-1}}$ & $V_\infty^2$ \\
$L_p$ & $QSB(B/2V_\infty)C_{\ell_{\widehat p}}^{\rm FW}$ & $\mathrm{N\,m\,s\,rad^{-1}}$ & $V_\infty$ \\
$N_r$ & $QSB(B/2V_\infty)C_{n_{\widehat r}}^{\rm FW}$ & $\mathrm{N\,m\,s\,rad^{-1}}$ & $V_\infty$ \\
\bottomrule
\end{tabular}
\end{table}

Because $Q\propto V_\infty^2$, Eqs.~\eqref{eq:FYV_derivation}--\eqref{eq:Nr_derivation} imply the two speed-scaling families
\begin{equation}
\{F_{YV},F_{ZW},L_p,N_r\}\propto V_\infty,
\qquad
\{M_\alpha,L_\beta,N_\beta\}\propto V_\infty^2.
\label{eq:derived_dynamics_speed_families}
\end{equation}

These equations determine magnitudes and units when the total fixed aircraft coefficient slope values are specified. The signs are consistent with those of the corresponding coefficient slopes adopted under MATLAB body-axis convention.

\section{Surrogate architecture, loss, and model-selection metrics}
\label{app:surrogate_math}

Let $\boldsymbol\mu_x$ and $\boldsymbol\sigma_x$ represent the means and standard deviation estimated from the 73,000 training samples. The inputs for all models use
\begin{equation}
\widetilde x_{ij}
=
\frac{x_{ij}-\mu_{x,j}}{\sigma_{x,j}},
\qquad
\widetilde y_{ij}
=
\frac{y_{ij}-\mu_{y,j}}{\sigma_{y,j}},
\label{eq:standardization_appendix}
\end{equation}
where the target statistics are also derived from training data alone. The compact baseline uses
\begin{equation}
\widehat{\widetilde{\mathbf y}}_{\rm compact}
=f_{\rm compact}(\widetilde{\mathbf x}_{\rm full}),
\label{eq:compact_network}
\end{equation}
with the layer widths $12$--$128$--$128$--$64$--$16$. for the ensemble member $m$, the two heads are
\begin{equation}
\widehat{\widetilde{\mathbf y}}_{A,m}
=f_{A,m}(\widetilde{\mathbf x}_{\rm sim}),
\qquad
\widehat{\widetilde{\mathbf y}}_{D,m}
=f_{D,m}(\widetilde{\mathbf x}_{\rm full}),
\label{eq:dual_head_networks}
\end{equation}
with widths $6$--$128$--$128$--$64$--$9$ and $12$--$128$--$128$--$64$--$7$, respectively. The activation is
\begin{equation}
\operatorname{SiLU}(u)=u\left(1+e^{-u}\right)^{-1}.
\label{eq:silu_appendix}
\end{equation}
Layer normalization occurs after each of the first two hidden activations. The output layers are linear.

Given standardized residual $e$, the smoothed $L_1$ cost function is
\begin{equation}
\rho_{\delta}(e)
=
\begin{cases}
\dfrac{e^2}{2\delta}, & |e|<\delta,\\[4pt]
|e|-\dfrac{\delta}{2}, & |e|\ge\delta,
\end{cases}
\qquad \delta=0.5.
\label{eq:smooth_l1_appendix}
\end{equation}
In the mini-batch $\mathcal B$, for observation $i$ and member $m$, suppose $w_{im}\sim \text{Poisson}(1)$. With aerodynamic/load target set $\mathcal A$ and dynamic target set $\mathcal D$, the member loss is
\begin{equation}
\mathcal L_m
=
\frac{
\displaystyle\sum_{i\in\mathcal B}w_{im}
\left[
\frac{1}{|\mathcal A|}\sum_{j\in\mathcal A}\rho_\delta(e_{ijm})
+
\frac{1}{|\mathcal D|}\sum_{j\in\mathcal D}\rho_\delta(e_{ijm})
\right]
}{
\displaystyle\sum_{i\in\mathcal B}w_{im}+10^{-8}
}.
\label{eq:member_loss_appendix}
\end{equation}
The joint training loss is
\begin{equation}
\mathcal L
=
\mathcal L_{\rm compact}
+
\frac{1}{5}\sum_{m=1}^{5}\mathcal L_m.
\label{eq:joint_training_loss_appendix}
\end{equation}
Since compact and dual-head networks have separate parameters, simultaneous training uses the same data order and optimizer configurations but not the hidden activations or gradients between models.

After the inverse transformation of targets, the ensemble estimators are
\begin{equation}
\widehat\mu_j(\mathbf x)
=
\frac{1}{M}\sum_{m=1}^{M}\widehat y_{jm}(\mathbf x),
\qquad
\widehat\sigma_j(\mathbf x)
=
\sqrt{
\frac{1}{M-1}\sum_{m=1}^{M}
\left(\widehat y_{jm}(\mathbf x)-\widehat\mu_j(\mathbf x)\right)^2
},
\label{eq:ensemble_estimators_appendix}
\end{equation}
with $M=5$. The dispersion in Eq.~\eqref{eq:ensemble_estimators_appendix} is a disagreement statistic and not a calibrated predictive standard deviation.

For target $j$, the normalized validation error is defined by
\begin{equation}
\mathrm{NRMSE}_j
=
\frac{
\sqrt{n^{-1}\sum_{i=1}^{n}(y_{ij}-\widehat y_{ij})^2}
}{s_{j,\rm train}},
\label{eq:nrmse_appendix}
\end{equation}
where $s_{j,\rm train}$ denotes the standard deviation.For sign-sensitive target $j$, define the boundary subset
\begin{equation}
\mathcal B_j
=
\left\{
 i:\ |y_{ij}|\le Q_{0.10}^{\rm train}(|Y_j|)
\right\}.
\label{eq:boundary_subset_appendix}
\end{equation}
Boundary sign accuracy is averaged over
\begin{equation}
\mathcal J_{\rm sign}
=
\{C_{m_\alpha},C_{\ell_p},C_{m_q},F_{YV},F_{ZW},M_\alpha,N_\beta,L_p,N_r\}.
\label{eq:sign_target_set_appendix}
\end{equation}
Rank quality is
\begin{equation}
R_{\rm rank}
=
\frac{1}{2}
\left[
\rho_S(C_D,\widehat C_D)
+
\rho_S(|C_{\rm bend}|,|\widehat C_{\rm bend}|)
\right],
\label{eq:rank_quality_appendix}
\end{equation}
where $\rho_S$ is the Spearman correlation. Plugging this expression into Eq.~\eqref{eq:model_selection_score_main} gives the frozen composite score.

The ridge baseline implemented standardized inputs and outputs with a regularization coefficient of $10^{-4}$. Sixteen independent LightGBM regressors used 150 trees, with a learning rate of 0.05, 31 leaves, minimum child count of 20, feature subsample of 0.9, and $\ell_2$ regularization of $10^{-4}$. The compact MLP consisted of 27,984 trainable parameters. The neural network had a batch size of 8192 on CUDA or 4096 on CPU, AdamW learning rate of $2\times10^{-3}$, weight decay of $10^{-5}$, and 12 epochs.

\section{Geometric identities and dimensional decoder}
\label{app:geometry_decoder}

For half of a linearly-tapered symmetric wing, the chord is
\begin{equation}
c(y)=c_r-\frac{c_r-c_t}{b}y
=c_r\left[1-(1-\lambda)\frac{y}{b}\right],
\qquad 0\le y\le b.
\label{eq:local_chord}
\end{equation}
Integrating over both half-wings gives
\begin{equation}
S=2\int_0^b c(y)\,\mathrm{d}y
=b(c_r+c_t)
=b c_r(1+\lambda),
\label{eq:area_derivation}
\end{equation}
which results in the derivation of Eq.~\eqref{eq:derived_geometry_main}. The wing span is $B = 2b$; therefore,
\begin{equation}
AR=\frac{B^2}{S}=\frac{4b^2}{b c_r(1+\lambda)}
=\frac{4b}{c_r(1+\lambda)}.
\label{eq:ar_derivation}
\end{equation}
Mean aerodynamic chord for a trapezoidal planform is derived following the second chord moment,
\begin{equation}
\bar c
=
\frac{2}{S}\int_0^b c^2(y)\,\mathrm{d}y
=
\frac{2}{3}c_r\frac{1+\lambda+\lambda^2}{1+\lambda}.
\label{eq:mac_derivation}
\end{equation}

The three load coefficients are defined by
\begin{equation}
C_{\mathrm{shear}}
=\frac{F_{\mathrm{root}}}{qS},
\qquad
C_{\mathrm{bend}}
=\frac{M_{\mathrm{bend}}}{qSB},
\qquad
C_{\mathrm{twist}}
=\frac{M_{\mathrm{twist}}}{qS\bar c},
\label{eq:load_coefficient_definitions}
\end{equation}
in which $B = 2b$. Thus, any lower and upper coefficient bounds
$[C_j^{\mathrm L},C_j^{\mathrm U}]$ can be decoded without resorting to an approximation, as the multiplicative factors $qS$, $qSB$, and $qS\bar c$ are known for each candidate. For a quantity that can switch signs, the conservative absolute bound is
\begin{equation}
|Q|^{\mathrm U}
=K\max\left(|C^{\mathrm L}|,|C^{\mathrm U}|\right),
\label{eq:absolute_decoded_bound}
\end{equation}
where $K$ is the respective dimensional factor. This expression is used for root shear, bending, and twisting constraints or objectives.

\section{Simultaneous conformal calibration}
\label{app:conformal_math}

Let $\widehat\mu_j(\mathbf{x})$ and $\widehat\sigma_j(\mathbf{x})$ be the ensemble mean and member standard deviation for the $j$ output. A target-specific floor $\epsilon_j>0$ is chosen in each case to avoid an unrealistically low scale in case of total agreement among the ensemble members. For calibration observation $i$ and target $j$, the normalized nonconformity score is
\begin{equation}
r_{ij}
=
\frac{|y_{ij}-\widehat\mu_j(\mathbf{x}_i)|}
{\widehat\sigma_j(\mathbf{x}_i)+\epsilon_j}.
\label{eq:normalized_nonconformity}
\end{equation}
For the set $\mathcal J_{\mathrm{opt}}$ of 14 outputs entering the optimization, we define a simultaneous score by
\begin{equation}
R_i=\max_{j\in\mathcal J_{\mathrm{opt}}}r_{ij}.
\label{eq:joint_nonconformity}
\end{equation}
When there are $n_{\mathrm{cal}}$ calibration scores, the finite-sample split-conformal quantile  is the order statistic
\begin{equation}
\widehat q_{1-\alpha}
=R_{(k)},
\qquad
k=\left\lceil(n_{\mathrm{cal}}+1)(1-\alpha)\right\rceil,
\label{eq:conformal_quantile}
\end{equation}
with the index capped at $n_{\mathrm{cal}}$ if necessary. The corresponding rectangular joint region is
\begin{equation}
\mathcal C_{1-\alpha}(\mathbf{x})
=
\prod_{j\in\mathcal J_{\mathrm{opt}}}
\left[
\widehat\mu_j(\mathbf{x})
-\widehat q_{1-\alpha}\left(\widehat\sigma_j(\mathbf{x})+\epsilon_j\right),
\widehat\mu_j(\mathbf{x})
+\widehat q_{1-\alpha}\left(\widehat\sigma_j(\mathbf{x})+\epsilon_j\right)
\right].
\label{eq:joint_conformal_region}
\end{equation}
Under exchangeability of calibration and future observations, the  max-score construction gives marginal finite-sample coverage for the whole output vector,
\begin{equation}
\Pr\left\{
\mathbf{Y}_{\mathrm{new}}
\in
\mathcal C_{1-\alpha}(\mathbf{X}_{\mathrm{new}})
\right\}
\ge 1-\alpha.
\label{eq:joint_coverage_statement}
\end{equation}
The guarantee is marginal for the future example and does not entail conditional coverage at all geometries. In terms of OOD diagnostics, the standardized distance to the $K=20$ nearest training points is
\begin{equation}
d_{20}(\mathbf{x})
=
\frac{1}{20}\sum_{m=1}^{20}
\left\|
\widetilde{\mathbf{x}}-
\widetilde{\mathbf{x}}_{(m)}
\right\|_2,
\label{eq:support_metric}
\end{equation}
where tilde denotes feature standardization, and $(m)$ denotes nearest neighbors. Calibration observations are stratified into support-score quantiles, and Eq.~\eqref{eq:conformal_quantile} is applied within each support bin to construct support-Mondrian diagnostic intervals.
There are three supporrt thresholds: 50th, 80th, and 95th percentiles. They are 1.195406, 1.283027, and 1.381042 for this particular experiment. The floor of the normalized scale is
\begin{equation}
\epsilon_j
=
\max\left
\{
Q_{0.10}\!\left(\widehat\sigma_j\right),
0.01\,s_j^{\mathrm{train}}
\right\},
\label{eq:sigma_floor_appendix}
\end{equation}
where $s_j^{\mathrm{train}}$ denotes  training-target standard deviation. The quantiles for global optimizations are 2.71379 and 3.33918 for 90\% and 95\% coverage, respectively. The corresponding 95\% support-bin quantiles are 2.85234, 3.34180, 3.92571, and 5.96246 from the best-supported to the most distant bin.

\section{Robust multipoint optimization statement}
\label{app:optimization_math}

Let the notation $\mathbf g=(c_r,\Lambda,b,\theta,\lambda,\Gamma)$ refer to geometry and let the three operating speeds be denoted by $V_k$, $k\in\{1,2,3\}$. The ensemble-mean lift is evaluated at the angular bounds for each of the pairs $(\mathbf g,V_k)$. A geometry is trim-reachable when $W$ is within those mean values. For reachable points, bisection generates $\alpha_k(\mathbf g)$ from
\begin{equation}
\widehat L\!\left(\mathbf g,\alpha_k,V_k\right)=W.
\label{eq:mean_trim_root}
\end{equation}
Bisection stops after 15 iterations. Equation~\eqref{eq:mean_trim_root} is a root location device, and not the feasibility statement.

Let $Q_k^{\mathrm L}(\mathbf g)$ and $Q_k^{\mathrm U}(\mathbf g)$ be the decoded global 95\% conformal bounds at the trimmed operating point. Robust Lift feasibility requires:
\begin{equation}
\max\left\{
|L_k^{\mathrm L}-W|,
|L_k^{\mathrm U}-W|
\right\}
\leq \tau_L W,
\qquad \tau_L=0.05,
\quad k=1,2,3.
\label{eq:robust_lift_constraint}
\end{equation}
The derivative constraints are
\begin{equation}
C_{m_\alpha,k}^{\mathrm U}<0,
\qquad
C_{\ell_p,k}^{\mathrm U}<0,
\qquad
C_{m_q,k}^{\mathrm U}<0,
\qquad k=1,2,3.
\label{eq:hard_sign_constraints}
\end{equation}
The support and load conditions are
\begin{align}
 d_{20,k}(\mathbf g)&\leq 1.38104168,\\
 \max\{|C_{\mathrm{shear},k}^{\mathrm L}|,|C_{\mathrm{shear},k}^{\mathrm U}|\}&\leq0.46592119,\\
 \max\{|C_{\mathrm{twist},k}^{\mathrm L}|,|C_{\mathrm{twist},k}^{\mathrm U}|\}&\leq0.002207061,
\label{eq:support_load_constraints}
\end{align}
for all three speeds.

The first objective is mission-weighted conservative drag,
\begin{equation}
J_D(\mathbf g)
=
\sum_{k=1}^{3}w_k D_k^{\mathrm U}(\mathbf g),
\qquad
(w_1,w_2,w_3)=(0.25,0.50,0.25).
\label{eq:drag_objective}
\end{equation}
The second objective is the largest conservative root-bending magnitude,
\begin{equation}
J_B(\mathbf g)
=
\max_{k=1,2,3}
\left[
q_kS(2b)
\max\{|C_{\mathrm{bend},k}^{\mathrm L}|,|C_{\mathrm{bend},k}^{\mathrm U}|\}
\right].
\label{eq:bending_objective}
\end{equation}
The third objective is the worst normalized directional deficit,
\begin{equation}
J_{\mathrm{dir}}(\mathbf g)
=
\max_{k=1,2,3}
\left\{
0,
-\frac{N_{\beta,k}^{\mathrm L}}{s_{N_\beta}},
\frac{N_{r,k}^{\mathrm U}}{s_{N_r}}
\right\},
\label{eq:directional_objective}
\end{equation}
where $s_{N_\beta}=331.510327996717$ and $s_{N_r}=44.1025032240466$ are training-partition median absolute values. Therefore, only if the entire certified interval is consistent with the adopted desired signs at every operating condition can the zero value be achieved.

The robust multiobjective optimization problem is:
\begin{equation}
\begin{aligned}
\min_{\mathbf g\in\mathcal G}\quad
&\left(J_D(\mathbf g),J_B(\mathbf g),J_{\mathrm{dir}}(\mathbf g)\right),\\
\text{subject to}\quad
&\text{Eqs.~\eqref{eq:robust_lift_constraint}, \eqref{eq:hard_sign_constraints}, and \eqref{eq:support_load_constraints}.}
\end{aligned}
\label{eq:full_robust_problem}
\end{equation}
In place of differentiation through trim, nearest-neighbor, conformal intervals, and nondominated sorting \cite{sobol1967distribution}, a randomized Sobol design of size $2^{14}$ is used.

Consider $\mathbf J_i = (J_{D,i}, J_{B,i}, J_{\text{dir}, i})$. The candidate $i$ is said to be dominated if some other feasible candidate $h$ fulfills $J_{h,m} \leq J_{i,m}$ for all objectives $m$ and is smaller than candidate $i$ for at least one objective. The Pareto include all candidates that are nondominated. To choose $K=20$ representatives, every objective is scaled throughout the Pareto set,
\begin{equation}
\widetilde J_{i,m}
=
\frac{J_{i,m}-\min_h J_{h,m}}
{\max_h J_{h,m}-\min_h J_{h,m}}.
\label{eq:objective_normalization_selection}
\end{equation}
The three objective minima are inserted first. The balanced knee is
\begin{equation}
i_{\mathrm{knee}}=\arg\min_i\|\widetilde{\mathbf J}_i\|_2.
\label{eq:knee_selection}
\end{equation}
Other remaining representatives are then greedily included by maximizing distance to the current chosen set,
\begin{equation}
i_{t+1}
=
\arg\max_{i\notin\mathcal S_t}
\min_{h\in\mathcal S_t}
\|\widetilde{\mathbf J}_i-\widetilde{\mathbf J}_h\|_2.
\label{eq:farthest_point_selection}
\end{equation}
This step maintains visible coverage of the three-objective front without affecting Pareto rank.

\section{External-validation protocol and metrics}
\label{app:external_validation_metrics}

The external-validation set consists of $n=60$ new Tornado operating points, and the full learned output vector has $p=16$ components. Point-prediction accuracy for each target $j$ is reported using
\begin{align}
\operatorname{MAE}_j
&=\frac{1}{n}\sum_{i=1}^n|\widehat y_{ij}-y_{ij}|,\\
\operatorname{RMSE}_j
&=\left[\frac{1}{n}\sum_{i=1}^n(\widehat y_{ij}-y_{ij})^2\right]^{1/2},\\
\operatorname{NRMSE}_j
&=\frac{\operatorname{RMSE}_j}{\operatorname{sd}(y_{1j},\ldots,y_{nj})}.
\label{eq:external_metrics}
\end{align}
For interval method $a$ and optimization-target index set $\mathcal J_{\rm opt}$, empirical joint coverage is calculated as
\begin{equation}
\widehat C_{\rm joint}^{(a)}
=\frac{1}{n}\sum_{i=1}^n
\mathbf 1\!\left[
 y_{ij}\in[\ell_{ij}^{(a)},u_{ij}^{(a)}]
 \ \text{for all }j\in\mathcal J_{\rm opt}
\right].
\label{eq:external_joint_coverage_metric}
\end{equation}
The 60 operating points were chosen adaptively during the optimization procedure and are cannot be assumed exchangeable with the calibration partition. For this reason, Equation~\eqref{eq:external_joint_coverage_metric} is to be interpreted as an empirical post-selection validation statistic rather than as a new finite-sample conformal guarantee.

For a design-level minimized objective $J_m$ with frozen upper bound $J_m^{U}$, the bound is deemed successful when
\begin{equation}
J_m^{\rm Tornado}\leq J_m^{U}.
\label{eq:objective_bound_success}
\end{equation}
A design is considered to be externally feasible only if all three operating points meet all solver-dependent hard constraints using the newly simulated values. The design should also satisfy the input support condition that was frozen prior to the simulation. Nondominance is then recomputed only for the fixed set of 20 verified designs and does not imply global Pareto optimality on the continuous space.

\begin{table}[p]
\centering
\caption{Target-wise point-prediction accuracy on the 60 new Tornado
operating points. MAE and RMSE use the native units shown; NRMSE and
$R^2$ are dimensionless.}
\label{tab:external_target_metrics_appendix}
\scriptsize
\setlength{\tabcolsep}{3.2pt}
\begin{tabular}{llrrrr}
\toprule
Target & Unit & MAE & RMSE & NRMSE & $R^2$ \\
\midrule
$C_L$ & -- & 0.003217 & 0.004000 & 0.0235 & 0.99945 \\
$C_D$ & -- & $7.90\times10^{-5}$ & $1.61\times10^{-4}$ & 0.0357 & 0.99873 \\
$C_m$ & -- & 0.001391 & 0.002019 & 0.0166 & 0.99973 \\
$C_{m_\alpha}$ & rad$^{-1}$ & 0.009765 & 0.013141 & 0.0158 & 0.99975 \\
$C_{\ell_p}$ & -- & $7.67\times10^{-4}$ & 0.001211 & 0.0257 & 0.99934 \\
$C_{m_q}$ & -- & 0.025950 & 0.040437 & 0.0147 & 0.99978 \\
$F_{YV}$ & N s m$^{-1}$ & 0.05251 & 0.06644 & 0.0243 & 0.99941 \\
$F_{ZW}$ & N s m$^{-1}$ & 5.442 & 6.941 & 0.0241 & 0.99942 \\
$M_\alpha$ & N m rad$^{-1}$ & 1049.8 & 1360.5 & 0.0188 & 0.99964 \\
$L_\beta$ & N m rad$^{-1}$ & 76.12 & 93.13 & 0.0135 & 0.99982 \\
$N_\beta$ & N m rad$^{-1}$ & 9.306 & 11.616 & 0.0261 & 0.99932 \\
$L_p$ & N m s rad$^{-1}$ & 137.56 & 175.63 & 0.0156 & 0.99976 \\
$N_r$ & N m s rad$^{-1}$ & 0.7333 & 0.9586 & 0.0310 & 0.99904 \\
$C_{\mathrm{shear}}$ & -- & 0.002447 & 0.002862 & 0.0339 & 0.99885 \\
$C_{\mathrm{bend}}$ & -- & $4.25\times10^{-4}$ & $5.18\times10^{-4}$ & 0.0284 & 0.99920 \\
$C_{\mathrm{twist}}$ & -- & $7.52\times10^{-6}$ & $9.43\times10^{-6}$ & 0.0175 & 0.99970 \\
\bottomrule
\end{tabular}
\end{table}

Simulator inputs, row identifiers, trained-network predictions, ensemble dispersions, conformal bounds, realized outputs, constraint indicators, and candidate-level objective comparisons are available from the corresponding author upon reasonable request. The external results were kept read-only throughout the evaluation.

\section{Frozen-model ablations and optimization sensitivity}
\label{app:sensitivity}

The entire analyses peresented in this appendix are based on the previously frozen training partitions, trained models, conformal parameters, Sobol geometries, and trim procedure. No model was retrained, no conformal quantile was recalibrated, and none of the 60 post-selection Tornado outputs were used. The results should therefore be interpreted as surrogate-level assessments of methodological and threshold sensitivity. They do not provide new external validation of another design set.

\subsection{Architecture and uncertainty ablations}

The structured representation and ensemble averaging contribute separate improvements. The single structured model is better than the unrestricted compact MLP in terms of mean NRMSE for both ID and structured-OOD test sets. Averaging the predictions of five independently initialized Poisson-bootstrap members lowers the error further.

\begin{table}[h]
\centering
\caption{Frozen architecture ablation. Lower NRMSE is better.}
\label{tab:architecture_ablation_appendix}
\begin{tabular}{lcc}
\toprule
Model & ID NRMSE & Structured-OOD NRMSE \\
\midrule
Compact unrestricted MLP & 0.03690 & 0.07670 \\
Physics-structured single member & 0.02692 & 0.06555 \\
Physics-structured five-member ensemble & \textbf{0.02234} & \textbf{0.05951} \\
\bottomrule
\end{tabular}
\end{table}

The uncertainty ablation brings us to a similar conclusion. y itself, ensemble dispersion does not deliver a simultaneous 95\% interval. Marginal target-wise calibration improves coverage for individual outputs, but it is insufficient in case of a decision involving 14 outputs simultaneously. Global max-score calibration provides the formal optimization certificate, while support conditioning primarily helps to improve the diagnostic result on the structured-OOD set.

\begin{table}[h]
\centering
\caption{Joint coverage across the 14 optimization outputs at 95\% nominal coverage.}
\label{tab:uncertainty_ablation_appendix}
\begin{tabular}{lcc}
\toprule
Interval method & ID joint coverage [\%] & Structured-OOD joint coverage [\%] \\
\midrule
Raw ensemble spread & 17.51 & 11.68 \\
Marginal normalized conformal & 58.29 & 27.81 \\
Global joint conformal & 94.71 & 73.81 \\
Support-Mondrian conformal & 94.97 & 91.08 \\
\bottomrule
\end{tabular}
\end{table}

\subsection{Optimization-setting sensitivity}

Table~\ref{tab:optimization_sensitivity_appendix} summarizes the numbers of the all-hard-feasible geometries and nondominated designs that are generated for alternative frozen settings. The baseline corresponds to the global joint 95\% configuration with $\pm5\%$ lift tolerance, 95th-percentile root-shear and root-twist thresholds, and the calibration 95th-percentile support limit.

\begin{table}[h]
\centering
\caption{Frozen-surrogate optimization sensitivity over the same 16,384 Sobol geometries.}
\label{tab:optimization_sensitivity_appendix}
\begin{tabular}{lrr}
\toprule
Scenario & Hard-feasible geometries & Nondominated geometries \\
\midrule
Mean-only bounds & 9,409 & 363 \\
Global joint 90\% & 5,736 & 271 \\
Global joint 95\% & 2,998 & 198 \\
95\% without support limit & 3,077 & 200 \\
Lift tolerance 3\% & 40 & 24 \\
Lift tolerance 5\% & 2,998 & 198 \\
Lift tolerance 7\% & 7,078 & 294 \\
Load limits $Q_{90}$ & 2,170 & 166 \\
Load limits $Q_{95}$ & 2,998 & 198 \\
Load limits $Q_{99}$ & 4,031 & 201 \\
\bottomrule
\end{tabular}
\end{table}

Among the declared thresholds, lift tolerance exercts the biggest impact on the size of the feasible set. Narowing the tolerance from 5\% to 3\% reduces the number of feasible geometries from 2,998 to 40, whereas relaxing it to 7\% increases the count to 7,078. Using the 90th- and 99th-percentile load thresholds instead of the 95th-percentile load threshold yields 2,170 and 4,031 feasible geometries respectively. Mean-only screening is far less conservative and retains 9,409 geometries, while the global 90\% and 95\% intervals retain 5,736 and 2,998 geometries. Deleting the support cap does not significantly affect the number of geometries (from 2,998 to 3,077) but permits designs to be selected outside the declared training-support region. Taken together, these results support the use of the frozen baseline as an explicitly defined study configuration, rather than suggesting that the chosen thresholds are universally applicable.

For reference, the root-load coefficient thresholds are $(0.438288,0.0020535)$ at $Q_{90}$, $(0.465921,0.0022071)$ at $Q_{95}$, and $(0.502253,0.0025068)$ at $Q_{99}$ for shear and twist, respectively.

\end{appendices}

\section*{Data and code availability}
The data and code supporting the findings of this study will be available from the corresponding author upon reasonable request.

{\small
\setlength{\bibsep}{2pt}
\interlinepenalty=10000
\bibliography{sn-bibliography}

@article{martins2022aerodynamic,
  title={Aerodynamic design optimization: Challenges and perspectives},
  author={Martins, Joaquim RRA},
  journal={Computers \& Fluids},
  volume={239},
  pages={105391},
  year={2022},
  publisher={Elsevier}
}

@article{abergo2023aerodynamic,
  title={Aerodynamic shape optimization based on discrete adjoint and RBF},
  author={Abergo, Luca and Morelli, Myles and Guardone, Alberto},
  journal={Journal of Computational Physics},
  volume={477},
  pages={111951},
  year={2023},
  publisher={Elsevier}
}

@article{melin2000vortex,
  title={A vortex lattice MATLAB implementation for linear aerodynamic wing applications},
  author={Melin, Tomas},
  journal={Royal Institute of Technology, Sweden},
  volume={208},
  year={2000}
}

@article{leifsson2014fast,
  title={Fast low-fidelity wing aerodynamics model for surrogate-based shape optimization},
  author={Leifsson, Leifur and Koziel, Slawomir and Bekasiewicz, Adrian},
  journal={Procedia Computer Science},
  volume={29},
  pages={811--820},
  year={2014},
  publisher={Elsevier}
}

@article{paulson1976applications,
  title={Applications of vortex lattice theory to preliminary aerodynamic design},
  author={Paulson Jr, John W},
  journal={Vortex-Lattice Utilization},
  year={1976}
}

@article{blackwell1976induced,
  title={Numerical method to calculate the induced drag or optimum loading for arbitrary non-planar aircraft},
  author={Blackwell Jr, James A},
  journal={NASA. Langley Res. Center Vortex-Lattice Utilization},
  year={1976}
}

@article{deyoung1976optimum,
  title={Optimum lattice arrangement developed from a rigorous analytical basis},
  author={DeYoung, John},
  journal={NASA. Langley Res. Center Vortex-Lattice Utilization},
  year={1976}
}

@book{Forrester2008,
  title={Engineering design via surrogate modelling: a practical guide},
  author={Forrester, Alexander and Sobester, Andras and Keane, Andy},
  year={2008},
  publisher={John Wiley \& Sons}
}

@article{Jones1998,
  title={Efficient global optimization of expensive black-box functions},
  author={Jones, Donald R and Schonlau, Matthias and Welch, William J},
  journal={Journal of Global optimization},
  volume={13},
  number={4},
  pages={455--492},
  year={1998},
  publisher={Springer}
}

@article{asouti2023radial,
  title={Radial basis function surrogates for uncertainty quantification and aerodynamic shape optimization under uncertainties},
  author={Asouti, Varvara and Kontou, Marina and Giannakoglou, Kyriakos},
  journal={Fluids},
  volume={8},
  number={11},
  pages={292},
  year={2023},
  publisher={MDPI}
}

@article{li2021data,
  title={Data-based approach for wing shape design optimization},
  author={Li, Jichao and Zhang, Mengqi},
  journal={Aerospace Science and Technology},
  volume={112},
  pages={106639},
  year={2021},
  publisher={Elsevier}
}

@article{sabater2022fast,
  title={Fast predictions of aircraft aerodynamics using deep-learning techniques},
  author={Sabater, Christian and St{\"u}rmer, Philipp and Bekemeyer, Philipp},
  journal={Aiaa Journal},
  volume={60},
  number={9},
  pages={5249--5261},
  year={2022},
  publisher={American Institute of Aeronautics and Astronautics}
}

@article{zhang2024active,
  title={Active learning for efficient data-driven aerodynamic modeling in spaceplane design},
  author={Zhang, Hao and Huang, Wei and Shen, Yang and Xu, Da-yu and Niu, Yao-bin},
  journal={Physics of Fluids},
  volume={36},
  number={6},
  year={2024},
  publisher={AIP Publishing}
}

@article{tao2024multi,
  title={Multi-fidelity deep learning for aerodynamic shape optimization using convolutional neural network},
  author={Tao, Guocheng and Fan, Chengwei and Wang, Wen and Guo, Wenjun and Cui, Jiahuan},
  journal={Physics of Fluids},
  volume={36},
  number={5},
  year={2024},
  publisher={AIP Publishing}
}

@article{nikolaou2025multi,
  title={Multi-Fidelity Surrogate-Assisted Aerodynamic Optimization of Aircraft Wings},
  author={Nikolaou, Eleftherios and Kilimtzidis, Spyridon and Kostopoulos, Vassilis},
  journal={Aerospace},
  volume={12},
  number={4},
  pages={359},
  year={2025},
  publisher={MDPI}
}

@article{liu2025review,
  title={Review of deep learning-based aerodynamic shape surrogate models and optimization for airfoils and blade profiles},
  author={Liu, Xiaogang and Yang, Shengyu and Sun, Haifeng and Wang, Zhongyi and Guan, Xue and Gu, Yuanqi and Wang, Yuhang},
  journal={Physics of Fluids},
  volume={37},
  number={4},
  year={2025},
  publisher={AIP Publishing}
}

@article{lakshminarayanan2017deep,
  title={Simple and scalable predictive uncertainty estimation using deep ensembles},
  author={Lakshminarayanan, Balaji and Pritzel, Alexander and Blundell, Charles},
  journal={Advances in neural information processing systems},
  volume={30},
  year={2017}
}

@article{ovadia2019trust,
  title={Can you trust your model's uncertainty? evaluating predictive uncertainty under dataset shift},
  author={Snoek, Jasper and Ovadia, Yaniv and Fertig, Emily and Lakshminarayanan, Balaji and Nowozin, Sebastian and Sculley, D and Dillon, Joshua V and Ren, Jie and Nado, Zachary},
  year={2019}
}

@article{lei2018distribution,
  title={Distribution-free predictive inference for regression},
  author={Lei, Jing and G’Sell, Max and Rinaldo, Alessandro and Tibshirani, Ryan J and Wasserman, Larry},
  journal={Journal of the American Statistical Association},
  volume={113},
  number={523},
  pages={1094--1111},
  year={2018},
  publisher={Taylor \& Francis}
}

@inproceedings{johnstone2021conformal,
  title={Conformal uncertainty sets for robust optimization},
  author={Johnstone, Chancellor and Cox, Bruce},
  booktitle={Conformal and Probabilistic Prediction and Applications},
  pages={72--90},
  year={2021},
  organization={PMLR}
}

@article{feldman2023calibrated,
  title={Calibrated multiple-output quantile regression with representation learning},
  author={Feldman, Shai and Bates, Stephen and Romano, Yaniv},
  journal={Journal of Machine Learning Research},
  volume={24},
  number={24},
  pages={1--48},
  year={2023}
}

@article{toal2011multipoint,
  title={Efficient multipoint aerodynamic design optimization via cokriging},
  author={Toal, David JJ and Keane, Andy J},
  journal={Journal of Aircraft},
  volume={48},
  number={5},
  pages={1685--1695},
  year={2011}
}

@inproceedings{kenway2016multipoint,
  title={Multipoint aerodynamic shape optimization investigations of the common research model wing},
  author={Kenway, Gaetan K and Burdette, David A and Martins, Joaquim RRA},
  booktitle={53rd AIAA aerospace sciences meeting},
  pages={0264},
  year={2015}
}

@article{li2023efficient,
  title={Efficient data-driven off-design constraint modeling for practical aerodynamic shape optimization},
  author={Li, Jichao and He, Sicheng and Martins, Joaquim RRA and Zhang, Mengqi and Cheong Khoo, Boo},
  journal={AIAA Journal},
  volume={61},
  number={7},
  pages={2854--2866},
  year={2023},
  publisher={American Institute of Aeronautics and Astronautics}
}

@article{chen2024data,
  title={Data-driven aerodynamic shape design with distributionally robust optimization approaches},
  author={Chen, Long and Rottmayer, Jan and Kusch, Lisa and Gauger, Nicolas R and Ye, Yinyu},
  journal={arXiv preprint arXiv:2310.08931},
  year={2023}
}

@article{zhang2024robustkriging,
  title={An efficient robust aerodynamic design optimization method based on a multi-level hierarchical Kriging model and multi-fidelity expected improvement},
  author={Zhang, Yu and Han, Zhong-hua and Song, Wen-ping},
  journal={Aerospace Science and Technology},
  volume={152},
  pages={109401},
  year={2024},
  publisher={Elsevier}
}

@article{wu2024efficient,
  title={Efficient aerodynamic shape optimization using transfer learning based multi-fidelity deep neural network},
  author={Wu, Ming-Yu and He, Xian-Jun and Sun, Xiao-Hui and Tong, Ting-Shuai and Chen, Zhi-Hua and Zheng, Chun},
  journal={Physics of Fluids},
  volume={36},
  number={11},
  year={2024},
  publisher={AIP Publishing}
}

@article{yang2025operationaware,
  title={Operation-Aware Aircraft Wing Design Using Cluster-Based Multipoint Aerodynamic Shape Optimization},
  author={Yang, Aobo and Lyu, Yuan and Li, Jichao and Liem, Rhea P},
  journal={Journal of Aircraft},
  volume={62},
  number={6},
  pages={1531--1547},
  year={2025},
  publisher={American Institute of Aeronautics and Astronautics}
}

@article{karniadakis2021piml,
  title={Physics-informed machine learning},
  author={Karniadakis, George Em and Kevrekidis, Ioannis G and Lu, Lu and Perdikaris, Paris and Wang, Sifan and Yang, Liu},
  journal={Nature Reviews Physics},
  volume={3},
  number={6},
  pages={422--440},
  year={2021},
  publisher={Nature Publishing Group UK London}
}

@article{willard2023scientific,
  title={Integrating scientific knowledge with machine learning for engineering and environmental systems},
  author={Willard, Jared and Jia, Xiaowei and Xu, Shaoming and Steinbach, Michael and Kumar, Vipin},
  journal={ACM Computing Surveys},
  volume={55},
  number={4},
  pages={1--37},
  year={2022},
  publisher={ACM New York, NY}
}

@article{bakarji2022buckingham,
  title={Dimensionally consistent learning with Buckingham Pi},
  author={Bakarji, Joseph and Callaham, Jared and Brunton, Steven L and Kutz, J Nathan},
  journal={Nature Computational Science},
  volume={2},
  number={12},
  pages={834--844},
  year={2022},
  publisher={Nature Publishing Group US New York}
}

@article{villar2023dimensionless,
  title={Dimensionless machine learning: Imposing exact units equivariance},
  author={Villar, Soledad and Yao, Weichi and Hogg, David W and Blum-Smith, Ben and Dumitrascu, Bianca},
  journal={Journal of Machine Learning Research},
  volume={24},
  number={109},
  pages={1--32},
  year={2023}
}

@inproceedings{trabucco2021com,
  title={Conservative objective models for effective offline model-based optimization},
  author={Trabucco, Brandon and Kumar, Aviral and Geng, Xinyang and Levine, Sergey},
  booktitle={International Conference on Machine Learning},
  pages={10358--10368},
  year={2021},
  organization={PMLR}
}

@article{tibshirani2019covshift,
  title={Conformal prediction under covariate shift},
  author={Tibshirani, Ryan J and Foygel Barber, Rina and Candes, Emmanuel and Ramdas, Aaditya},
  journal={Advances in neural information processing systems},
  volume={32},
  year={2019}
}

@inproceedings{patel2024ccro,
  title={Conformal contextual robust optimization},
  author={Patel, Yash P and Rayan, Sahana and Tewari, Ambuj},
  booktitle={International Conference on Artificial Intelligence and Statistics},
  pages={2485--2493},
  year={2024},
  organization={PMLR}
}

@article{chenreddy2024ecro,
  title={End-to-end conditional robust optimization},
  author={Chenreddy, Abhilash and Delage, Erick},
  journal={arXiv preprint arXiv:2403.04670},
  year={2024}
}

@article{yeh2025endtoend,
  title={End-to-end conformal calibration for optimization under uncertainty},
  author={Yeh, Christopher and Christianson, Nicolas and Wu, Alan and Wierman, Adam and Yue, Yisong},
  journal={arXiv preprint arXiv:2409.20534},
  year={2024}
}

@article{breiman1996bagging,
  title={Bagging predictors},
  author={Breiman, Leo},
  journal={Machine learning},
  volume={24},
  number={2},
  pages={123--140},
  year={1996},
  publisher={Springer}
}

@book{katzPlotkin2001,
  title={Low-speed aerodynamics},
  author={Katz, Joseph and Plotkin, Allen},
  volume={13},
  year={2001},
  publisher={Cambridge university press}
}

@misc{drelaYoungrenAVL,
  author       = {Drela, Mark and Youngren, Harold},
  title        = {{AVL}: Aerodynamic Analysis, Trim Calculation, Dynamic Stability Analysis, and Aircraft Configuration Development},
  howpublished = {Massachusetts Institute of Technology},
  year         = {2026},
  note         = {Extended vortex-lattice software and documentation; accessed 28 July 2026},
  url          = {https://web.mit.edu/drela/Public/web/avl/}
}

@misc{MathWorksAeroFixedWing,
  author={{MathWorks}},
  title={{Aero.FixedWing}: Define Fixed-Wing Aircraft},
  year = {2026},
  howpublished={MATLAB Aerospace Toolbox documentation},
  note = {Accessed 26 July 2026},
  address={Natick, MA, USA},
  url = {https://www.mathworks.com/help/aerotbx/ug/aero.fixedwing-class.html}
}

@misc{MathWorksForcesAndMoments,
  author={{MathWorks}},
  title={{Aero.FixedWing.forcesAndMoments}: Calculate Forces and Moments of Fixed-Wing Aircraft},
  year = {2026},
  howpublished={MATLAB Aerospace Toolbox documentation},
  note = {Accessed 26 July 2026},
  address={Natick, MA, USA},
  url = {https://www.mathworks.com/help/aerotbx/ug/aero.fixedwing.forcesandmoments.html}
}

@misc{MathWorksStaticStability,
  author={{MathWorks}},
  title={{Aero.FixedWing.staticStability}: Calculate Static Stability of Fixed-Wing Aircraft},
  year = {2026},
  howpublished={MATLAB Aerospace Toolbox documentation},
  note = {Accessed 26 July 2026},
  address={Natick, MA, USA},
  url = {https://www.mathworks.com/help/aerotbx/ug/aero.fixedwing.staticstability.html}
}

@book{roskam1995flightdynamics,
  title={Airplane flight dynamics and automatic flight controls},
  author={Roskam, Jan},
  year={1998},
  publisher={DARcorporation}
}

@article{elfwing2018silu,
  title={Sigmoid-weighted linear units for neural network function approximation in reinforcement learning},
  author={Elfwing, Stefan and Uchibe, Eiji and Doya, Kenji},
  journal={Neural networks},
  volume={107},
  pages={3--11},
  year={2018},
  publisher={Elsevier}
}

@article{ba2016layernorm,
  title={Layer normalization},
  author={Ba, Jimmy Lei and Kiros, Jamie Ryan and Hinton, Geoffrey E},
  journal={arXiv preprint arXiv:1607.06450},
  year={2016}
}

@article{loshchilov2019adamw,
  title={Decoupled weight decay regularization},
  author={Loshchilov, Ilya and Hutter, Frank},
  journal={arXiv preprint arXiv:1711.05101},
  year={2017}
}

@incollection{huber1964robust,
  title={Robust estimation of a location parameter},
  author={Huber, Peter J},
  booktitle={Breakthroughs in statistics: Methodology and distribution},
  pages={492--518},
  year={1992},
  publisher={Springer}
}

@inproceedings{oza2001online,
  title={Online bagging and boosting},
  author={Oza, Nikunji C},
  booktitle={IEEE Conference on Systems, Man, and Cybernetics, Special Session on Ensemble Methods for Extreme Environments},
  year={2005}
}

@article{hoerl1970ridge,
  title={Ridge regression: Biased estimation for nonorthogonal problems},
  author={Hoerl, Arthur E and Kennard, Robert W},
  journal={Technometrics},
  volume={42},
  number={1},
  pages={80--86},
  year={2000},
  publisher={Taylor \& Francis}
}

@article{ke2017lightgbm,
  title={Lightgbm: A highly efficient gradient boosting decision tree},
  author={Ke, Guolin and Meng, Qi and Finley, Thomas and Wang, Taifeng and Chen, Wei and Ma, Weidong and Ye, Qiwei and Liu, Tie-Yan},
  journal={Advances in neural information processing systems},
  volume={30},
  year={2017}
}

@inproceedings{bostrom2020mondrian,
  title={Mondrian conformal regressors},
  author={Bostr{\"o}m, Henrik and Johansson, Ulf},
  booktitle={Conformal and probabilistic prediction and applications},
  pages={114--133},
  year={2020},
  organization={PMLR}
}

@article{sobol1967distribution,
  title={Distribution of points in a cube and approximate evaluation of integrals},
  author={Sobol, Ilya M},
  journal={USSR Computational mathematics and mathematical physics},
  volume={7},
  pages={86--112},
  year={1967}
}
}

\end{document}